\documentclass[twoside,11pt]{article}

\usepackage{amsthm}
\usepackage[preprint]{jmlr2e}

\usepackage{amsmath,mathtools}
\usepackage{microtype,needspace}
\usepackage{booktabs,longtable,enumitem,array}
\usepackage{xurl}
\usepackage{tikz}
\usetikzlibrary{arrows.meta,positioning,calc,decorations.pathreplacing}
\usepackage{lastpage}
\usepackage{dsfont}
\usepackage{algorithm}
\usepackage{algpseudocode}
\hypersetup{hidelinks}

\renewenvironment{proof}[1][Proof]{\par\noindent{\bf #1\ }}{\hfill\BlackBox\\[2mm]}

\newcommand{\R}{\mathbb R}
\newcommand{\E}{\mathbb E}
\newcommand{\Prob}{\mathbb P}
\newcommand{\eps}{\varepsilon}
\newcommand{\norm}[1]{\lVert#1\rVert}
\newcommand{\ip}[2]{\langle#1,#2\rangle}

\newcommand{\argmin}{\mathop{\rm argmin}}

\newcommand{\one}{\mathds 1}

\allowdisplaybreaks[1]
\jmlrheading{}{2026}{}{}{}{}{Yuxing Peng, Zhiqing Tang and Weijia Jia}
\ShortHeadings{Finite-Sum Minimax with Exact Proximal Oracles}{Peng, Tang and Jia}
\firstpageno{1}
\usepackage{xcolor,colortbl}
\begin{document}
\title{On the Complexity of Finite-Sum Nonconvex--Strongly-Concave Minimax Optimization with Exact Proximal Oracles}
\author{\name Yuxing Peng \email pyx26@mail.bnu.edu.cn\\
\addr Faculty of Arts and Sciences\\
Beijing Normal University, Zhuhai 519087, China
\AND
\name Zhiqing Tang\thanks{Corresponding author.} \email zhiqingtang@bnu.edu.cn\\
\addr Institute of Artificial Intelligence and Future Networks\\
Faculty of Arts and Sciences, Beijing Normal University, Zhuhai 519087, China
\AND
\name Weijia Jia \email jiawj@bnu.edu.cn\\
\addr Institute of Artificial Intelligence and Future Networks\\
Faculty of Arts and Sciences, Beijing Normal University, Zhuhai 519087, China\\
Guangdong Key Lab of AI and Multi-Modal Data Processing\\
Beijing Normal-Hong Kong Baptist University, Zhuhai 519087, China}
\editor{}
\maketitle
\begin{abstract}%
Finite-sum nonconvex--strongly-concave minimax problems arise in adversarial training and distributionally robust learning. Known lower bounds for these problems cover only linear-span algorithms, zero-respecting algorithms, or a restricted proximal protocol. This leaves open whether unrestricted adaptivity or exact component proximal steps, which return global saddle points of regularized component problems, can improve the polynomial complexity. We show that they cannot. Consider $n\ge2$ components with joint mean-squared smoothness $L$ and condition number $\kappa\ge1$, and a starting point with primal gap at most $\Delta$ and joint gradient norm at most $\sqrt{2L\Delta}$. For $\varepsilon^2/(L\Delta)$ below a universal constant, every randomized adaptive algorithm with exact component values, gradients, and proximal responses requires at least $\Omega(n+\min\{\sqrt n\,\kappa,n^{3/4}\sqrt\kappa\}L\Delta/\varepsilon^2)$ expected component queries in the worst case to find an $\varepsilon$-stationary point of the value function. A method that uses only component gradients attains this rate up to a factor $\log^2(2+\kappa)$, including its initialization cost. Zhou concurrently proves the same rate for zero-respecting first-order methods. For large condition numbers, our hard instances link primal stages through dual paths whose edges belong to random components, and with high probability a proximal query to a component reveals a path only up to the first edge owned by another component.
\end{abstract}

\begin{keywords}
minimax optimization, oracle complexity, finite-sum optimization, randomized lower bounds, proximal methods
\end{keywords}
\section{Introduction}
Finite-sum minimax problems arise when a learning objective is optimized against an adversary or against a worst-case weighting of the training examples, with one component per example, as in adversarial training \citep{madry2018}, distributionally robust optimization \citep{namkoong2016}, and agnostic federated learning \citep{mohri2019}. When the adversary's objective is concave and carries a strongly convex penalty, the inner maximization is strongly concave, and we obtain the nonconvex--strongly-concave (NC--SC) problem
\begin{equation}\label{eq:problem}
 \min_{p\in\R^{d_p}}V(p),\qquad V(p)=\max_{y\in\R^{d_y}}F(p,y),
 \qquad F=\frac1n\sum_{i=1}^nF_i,
\end{equation}
where $F$ is nonconvex in $p$, $\mu$-strongly concave in $y$, and the component gradients have joint mean-squared smoothness $L$. The condition number $\kappa=L/\mu$ measures the difficulty of the inner maximization. The goal is a point with a small gradient of the value function $V$, also called the primal function \citep{zhang2021,zhou2026ncsc}. Gradient descent ascent and its variance-reduced variants are standard methods for this problem \citep{lin2020gda,luo2020sreda}, and the oracle complexity of the problem is the number of component queries that the best method requires in the worst case.

Known lower bounds for finite-sum NC--SC problems cover linear-span algorithms \citep{zhang2021}, a restricted proximal protocol \citep{han2024}, and zero-respecting algorithms \citep{zhou2026ncsc}. None of them covers arbitrary randomized adaptive algorithms, which may choose components and query points based on the full transcript and a random seed. An exact component proximal query also carries more information than a gradient. It is available at every legal parameter, for which the regularized component problem is globally convex--concave and has a saddle point, and it returns a global saddle point, which can expose hidden directions that gradients do not reveal. Whether unrestricted adaptivity or exact proximal access can lower the complexity has therefore remained open.

We prove that neither of them changes the polynomial complexity. Write $S=L\Delta/\eps^2$ and $h(n,\kappa)=\min\{\sqrt n\,\kappa,n^{3/4}\sqrt\kappa\}$. On the class of Section~\ref{sec:model}, whose instances start from a point with primal gap at most $\Delta$ and joint gradient norm at most $\sqrt{2L\Delta}$, our lower and upper bounds on the oracle complexity are
\[
 \Omega\bigl(n+h(n,\kappa)S\bigr)\quad\text{and}\quad
 O\bigl((n+h(n,\kappa)S)\log^2(2+\kappa)\bigr),
\]
for $n\ge2$, every $\kappa\ge1$, and $\eps^2$ below a universal multiple of $L\Delta$, so they match up to the factor $\log^2(2+\kappa)$. Both bounds hold for the expected cost and for a deterministic query cap with success probability $2/3$ (Theorems~\ref{ak:thm:all} and~\ref{ak:thm:upper}), and for $\kappa\ge2^{18}\sqrt n$ the lower bound also holds under the protocol of \citet{han2024} (Corollary~\ref{cor:han}).

Where does the factor $n^{3/4}\sqrt\kappa$ come from at large condition numbers? Our hard instances chain $M\asymp S$ nonlinear primal stages through dual paths with $m$ nodes, and each internal edge of a path is assigned independently to one uniformly random component, its owner. A query to another component only rules that component out as the owner, so an algorithm is unlikely to find the owner with far fewer than order $n$ queries. For the mean $F$ to carry an edge with weight $a$, the owner must carry it with weight $na$, which adds a term of order $\sqrt{n^{-1}(na)^2}=\sqrt n\,a$ to the mean-squared smoothness constant and limits $a$ to at most order $L/\sqrt n$; we take $a$ of this order. A path with $m$ nodes and edge weight $a$ preserves the amplitude of its input from one end to the other up to a constant factor when its strong concavity is of order $a/m^2$, and this strong concavity must be at least $\mu$. Paths can therefore have $m\asymp\sqrt{a/\mu}\asymp\sqrt\kappa/n^{1/4}$ nodes. Each of the $M(m-1)$ owner searches costs at least order $n$ queries in expectation, so the total expected cost is at least of order
\begin{equation*}
 n\cdot m\cdot M\asymp n^{3/4}\sqrt\kappa\,\frac{L\Delta}{\eps^2}.
\end{equation*}

Exact proximal responses threaten this count in two ways. If the legal parameters depended on the component, testing legality alone would reveal owners; and since a response is a global saddle point, one query could reveal edges far beyond the current one. A public coordinate, with the same curvature in every component, makes the legal parameters identical for all components. A reflection symmetry confines the saddle point of a censored problem, in which the unrevealed hidden directions are set to zero, to the spans of the current query centers and of the blocks already revealed. Edge interactions vanish while the hidden coordinates are small, so with high probability this saddle point is the actual response, and a query to a component reveals only its run of consecutive edges and stops at the first edge with another owner.

Concurrent work by \citet[Theorems 1--3]{zhou2026ncsc} establishes the same polynomial rate for randomized zero-respecting incremental first-order methods, starting from an exact dual maximizer $y^*(0)=0$. Our lower bound removes the zero-respecting restriction, adds exact values and proximal responses, and also holds under that start (Section~\ref{sec:assumptions}). Our upper bound starts instead from a joint-gradient bound and pays for the resulting initialization. \citet{wu2026ncfinite} prove lower bounds for arbitrary randomized algorithms in the related finite-sum nonconvex--concave (NC--C) problem over a bounded dual domain, with values and gradients only. Table~\ref{tab:comparison} compares these bounds, and Section~\ref{sec:assumptions} compares our assumptions with those of prior work.

\begin{table}[t]
\centering
\renewcommand{\arraystretch}{1.6}
\setlength{\tabcolsep}{5pt}
\begin{tabular}{@{}lccl@{}}
\toprule
 & \multicolumn{2}{c}{Oracle} & \\
\cmidrule(lr){2-3}
Algorithm class & Component gradients & \shortstack{Exact values, gradients\\and proximal responses} & Source \\
\midrule
\shortstack[l]{Linear-span or\\restricted protocol} & $\Omega(n+\sqrt{n\kappa}\,S)$, $\kappa\ge2n$ & $\Omega(n+\sqrt{n\kappa}\,S)$, $\kappa=\Omega(n)$ & \shortstack[l]{\citeauthor{zhang2021};\\\citeauthor{han2024}} \\
\midrule
Zero-respecting & $\Omega(n+hS)$ & -- & \citeauthor{zhou2026ncsc} \\
\midrule
\shortstack[l]{Arbitrary randomized\\adaptive} & \cellcolor{gray!16}$\Omega(n+hS)$ & \cellcolor{gray!16}$\Omega(n+hS)$ & Thm.~\ref{ak:thm:all} \\
\midrule
Matching upper bound & \multicolumn{2}{c}{\cellcolor{gray!16}$O\bigl((n+hS)\log^2(2+\kappa)\bigr)$} & Thm.~\ref{ak:thm:upper} \\
\bottomrule
\end{tabular}
\caption{Bounds on component queries for finite-sum NC--SC problems with joint mean-squared smoothness, for small $\eps$ and every $\kappa\ge1$ unless stated otherwise; $h=\min\{\sqrt n\kappa,n^{3/4}\sqrt\kappa\}$ and $S=L\Delta/\eps^2$. Shaded: this paper. The first row combines \citet{zhang2021}, for linear-span methods, and \citet[Theorem 25]{han2024}, for their restricted proximal protocol with primal nonconvexity parameter $L$; the second is the concurrent lower bound of \citet[Theorem 2]{zhou2026ncsc}, whose Theorem 1 also gives a matching upper bound up to logarithmic factors from an exact dual maximizer. A dash marks a combination not treated separately. Our lower bound $\Omega(n+hS)$ covers the largest algorithm class and the stronger oracle; a method using only component gradients attains this rate up to the factor $\log^2(2+\kappa)$.}
\label{tab:comparison}
\end{table}

\section{Models and Main Results}\label{sec:model}
Our lower and upper bounds are stated for the same class of finite sums and the same cost model. They differ in the oracle: the lower bounds allow exact proximal responses, whereas our upper bounds use only component gradients.
\subsection{Strongly Concave Finite Sums and Exact Indexed Oracles}
Write $[n]=\{1,\ldots,n\}$, $w=(p,y)$, and $S=L\Delta/\eps^2$. With the dual space $\mathcal Y=\R^{d_y}$, the class $\mathcal F(n,L,\mu,\Delta)$ consists of the finite sums $F=\frac1n\sum_{i=1}^nF_i$ with continuously differentiable components that satisfy, globally,
\begin{align}
 \frac1n\sum_i\norm{\nabla F_i(w)-\nabla F_i(w')}^2
 &\le L^2\norm{w-w'}^2,\label{eq:rms}\\
 F(p,\cdot)&\text{ is $\mu$-strongly concave},\label{eq:sc}\\
 V(0)-\inf_pV(p)&\le\Delta,\label{eq:gap}\\
 \norm{\nabla F(0,0)}^2&\le2L\Delta.\label{eq:start}
\end{align}
All norms are Euclidean; for matrices and multilinear maps, $\norm\cdot$ is the induced operator norm. We call condition~\eqref{eq:rms} joint mean-squared smoothness, or root-mean-square (RMS) smoothness for short. It bounds only an average over the components: it implies that the mean $F$ is $L$-smooth and that each component is at most $\sqrt n\,L$-smooth. By the strong concavity~\eqref{eq:sc}, for every $p$ the function $F(p,\cdot)$ has a unique maximizer $y^*(p)$, and $\nabla V(p)=\nabla_pF(p,y^*(p))$. We call conditions~\eqref{eq:gap} and~\eqref{eq:start} together a normalized start. They describe the starting point by its primal gap and its joint gradient, without requiring that its dual part be an exact maximizer. We write $G_0=\sup_yF(0,y)-\inf_pF(p,0)$ for the initial primal--dual gap; it differs from the primal gap in~\eqref{eq:gap}, and the class places no bound on it.

For a component index $i$ and a point $w$, an exact indexed first-order query returns
\begin{equation}\label{eq:oracle}
 \mathcal O(i,w)=(F_i(w),\nabla_pF_i(w),\nabla_yF_i(w)).
\end{equation}
The component saddle-proximal oracle performs a proximal point step \citep{rockafellar1976} on one component. For a parameter $\gamma>0$, it returns, together with the first-order information~\eqref{eq:oracle}, a saddle pair of
\begin{equation}\label{eq:prox}
 H_{i,\gamma}(p,y)=F_i(p,y)+\frac{\norm{p-\bar p}^2-\norm{y-\bar y}^2}{2\gamma}
\end{equation}
where $(\bar p,\bar y)$ is the queried center. Since a component may be nonconvex in $p$, the regularized problem~\eqref{eq:prox} need not be convex--concave for every $\gamma$, so the oracle answers only at legal parameters. As in Equation~2 and Remark 12(iv) of \citet{han2024}, a finite $\gamma>0$ is legal when $H_{i,\gamma}$ is globally convex in $p$ and concave in $y$ and has a saddle point. A saddle pair $(p^+,y^+)$ satisfies $H_{i,\gamma}(p^+,y)\le H_{i,\gamma}(p^+,y^+)\le H_{i,\gamma}(p,y^+)$ for all $p$ and $y$.

One query to either oracle costs one unit, a full gradient costs $n$ units, and $Q$ denotes the total cost of a run. At each query, the algorithm chooses the component index, a finite query point and, for a proximal query, a legal parameter based on the entire exact transcript and a private random seed. Its computation and its primal output are unrestricted, and it may stop at a random time. Apart from the oracle answers, the algorithm is given only the dimensions, the target accuracy $\eps$, and the class parameters, which are the same for every instance of the class.

\begin{theorem}[Unrestricted lower bound for every condition number]
\label{ak:thm:all}
There are universal constants $c_\star\in(0,1/2]$ and $c>0$ such that,
for $n\ge2$, $L,\mu,\Delta>0$, $\kappa=L/\mu\ge1$, and
\begin{equation}\label{ak:eq:all-window}
 0<\eps^2\le2^{-52}c_\star^2L\Delta,
\end{equation}
every algorithm using either exact indexed oracle above, terminating
almost surely and guaranteeing $\E\norm{\nabla V(p_{\rm out})}\le\eps$ uniformly
on $\mathcal F(n,L,\mu,\Delta)$ has worst-case expected cost at least
\begin{equation}\label{ak:eq:all-rate}
 c\left(n+\min\{\sqrt n\,\kappa,n^{3/4}\sqrt\kappa\}
                 \frac{L\Delta}{\eps^2}\right).
\end{equation}
A lower bound of the same order holds for success probability $2/3$ under a
deterministic query cap. The hard instances have polynomial dimension,
$C^{1,1}$ components that are individually $\mu$-strongly concave in $y$,
$y^*(0)=0$, finite initial primal--dual gap $G_0$, and a unique response at every legal
proximal parameter. Their exact legal domain is common to all
components, all hidden realizations and all proximal centers.
\end{theorem}
The accuracy condition~\eqref{ak:eq:all-window} asks only that $\eps^2/(L\Delta)$ be below a universal constant. We prove Theorem~\ref{ak:thm:all} in Appendix~\ref{ak:app:all}, where we obtain $c=c_\star^2/2^{96}$.

\begin{theorem}[Upper bound for all condition numbers and matching polynomial rate]
\label{ak:thm:upper}
For $n\ge2$, $\kappa\ge1$ and $S=L\Delta/\eps^2\ge2$, there is an indexed
component-gradient algorithm on $\mathcal F(n,L,\mu,\Delta)$ that terminates
almost surely and satisfies
\begin{align}
 \E\|\nabla V(p_{\rm out})\|^2&\le\eps^2,\label{ak:eq:upperacc}\\
 \E Q&\le C\bigl[n+h(n,\kappa)(1+S)\bigr]
                    \log^2(2+\kappa),\label{ak:eq:upper}\\
 h(n,\kappa)&=\min\{\sqrt n\kappa,n^{3/4}\sqrt\kappa\}.\notag
\end{align}
Here $C>0$ is universal, and all initialization, snapshots and final
refinement calls are included. The same order holds for a deterministic
query cap with success probability $2/3$.
Under \eqref{ak:eq:all-window}, we have $S\ge2$ and
$\E\norm{\nabla V(p_{\rm out})}\le(\E\norm{\nabla V(p_{\rm out})}^2)^{1/2}\le\eps$, so this upper bound is at most a constant multiple of
$\log^2(2+\kappa)$ times the lower bound of Theorem~\ref{ak:thm:all}.
\end{theorem}
We prove Theorem~\ref{ak:thm:upper} in Appendix~\ref{nu:sec:assembly} by combining two methods: the method of Appendix~\ref{nu:app:upper} for $\kappa\le2^{18}\sqrt n$, and the method of the next theorem for larger $\kappa$. We state the regime of large condition numbers separately because there the construction with long paths gives a larger constant $c$, the bound $G_0\le1025\Delta$, the dimension estimate~\eqref{eq:dimension-main}, and a single hard instance for all early horizons under the protocol of \citet{han2024}.

\begin{theorem}[Special case of large condition numbers and matching upper bound]\label{rb:thm:main}
There are universal constants $c_\star\in(0,1/2]$ and $c,C>0$ such that, for $n\ge2$, $L,\mu,\Delta>0$, and
\begin{equation}\label{rb:eq:window}
 \kappa=L/\mu\ge2^{18}\sqrt n,\qquad
 0<\eps^2\le2^{-52}c_\star^2L\Delta,
\end{equation}
every algorithm using either oracle above, terminating almost surely and guaranteeing $\E\norm{\nabla V(p_{\rm out})}\le\eps$ uniformly on $\mathcal F(n,L,\mu,\Delta)$, has worst-case expected query cost at least
\begin{equation}\label{rb:eq:main-rate}
 c n^{3/4}\sqrt\kappa\,S.
\end{equation}
For each algorithm, the hard instance that witnesses this bound can be chosen with $C^{1,1}$ components, each $\mu$-strongly concave in $y$, and with $y^*(0)=0$ and $G_0\le1025\Delta$.

For every $\eps>0$ and $\kappa\ge2^{18}\sqrt n$, without the accuracy condition in Equation~\ref{rb:eq:window}, our method of Appendix~\ref{app:upper}, which combines existing first-order methods, achieves
\begin{align}
 \E\norm{\nabla V(p_{\rm out})}^2&\le\eps^2,\label{eq:upperacc}\\
 \E Q&\le Cn^{3/4}\sqrt\kappa(1+S)\log^2(2+\kappa).\label{eq:upper}
\end{align}
In the respective regimes of the lower and upper bounds, the same orders hold for success probability $2/3$ under a deterministic query cap.
\end{theorem}
Our proof gives $c=133c_\star^2/(3\cdot2^{73})$, where $c_\star$ is determined by the minima of continuous residuals over three fixed-dimensional compact sets (Lemma~\ref{rb:match:lem:uniform-barrier}). The primal and dual dimensions $d_p$ and $d_y$ of the witnessing hard instances are fixed before the algorithm runs and satisfy
\begin{equation}\label{eq:dimension-main}
 d_p+d_y=O\!\left(\left(Sn^{3/4}\sqrt\kappa+
 S^2\kappa^3n^{-3/2}\right)\log(n\kappa S)\right).
\end{equation}

\begin{corollary}[The Han et al.\ proximal protocol]\label{cor:han}
\looseness=-1 In the regime of Theorem~\ref{rb:thm:main}, a lower bound of the same order also holds for the proximal incremental first-order oracle (PIFO) algorithms of \citet[Definition 11]{han2024}, with their complexity $\sup_F\inf_t$ of Definition 13 measured by the expected gradient norm.
\end{corollary}
In the proof of Corollary~\ref{cor:han} (Appendix~\ref{app:han}), one fixed hard instance serves every early output horizon, and the cost counts both component calls and full-gradient refreshes. The upper bound in Theorem~\ref{rb:thm:main} is stated in the indexed oracle model.

\subsection{Comparison with Prior Assumptions}\label{sec:assumptions}
\par\textit{Smoothness.} The joint mean-squared smoothness condition~\eqref{eq:rms} is Assumption~1 of \citet{zhou2026ncsc} and the smoothness condition of the finite-sum lower bounds of \citet{zhang2021} and \citet{han2024}. It is weaker than $L$-smoothness of every component, which \citet{wu2026ncfinite} use for their lower bounds, and stronger than $L$-smoothness of the mean alone. Our upper bounds need only~\eqref{eq:rms}. The components of our hard instances use the extra freedom that~\eqref{eq:rms} allows: their individual smoothness constants grow proportionally to $\sqrt n\,L$.

\par\textit{Dual variable.} As in \citet[Assumption 3]{zhou2026ncsc}, the dual space is $\R^{d_y}$ and the mean is $\mu$-strongly concave; the components of our hard instances are even individually $\mu$-strongly concave. \citet{wu2026ncfinite} instead use a known compact convex dual set of diameter at most $D_y$ with a concave mean, and they measure stationarity through the Moreau envelope of the value function.

\par\textit{Initialization.} The normalized start~\eqref{eq:gap}--\eqref{eq:start} bounds the primal gap and the joint gradient at the origin. \citet{zhou2026ncsc} assumes instead a bound on the primal gap and an exact dual maximizer $y^*(0)=0$, and \citet{wu2026ncfinite} assume only a bound on the primal gap; in their setting, the bounded dual domain controls the dual side. Neither our start nor the exact dual start of \citet{zhou2026ncsc} implies the other: $y^*(0)=0$ does not imply the bound~\eqref{eq:start}, and a bounded joint gradient does not give $y^*(0)=0$. Our hard instances satisfy both conditions, so Theorem~\ref{ak:thm:all} also holds under the start of \citet{zhou2026ncsc}. Every hard instance also has a finite initial primal--dual gap (Appendix~\ref{ak:sec:assembly}).

\par\looseness=-1\textit{Accuracy and cost.} Theorem~\ref{ak:thm:all} bounds the expected cost of algorithms with $\E\norm{\nabla V(p_{\rm out})}\le\eps$, and the deterministic cap needed for confidence $2/3$; Theorem~\ref{ak:thm:upper} guarantees $\E\norm{\nabla V(p_{\rm out})}^2\le\eps^2$. \citet{zhou2026ncsc} characterizes expected query counts under $\E\norm{\nabla V}^2\le\eps^2$, with a fixed budget for $\kappa\le\sqrt n$, and that lower bound also covers the criterion $\E\norm{\nabla V(p_{\rm out})}\le\eps$ and success probability $2/3$ under a fixed budget; \citet{wu2026ncfinite} use fixed budgets and confidence $2/3$. The remaining gap between the upper and lower bounds is $\log^2(2+\kappa)$ here and logarithmic for \citet{zhou2026ncsc}; for \citet{wu2026ncfinite} there is no gap when $LD_y/\eps\le n^{1/4}$ and a single logarithmic factor otherwise.

\section{Technical Overview}
Our lower bounds hide a chain of nonlinear primal stages that an algorithm must uncover one at a time, and for $\kappa\ge64$ the hard instances link consecutive stages through dual paths whose edges belong to random components. We first describe the hard instances for large condition numbers and then explain how they stay hidden from arbitrary queries and from exact proximal responses. The symbols of this section are local to the lower bounds.

\subsection{Dual Paths with Component Assignments}\label{sec:construction}
\begin{figure}[t]
\centering
\begin{tikzpicture}[>=Stealth, node distance=1.35cm, every node/.style={font=\small}]
\node[draw,rounded corners,minimum height=.65cm] (x) {primal stage $j$};
\node[circle,draw,right=1cm of x] (u1) {$U_1$};
\node[circle,draw,right=of u1] (u2) {$U_2$};
\node[circle,draw,right=of u2] (u3) {$U_3$};
\node[circle,draw,right=of u3] (u4) {$U_4$};
\node[draw,rounded corners,right=1cm of u4,minimum height=.65cm] (z) {stage $j+1$};
\draw[->] (x)--node[above=5pt,font=\scriptsize]{public}(u1);
\draw[thick] (u1)--node[above]{owner $2$}(u2);
\draw[thick] (u2)--node[above]{owner $2$}(u3);
\draw[dashed] (u3)--node[above]{owner $5$}(u4);
\draw[->] (u4)--node[above=5pt,font=\scriptsize]{public}(z);
\draw[->,thick] ([yshift=-.6cm]u1.south)--node[below]{query $i=2$: traverse the entire same-owner run}([yshift=-.6cm]u3.south);
\node[below=1.05cm of u3,align=center] {next edge requires a new owner search};
\end{tikzpicture}
\caption{A dual path. In the ideal exposure process of Appendix~\ref{rp:sec:proof}, a query to component $i$ reveals the entire run of consecutive edges owned by $i$ and stops at the first edge with another owner; the query reveals only that $i$ does not own that edge, and no later owner or axis. There are of order $M m$ serial searches, and each costs at least order $n$ queries in expectation.}
\label{fig:owner-path}
\end{figure}
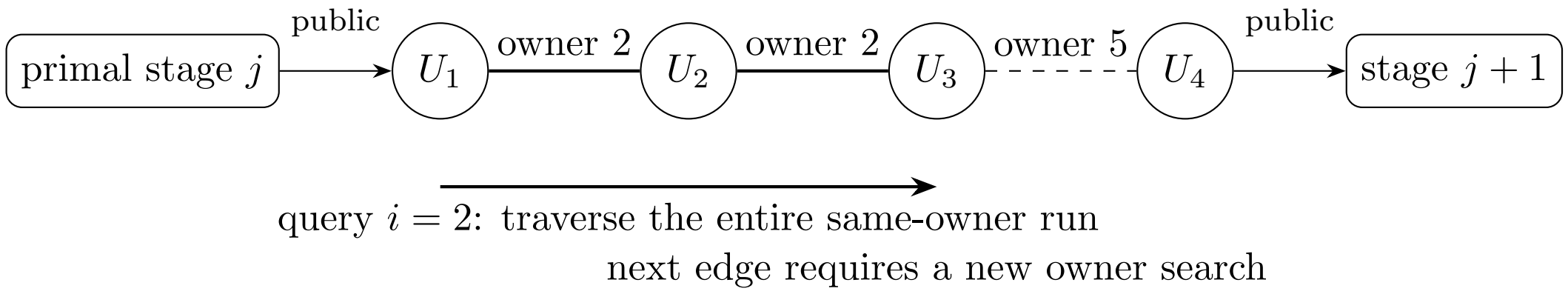

As described in the introduction, each primal stage passes its progress to the next through a dual path with $m$ nodes, and each internal edge of a path belongs to one uniformly random component, its owner, which carries the edge interaction with weight $na$; the edge weight $a$ is of order $L/\sqrt n$. When the flat zones introduced below have zero width and the cutoffs are inactive, the quadratic dual term of the mean is $-\frac a2t^\top Qt$ in the hidden coordinates $t$ of a path, with $Q=P_m+m^{-2}I$, where $P_m$ is the Laplacian of the path graph on $m$ vertices. The ridge $m^{-2}$ supplies the strong concavity $a/m^2\ge\mu$, and Lemma~\ref{rb:lem:green} gives the endpoint entries $u=(Q^{-1})_{11}\asymp m$ and $v=(Q^{-1})_{1m}\asymp m$ of the Green matrix, with $1<u/v\le\cosh1$. A path with $m\asymp\sqrt{a/\mu}\asymp\sqrt\kappa/n^{1/4}$ nodes therefore passes the forcing at one endpoint to the other up to a constant factor. Figure~\ref{fig:owner-path} shows how queries interact with the owners, and Appendix~\ref{app:robust} fixes the constants in Equations~\ref{rb:eq:base-parameters} and~\ref{rb:eq:green-parameters}.

Every node of every path has an independent hidden axis, and the owners are independent of the axes. The primal variable $P=(P_1,\ldots,P_M)$ has one block $P_j=(X_j,Z_{j+1})$ per stage, and public identity maps pair its two subblocks with the first and the last node of path $j$. Each component has the form
\begin{equation*}
\begin{aligned}
 f_{i,\eta}(P,y)={}&\mathcal P(P)-\frac\nu2\norm y^2+b\chi_*(Y_\partial)\ip{\mathcal A(P)}{Y_\partial}+c\sum_v\chi_*(y_v)\phi_\eta(t_v)\\
 &-na\sum_{e:\Sigma_e=i}A_\eta(Y_e)+(\text{public radial terms});
\end{aligned}
\end{equation*}
see Equation~\ref{rb:eq:components}. Here $\mathcal P$ collects a scaled source chain built from the scalar functions of \citet{carmon2020}, the compensation $-d\norm{\mathcal A}^2$, and a radial penalty. The forcing $\mathcal A(P)$ stacks the gated and capped primal blocks and acts on the concatenated path endpoints $Y_\partial$. Each dual node $v$ has a hidden coordinate $t_v$ along its axis, $\phi_\eta(t)=\frac12(|t|-\eta)_+^2$ vanishes on the flat zone $|t|\le\eta$, and $\chi_*$ is a cutoff. The owned term $A_\eta(Y_e)=\frac12(|t_e|-h_\eta(\norm{Y_e}))_+^2$ acts on the difference $t_e$ of the two hidden coordinates of an edge $e$ with endpoint pair $Y_e$, and only the owner $\Sigma_e$ of $e$ contains it. The width $\eta\ge0$ of the flat zones is fixed in Appendix~\ref{rb:sec:width}, and $a,b,c,d,\nu>0$ are constants.

Why do the paths not weaken the primal chain? With the primal stage scale $\lambda$ of Equation~\ref{rb:eq:green-parameters}, we calibrate the ridge $\nu=a/u$, the coupling $b^2=12a\lambda/v$, and the compensation $d=6\lambda(u/v-1)$. At zero width, eliminating the dual variable then contributes, for each stage $j$, the squared gate $h_j^2$ of that stage times the completed quadratic (Lemma~\ref{rb:lem:uncut})
\begin{equation}\label{eq:macro-quadratic}
 6\lambda(x_j-z_{j+1}/2)^2+6\lambda\|A_{j,\perp}\|^2
       +\tfrac32\lambda\|B_{j+1,\perp}\|^2,
\end{equation}
where $x_j$ and $z_{j+1}$ are the coordinates of the capped subblocks of $P_j$ along the first and the last hidden axis of path $j$, and $A_{j,\perp}$ and $B_{j+1,\perp}$ are their orthogonal parts. Dual elimination thus preserves the same normalized primal quadratic, including in the orthogonal directions that unrestricted outputs can use. The path and source ingredients follow \citet{li2021} and \citet{carmon2020}.

\Needspace{12\baselineskip}
\subsection{Global Geometry and Exact Hiding}\label{sec:localization}
Against zero-respecting algorithms \citep{zhou2026ncsc}, every query lies in the span of the coordinates that earlier gradients have touched, so the unexposed coordinates are exactly zero at every query. An arbitrary algorithm can instead query points with nonzero projections onto unexposed hidden axes and with arbitrary norms, and an exact proximal response is a global saddle point. What keeps the hidden structure invisible in this case?

Smooth radial maps \citep[cf.][Section 5.2]{carmon2020} compress the primal blocks to bounded images, so the forcing and its derivatives stay bounded at every query (Lemma~\ref{rb:match:lem:forcing}). Dual cutoffs with bounded support control the primal Hessian of the coupling. An owned edge interaction vanishes on its flat zone (cf.~\citealp[Appendix C]{woodworth2016}), so a small projection onto unexposed axes leaves the edge invisible. A flat zone of fixed width would not hide an edge from queries of large norm, so the width grows with the radius $r=\norm{Y_e}$ of the edge pair beyond a localization radius $R_*$:
\begin{align}
 h_\eta(r)&=\eta\varrho(r/R_*),\label{eq:radial-width-main}\\
 \varrho(u)&=\begin{cases}1,&u\le1,\\1+(u-1)^2/2,&1<u<2,\\u-1/2,&u\ge2.\end{cases}
 \label{eq:profile-main}
\end{align}
Since $\varrho(u)\ge u/2$, flatness holds on a cone at every radius.

These devices perturb the instance, and the perturbation must not destroy the terminal barrier. Public radial penalties, common to all components, dominate the positive curvature of the cutoffs without revealing owners. For every primal point, the dual maximizer of the mean without cutoffs and radial corrections has every relevant node, edge pair, and endpoint tuple strictly inside radius $3R_*/16$. There the cutoffs equal one and the radial corrections vanish, so strong concavity makes it the maximizer of the actual mean as well. Using the radial calculus of Lemma~\ref{rb:lem:radial}, Lemma~\ref{rb:lem:class} shows that the components are smooth and componentwise strongly concave, and Lemma~\ref{rb:lem:width-start} completes class membership. With the primal length scale $s=8\eps/(c_\star\lambda)$, the barrier of the zero-width chain on the last stage coordinate $z_{M+1}$ survives the perturbation (Lemmas~\ref{rb:match:lem:uniform-barrier} and~\ref{rb:lem:width-start}):
\begin{equation*}
 |z_{M+1}|\le s/8\quad\Longrightarrow\quad\|\nabla V(P)\|\ge6\eps.
\end{equation*}
A point whose last stage coordinate is this small is therefore far from $\eps$-stationary.

\subsection{Exact Oracle Responses and Component Search}\label{sec:search}
We now address the two threats from exact proximal responses described in the introduction: a legal domain that depends on the component, and a global saddle point that reveals edges far beyond the current one.

Against the first threat, we add one public primal coordinate $\xi$ with the common term $(L/8)(1-\cos\xi)$ to every component. This term preserves all initial budgets and gradient barriers. Its curvature reaches $-L/8$, while the hidden part of every component has primal curvature below $L/32$ (Lemma~\ref{rb:lem:class}), so the public term alone decides legality: the exact legal domain is $\tau\ge L/8$ for the inverse step $\tau=1/\gamma$, the same for every component (Lemma~\ref{rp:lem:public-legality}). In the regularized problem the objective is strongly convex--concave in $(P,y)$, and the separate scalar problem in $\xi$ is strictly convex even at the boundary.

Against the second threat, consider a legal call to a component $i$. It can reveal the whole run of consecutive edges that $i$ owns (Figure~\ref{fig:owner-path}), and we show that, with high probability, it reveals nothing beyond that run except that the next edge, if any, is not owned by $i$. Beyond the run, we set the unrevealed hidden axes to zero in every term and keep all public terms; this censored problem on the full space has a unique saddle point. We reflect the unrevealed blocks through the spans of their current query centers, using matching reflections for paired primal and dual endpoints. Invariance and uniqueness then place the unrevealed blocks of the censored saddle point in these spans, each of dimension at most two: one for an interior node, and two at an endpoint, whose span also contains the paired primal center. This step uses no property of the unrevealed axes; their random directions enter only in the next step. With high probability, the unrevealed axes have small projections onto these spans, so the actual and the censored problems have the same values and full gradients wherever the unrevealed blocks lie in these spans, and the censored saddle point is the actual response (Lemma~\ref{rp:lem:localization}). A separate projection argument handles the arbitrary final output.

Locating the owners is a sequence of $E=M(m-1)$ serial searches. Given the history of these searches, the owner of the first unrevealed edge is uniform over the components that have not yet been excluded. A nonnegative posterior potential starts at $E(n-1)/2$ and decreases in conditional expectation by at most one per call, including calls that reveal several consecutive edges (Lemma~\ref{rp:lem:posterior-cost}). Together with the terminal barrier, this bound on completing the searches shows that, even under random stopping, every algorithm with $\E\norm{\nabla V(p_{\rm out})}\le\eps$ uniformly on the class makes at least order $nMm$ queries in expectation over the random owners and axes. Some fixed choice of component functions therefore inherits this bound, as in Yao's minimax principle \citep{yao1977}.

\subsection{The Remaining Condition Numbers}
For $64\le\kappa\le2^{18}\sqrt n$, we use paths with two nodes. With $r=\mu/a$, they keep the calibration $\lambda=2^{-39}Lv/u=2^{-39}L/(1+r)$ and preserve~\eqref{eq:macro-quadratic}, and their owner searches cost at least order $nM$ queries in expectation, which is of order $\sqrt n\,\kappa S$ (Corollary~\ref{ak:cor:middle} in Appendix~\ref{ak:sec:middle}). We use paths only for $\kappa\ge64$: their dual ridge and cutoff barriers add about $47\nu$ to our smoothness bound (Lemma~\ref{rb:lem:class}), and $\nu\ge\mu=L/\kappa$.

For $1\le\kappa<64$, the hard family has no hidden dual path: the dual variable enters only through a decoupled strongly concave term, and each stage is linked to the next by a primal cross term $C_\eta(t_1,t_2)=\phi_\eta(t_1+t_2)-\phi_\eta(t_1)-\phi_\eta(t_2)$ owned by one random component, where $\phi_\eta(t)=\tfrac12(|t|-\eta)_+^2$. This cross term equals $t_1t_2$ at zero width and is flat near the hidden origin. Independent owners cost only a factor $\sqrt n$ in RMS smoothness, but a single component can then have primal curvature up to order $\sqrt n\,L$, so without a correction the legal parameters of a component would depend on what it owns. Each component $i$ therefore has its own public coordinate $\xi_i$ with the term $B(1-\cos\xi_i)$, where $B=\sqrt n\,L/4$. These orthogonal coordinates cost only $B/\sqrt n=L/4$ in RMS smoothness and give every component the exact legal domain $\tau\ge B$. Appendices~\ref{ak:sec:primal} and~\ref{ak:sec:primal-oracle} treat this family (Lemmas~\ref{ak:lem:primal-class}--\ref{ak:lem:primal-cost}). Each family is used only within its accuracy window (Equation~\ref{ak:eq:path-window}, with its own parameter $\lambda$). Outside this window, and for the term $n$, we use a separate instance in which one component, unknown to the algorithm, carries a random sign (Lemma~\ref{ak:lem:n-baseline}).

\section{Matching Upper Bound and Initialization Cost}\label{sec:upper}
Our upper bound uses only component gradients, and it needs strong concavity only of the mean. We match the two branches of $h(n,\kappa)$ with two methods: a single-loop method for $\kappa\le2^{18}\sqrt n$ and an accelerated method for larger $\kappa$. Both methods begin at the normalized start and pay for computing a good dual point themselves.

For $\kappa\le2^{18}\sqrt n$, Algorithm~\ref{alg:normalized} in Appendix~\ref{nu:app:upper} combines the stochastic projected damped extragradient (SPDE) updates of \citet{zhang2026spde} with recursive finite-sum gradient estimates \citep{nguyen2017sarah,fang2018spider} and loopless stochastic variance-reduced extragradient (L-SVRE) steps \citep{alacaoglu2022,luo2022}, which need strong monotonicity only of the mean operator. The analysis tracks an energy that combines the gap of the Moreau envelope $\mathcal M(z)=\min_p\{V(p)+L\norm{p-z}^2\}$ at a moving primal center $z$ with a term that controls the distance to the saddle point of $F(p,y)+L\norm{p-z}^2$. The normalized start bounds this distance only by $\kappa\sqrt{2\Delta/L}$. We prove the L-SVRE contraction directly (Lemma~\ref{nu:lem:lsvre}), and running L-SVRE at the input center, at expected query cost $O((n+\sqrt n\kappa)\log(2\kappa))$, brings the expected energy below $2\Delta$. At the end, the same solver refines a sampled endpoint, whose distance to its proximal saddle point is already controlled by the SPDE dissipation. \citet{zhang2026spde} analyze SPDE with a compact dual domain; we apply their one-step inequalities on the whole space by placing each step in balls on which the projections are inactive (Lemma~\ref{nu:lem:pathwise}). Our cost count includes every full snapshot and both gradients evaluated in each recursive difference. With the pathwise drift and motion bounds of Lemma~\ref{nu:lem:pathwise}, Theorem~\ref{nu:thm:normalized} then gives the bound $O(n+\sqrt n\kappa S+(n+\sqrt n\kappa)\log(2\kappa))$, with no logarithmic factor on the accuracy term.

For larger $\kappa$, the method of Appendix~\ref{app:upper} uses Catalyst acceleration. We normalize $f_i=F_i/L$, $\nu=\mu/L$, $D_0=\Delta/L$, and $e_0=\eps/L$, and set
\begin{equation}\label{eq:upperparams}
 s=n^{-1/2},\qquad \tau=s-\nu,\qquad q=\nu/s,\qquad\rho=0.9\sqrt q.
\end{equation}
Catalyst subtracts the dual regularizer $\frac\tau2\norm{y-z_k}^2$ at dual centers $z_k$, so each subproblem is $s$-strongly concave, and the accuracy targets decrease like $(1-\rho)^k$. The regime of Theorem~\ref{rb:thm:main} ensures $\tau>0$ and $q\le1/64$. Since our start does not supply $y^*(0)$, the method first computes a dual warm start: the same L-SVRE solver with an exact residual stopping test (Lemma~\ref{lem:lsvre-cert}), combined with Catalyst for strongly convex problems \citep{lin2018}, returns a point satisfying
\begin{equation}\label{eq:warm}
 \norm{y_{\rm w}-y^*(0)}^2\le D_0
\end{equation}
on every terminating run, at expected cost $O(nq^{-1/2}\log^2(2+\kappa))$.

The tolerances of the accelerated scheme of \citet{zhang2021} depend on the saddle gap of each outer subproblem, which the algorithm does not know. At an outer point $w_t$, let $\widehat f_t(p,y)=f(p,y)+\norm{p-p_t}^2$, let $G_t$ be its saddle gap, and let $R_t^2=\norm{\nabla \widehat f_t(w_t)}^2$, which costs $n$ queries to compute. If $R_t=0$, the point is stationary. Otherwise
\begin{equation}\label{eq:gapcert}
 B_t:=\frac{\nu R_t^2}{72}\le G_t,\qquad \frac{G_t}{B_t}\le\frac{36}{\nu^2}.
\end{equation}
Thus $B_t$ is a computable lower bound on $G_t$ that is off by at most the factor $36/\nu^2$. Replacing the unavailable gap by $B_t$ in the acceleration tolerances preserves the bounds of \citet[Theorems 4.1--4.3]{zhang2021} at the cost of a logarithmic factor, and the residual estimate also controls each auxiliary warm start directly. Each of the $O(1+S)$ outer solves costs $O(nq^{-1/2}\log^2(2+\kappa))$ in expectation, and $nq^{-1/2}=n^{3/4}\sqrt\kappa$. Appendix~\ref{app:upper} also counts the warm-start solves that start from $y=0$, the exact residual checks, and the conversion to fixed confidence, and Appendix~\ref{nu:sec:assembly} combines the two regimes.

\section{Related Work}
For finite-sum nonconvex--concave problems, the lower bounds of \citet[Theorems 3.1 and 3.3]{wu2026ncfinite} differ from ours in the dual geometry and the stationarity measure. As their Section~4 notes, substituting the effective condition number of a regularized dual into the NC--SC rate is only a formal comparison, and converting their Moreau envelope guarantee into a bound on the gradient of the value function can lose a factor of order $\kappa$. For finite-sum monotone inclusions under mean-squared Lipschitz continuity, \citet[Theorems 1 and 2]{zhou2026inclusion} gives an $O(n+\sqrt nLR/\eps)$ method and a matching $\Omega(n+\sqrt nLR/\eps)$ lower bound, where $R$ is an initial distance to a solution. That lower bound concerns randomized linear-span algorithms with component operator evaluations, and the resolvent there is that of the common operator, not of the components.

Our construction builds on established precedents for the path and the nonlinear source. \citet{li2021} use dual paths and endpoint compensation with the source functions of \citet{carmon2020}, and \citet{zhou2026ncsc} assigns internal dual edges to random component owners. \citet[Section 4.2 and Theorems 66 and 74]{han2024} develop component grouping and path bounds for convex finite sums. Our construction preserves the primal and dual scales while satisfying global smoothness and hiding the component assignments from arbitrary exact proximal queries.

Exact hiding also has distinct antecedents. \citet[Appendix C, Property 1 and Section C.3]{woodworth2016} use flat edge penalties to limit what proximal responses reveal, \citet[Section 5.2]{carmon2020} bound the effect of unrestricted query norms by radial compaction, and \citet[Sections 2.3--2.4]{bubeck2019} use a convex wall outside correlation cones. In randomized information arguments, \citet[Theorem 3 and Appendix B.1.2]{arjevani2023} condition on preceding projected queries. \citet{peng2026dense} develop dense weak hiding for individually smooth nonconvex and PL finite sums with exact values and gradients. Our construction combines blockwise exact transcript coupling with radius-dependent flat edges and componentwise strong concavity, and it controls the entire saddle-proximal response. In the related NC--C setting with a bounded dual domain, \citet[Theorem 4.1]{wu2026ncc} prove a lower bound for deterministic zero-respecting algorithms, under a different stationarity measure.

\section{Discussion}\label{sec:discussion}
Exact proximal access and unrestricted adaptivity do not change the polynomial complexity of the problem for any condition number. Although an exact component proximal step solves a saddle problem in each call, on this class it saves at most a factor $O(\log^2(2+\kappa))$ in worst-case component queries over component gradients, for $\eps$ in the range of Theorem~\ref{ak:thm:all}. Two parts of our upper bounds apply to other finite-sum minimax methods: the shared L-SVRE solver needs strong concavity only of the mean, not of each component, and the residual estimate $B_t$ of Section~\ref{sec:upper} makes the subproblem tolerances of the accelerated scheme of \citet{zhang2021} computable.

Our lower bounds use hard instances whose dimension grows with the parameters. As in the lower bound of \citet{zhou2026ncsc}, our lower bound requires $\eps^2/(L\Delta)$ to be below a universal constant, and we have not optimized this constant or the other universal constants. Under~\eqref{ak:eq:all-window}, for $\kappa\le2^{18}\sqrt n$ and $S\ge\log(2\kappa)$, Theorem~\ref{nu:thm:normalized} gives $O(n\log(2\kappa)+h(n,\kappa)S)$, so the upper and lower bounds match in the accuracy term and differ by at most a factor $\log(2\kappa)$ in the term $n$. For larger $\kappa$, the accelerated method of Appendix~\ref{app:upper} leaves a gap of a factor $\log^2(2+\kappa)$.

Two questions remain open. Finite sums with individually $L$-smooth components form a smaller class than those with joint mean-squared smoothness $L$. On a bilinear subclass, \citet[Theorem 4]{zhou2026ncsc} proves the lower bound $\Omega(n+\min\{\kappa,\sqrt{n\kappa}\}L\Delta/\eps^2)$ for zero-respecting methods; what is the complexity of the individually smooth class for unrestricted algorithms? For nonconvex--concave finite sums over bounded dual domains, \citet{wu2026ncfinite} determine the complexity with exact values and gradients up to a single logarithmic factor; does exact component proximal access change that complexity?

\par\textit{Formal verification.}
We provide a Lean~4 formalization of the lower-bound assertions of Theorem~\ref{ak:thm:all} as supplementary material and as ancillary files of the arXiv version. It covers the expected query cost and the fixed-confidence complexity under a deterministic cap, in all three ranges of the condition number and for the baseline instance that gives the term $n$. The formal proof includes the path calibration of Lemma~\ref{rb:lem:green}, the posterior owner cost of Lemma~\ref{rp:lem:posterior-cost}, and the public cosine terms in~\eqref{rp:eq:public-component} and~\eqref{ak:eq:primal-components} together with their exact legal boundary. In place of the Haar-distributed axes, it uses a finite Rademacher construction, and it obtains a barrier constant that plays the role of the constant of Lemma~\ref{rb:match:lem:uniform-barrier} by the same compactness argument. The files \texttt{CLAIMS} and \texttt{TRUSTED\_DEFINITIONS} of the formalization state the formal theorems, the dimension bounds, and the remaining differences from the written proof. The Haar conditioning argument, the zero-axis extension, the equality of the two barrier constants, Corollary~\ref{cor:han}, the upper bounds, and the dimension estimate~\eqref{eq:dimension-main} are not formalized.

\appendix
\section{The Hard Family and Its Global Geometry}\label{app:robust}
We use one path family for every $\kappa\ge64$: long paths prove
Theorem~\ref{rb:thm:main}, and two-node paths cover the intermediate
range in Appendix~\ref{ak:app:all}. The scalar source functions are those of
\citet{carmon2020}, and the dual paths with endpoint compensation follow \citet{li2021}. Symbols defined in Appendices~\ref{app:robust}--\ref{ak:app:all} are local to the lower-bound proofs.

Let $P_m$ be the Laplacian of the $m$-vertex path, with diagonal
$(1,2,\ldots,2,1)$. Fix
\begin{align}
 a&=\frac{L}{2^{16}\sqrt n},& t&=\sqrt{a/\mu},&
 m&=\max\{2,\lfloor t\rfloor\},\nonumber\\
 r&=\begin{cases}\mu/a,&t<2,\\m^{-2},&t\ge2,\end{cases}&
 \mu_0&=ar,& Q&=P_m+rI.\label{rb:eq:base-parameters}
\end{align}
Write $u=(Q^{-1})_{11}$, $v=(Q^{-1})_{1m}$ and $g=u/v$, and put
\begin{equation}\label{rb:eq:green-parameters}
 \delta_0=2^{-39},\quad \lambda=\frac{\delta_0L}{g},\quad
 \nu=\frac au,\quad c=\nu-\mu_0,\quad
 b^2=\frac{12a\lambda}{v},\quad d=6\lambda(g-1).
\end{equation}
Instantiate the path family only when
\begin{equation}\label{ak:eq:path-window}
 X:=\frac{c_\star^2\lambda\Delta}{2048\eps^2}\ge2,
 \quad\text{equivalently}\quad
 \eps^2\le c_\star^2\lambda\Delta/4096.
\end{equation}
Outside this window the indexed baseline will suffice. Within it set
\begin{align}
 s&=\frac{8\eps}{c_\star\lambda},& M&=\lfloor X\rfloor,&
 K&=Mm,&E&=M(m-1).\label{rb:eq:macro-parameters}
\end{align}
Thus $m,M\ge2$ and
\begin{equation}\label{ak:eq:macro-floor}
 M\ge\frac{c_\star^2\lambda\Delta}{4096\eps^2},\qquad
 M\lambda s^2\le\Delta/32,\qquad \lambda s^2\le\Delta/64.
\end{equation}
The width and hiding dimension are fixed by
\eqref{rb:eq:positive-width} and \eqref{rp:eq:dimension-sufficient}.
With $A_1=2433$, the bound on $\norm{D\mathcal A}$ in Lemma~\ref{rb:match:lem:forcing}, and $C_0=294912A_1/c_\star$, their ratio is
\begin{equation}\label{rb:eq:dimension-ratio}
 \frac{R_*}{\eta}=\max\left\{C_0\frac gr\sqrt K,16\sqrt n\right\},
 \qquad \frac{R_*}{\eta_0}=C_0\frac gr\sqrt K>128.
\end{equation}
In particular $H_n=\max\{C_0^2(g/r)^2K,256n\}$, and the dimension
is fixed before execution, independently of the stopping time.

\subsection{Path Geometry, Hidden Axes, and the Scalar Components}\label{rb:sec:path}
There are $M$ publicly identified dual paths. Every node
$y_{j,k}\in\R^D$ has an independent Haar-uniform hidden unit axis $U_{j,k}$.
The primal coordinate block $P_j=(X_j,Z_{j+1})\in\R^D\times\R^D$ is publicly
identified with the two endpoint coordinate blocks $(j,1)$ and $(j,m)$, respectively.
Thus its source and outgoing connector axes are $U_{j,1}$ and $U_{j,m}$.
Interior axes are independent of all endpoint axes. For each edge
$(j,k)$, $k<m$, draw a fixed independent owner $\Sigma_{j,k}$ uniformly
from $[n]$. These draws and the axes are independent of the algorithm's
private seed; they will be fixed before invoking its oracle.

\begin{lemma}[Common path bounds and branch calibration]\label{rb:lem:green}
In both branches,
\begin{equation}\label{rb:eq:common-path-bounds}
 \begin{gathered}
 \mu_0\ge\mu,\quad 0\le c\le a,\quad \nu\le\mu+a,\quad 0<v\le u,\\
 \frac{b^2}{\nu}=12\delta_0L,\qquad
 0\le d=6\delta_0L(1-1/g)\le6\delta_0L.
 \end{gathered}
\end{equation}
Every entry in either endpoint inverse column is at most $u$.
For $t\ge2$, that is, $\kappa\ge2^{18}\sqrt n$,
\begin{equation}\label{rb:eq:green-bounds}
 m\le u\le m^2,\quad \frac m{\sinh1}\le v\le m,\quad
 1<g\le\cosh1<5/3.
\end{equation}
Then $\mu\le\mu_0\le4\mu$, $\nu,c\le a/2$, $d<4\lambda$ and
$\lambda\ge2^{-40}L$. Hence the window~\eqref{rb:eq:window} of Theorem~\ref{rb:thm:main}
implies~\eqref{ak:eq:path-window} and
\begin{equation}\label{rb:eq:floor-bounds}
 m\ge\frac{\sqrt\kappa}{512n^{1/4}},\qquad
 M\ge\frac{c_\star^2S}{2^{52}}.
\end{equation}
For $t<2$, $m=2$, $r=\mu/a>1/4$, $\mu_0=\mu$, and
\begin{equation}\label{ak:eq:middle-green}
 Q^{-1}=\frac1{r(r+2)}
 \begin{pmatrix}r+1&1\\1&r+1\end{pmatrix},\quad
 g=1+r,\quad \nu=\mu+\frac{ar}{r+1},\quad c=\frac{ar}{r+1}.
\end{equation}
The two calibrations agree at $t=2$.
\end{lemma}
\begin{proof}
The two-node formula is direct. For the long path and any mass $r>0$,
put $\theta=\operatorname{arcosh}(1+r/2)$. The leading-minor recurrence
$d_j=(2+r)d_{j-1}-d_{j-2}$, with first endpoint entry $1+r$,
gives $d_j=\cosh((j+1/2)\theta)/\cosh(\theta/2)$ before the last
endpoint, and $\det(P_m+rI)=r\sinh(m\theta)/\sinh\theta$.
The endpoint cofactors therefore give
\[
 v=\frac{\sinh\theta}{r\sinh(m\theta)},\qquad
 g=\frac{\cosh((m-1/2)\theta)}{\cosh(\theta/2)}.
\]
For $r=m^{-2}$, $\theta=2\operatorname{asinh}(1/(2m))$,
$m\theta\le1$, $\sinh\theta\ge1/m$ and
$\sinh(m\theta)\ge m\sinh\theta$, proving the bounds on $v,g$.
The constant eigenvector contributes $m$ to each diagonal entry
of $Q^{-1}$, and $\norm{Q^{-1}}=m^2$, giving the bounds on $u$.
The endpoint column maximum principle gives the entrywise claim in
both branches. The remaining bounds follow by substitution and
$\lfloor t\rfloor\ge t/2$ for $t\ge2$.
\end{proof}

For the high branch, $r=m^{-2}$ and $K=Mm$, so \eqref{rb:eq:dimension-ratio} specializes to
\begin{align*}
 H_n&=\max\{C_0^2g^2Mm^5,256n\},\\
 D&=O\!\left((n+S\kappa^{5/2}n^{-5/4})\log(n\kappa S)\right).
\end{align*}
Indeed $M=O(S)$, $m=O(\sqrt\kappa n^{-1/4})$ and $g\le\cosh1$.
The primal dimension is $2MD+1$, including the public coordinate
in~\eqref{rp:eq:public-component}, and the dual dimension is $KD$, so the total dimension $(2M+K)D+1$ is~\eqref{eq:dimension-main}.

The source uses the scalar functions of \citet{carmon2020} and \citet{li2021}:
\begin{equation}\label{rb:match:eq:scalars}
 \Psi(t)=\begin{cases}0,&t\le1/2,\\e^{1-(2t-1)^{-2}},&t>1/2,\end{cases}
 \qquad \Phi(t)=\sqrt e\int_{-\infty}^t e^{-r^2/2}\,\mathrm dr.
\end{equation}
The cap and primal penalty are
\begin{align}
 w_j=\rho(P_j)&=P_j/(1+\norm{P_j}^2/R^2)^{1/2},& R&=128s,
 \label{rb:match:eq:cap}\\
 p(P_j)&=\frac\lambda{1024}(\norm{P_j}-R)_+^2,&
 x_j&=\ip{U_{j,1}}{w_{j,X}},\quad z_{j+1}=\ip{U_{j,m}}{w_{j,Z}}.
 \label{rb:eq:primal-coordinates}
\end{align}
Write $w_{j,X}=x_jU_{j,1}+A_{j,\perp}$ and
$w_{j,Z}=z_{j+1}U_{j,m}+B_{j+1,\perp}$. Let
\begin{equation}\label{rb:match:eq:source}
 J_M=-\Phi(x_1/s)+\sum_{j=2}^M
 [\Psi(-z_j/s)\Phi(-x_j/s)-\Psi(z_j/s)\Phi(x_j/s)].
\end{equation}
For $1/2<t<1$ define
\begin{equation}\label{rb:match:eq:gate}
\begin{aligned}
 w(t)&=\sqrt{\Psi(t)},& \omega(w)&=1-(1-w)^3,\\
 \theta_+(t)&=\tfrac\pi2\omega(w(t)),&
 a_+(t)&=\sin\theta_+(t),\quad b_+(t)=\cos\theta_+(t).
\end{aligned}
\end{equation}
extending $\theta_+$ by $0$ for $t\le1/2$ and by $\pi/2$ for $t\ge1$.
Set
\begin{align}
 \alpha&=a_+^2,&q_j&=\prod_{k=2}^j\alpha(z_k/s),\quad q_1=1,
 \label{rb:match:eq:masses}\\
 h_j&=\Bigl(\prod_{k=2}^j a_+(z_k/s)\Bigr)b_+(z_{j+1}/s),&
 \mathcal A_j&=h_j\Gamma w_j,\qquad \Gamma=\operatorname{diag}(I,-I/2).
 \label{rb:match:eq:forcing}
\end{align}
The concatenated endpoint forcing obeys
\begin{equation*}
 h_j^2=q_j[1-\alpha(z_{j+1}/s)],\qquad \sum_j h_j^2\le1,
 \qquad \norm{\mathcal A}\le R.
\end{equation*}
Define the common primal part
\begin{equation}\label{rb:eq:primal-part}
 \mathcal P(P)=\lambda s^2J_M(P)-d\norm{\mathcal A(P)}^2+\sum_jp(P_j).
\end{equation}
Let $t_{j,k}=\ip{U_{j,k}}{y_{j,k}}$,
$\phi_\eta(t)=\frac12(|t|-\eta)_+^2$, and write $Y_\partial$ for the
concatenation of all endpoint coordinate blocks, ordered like $\mathcal A$.
For an edge $e=((j,k),(j,k+1))$, write $Y_e=(y_{j,k},y_{j,k+1})$ and
$t_e=t_{j,k}-t_{j,k+1}$. For a node $v$ write $y_v,t_v$.

The cutoff functions are
\begin{align}
 \chi(t)&=\begin{cases}1&t\le1,\\1-10r^3+15r^4-6r^5&t=1+r,\ 0<r<1,\\0&t\ge2,\end{cases}
 \label{rb:eq:cutoff}\\
 \psi(t)&=\begin{cases}0&t\le1,\\(t-1)^3/6&1<t<2,\\
 (t-2)^2/2+(t-2)/2+1/6&t\ge2.
 \end{cases}\label{rb:eq:dual-radial}
\end{align}
We localize the cutoff $\chi$ and the dual radial profile $\psi$ at the radius $R_*$ by setting
\begin{equation}\label{rb:eq:localized-radii}
 R_*=16bR/\nu,\qquad
 \chi_*(z)=\chi(\norm z/R_*),\qquad
 \psi_*(z)=R_*^2\psi(\norm z/R_*).
\end{equation}
For the owned edges use the profile $\varrho$ in~\eqref{eq:profile-main} and define
\begin{equation}\label{sc:eq:radial-edge}
 h_\eta(r)=\eta\varrho(r/R_*),\qquad
 A_\eta(Y_e)=\tfrac12\bigl(|t_e|-h_\eta(\norm{Y_e})\bigr)_+^2,
 \qquad B_T(z)=\tfrac12(\norm z-T)_+^2.
\end{equation}
Node and endpoint cutoffs keep radius $R_*$; we use the positive width
\eqref{rb:eq:positive-width} and width zero.

The globally defined components are precisely
\begin{align}
 f_{i,\eta}(P,y)={}&\mathcal P(P)-\frac\nu2\norm y^2
 +b\chi_*(Y_\partial)\ip{\mathcal A(P)}{Y_\partial}
 -45\nu\psi_*(Y_\partial)\nonumber\\
 &+\sum_v[c\chi_*(y_v)\phi_\eta(t_v)-720c\psi_*(y_v)]\nonumber\\
 &-na\sum_{e:\Sigma_e=i}A_\eta(Y_e)
 -a\sqrt n\sum_e B_{R_*/2}(Y_e).\label{rb:eq:components}
\end{align}
Only the edge sum with a flat region depends on owners; every radial correction
uses public norms and is common to all components.

The radii are nested as follows. Since $\norm{\mathcal A}\le R$, every node, edge pair and endpoint tuple of the uncut dual maximizer has norm at most $2bR/\nu=R_*/8$ (Lemma~\ref{rb:lem:uncut}), and after the width perturbation these stay inside $3R_*/16$ (Lemma~\ref{rb:lem:width-start}). There the cutoffs equal one and the barriers $\psi_*$ and $B_{R_*/2}$ vanish, so localization does not move the maximizer. The barriers act only beyond $R_*/2$, where $B_{R_*/2}$ dominates the curvature of the flat edges (Lemma~\ref{rb:lem:radial}).

\subsection{Forcing Derivatives}
\begin{lemma}[Gate and forcing estimates]\label{rb:match:lem:forcing}
The nonnegative amplitudes $a_+,b_+$ are $C^2$, flat to second order
at both joins, and satisfy $a_+^2+b_+^2=1$. With $C=23$,
\begin{align}
 \Psi(t)&\le\alpha(t)\le C\Psi(t),&&1/2<t<1,
 \label{rb:match:eq:gate-comparison}\\
 \alpha(t)/C&\le\Psi(t)\le e\alpha(t),&&t\ge0,
 \label{rb:match:eq:source-comparison}\\
 0&\le\alpha'(t)\le C\Psi'(t),&&1/2<t<1.
 \label{rb:match:eq:gate-derivative}
\end{align}
Outside the transition $\alpha'=0$. Uniformly in length, dimension,
owners, signs and raw radii,
\begin{equation}\label{rb:match:eq:forcing-bounds}
 \norm{D\mathcal A}\le2433,\qquad
 \norm{D^2\mathcal A}<231000/s,
\end{equation}
where the second norm is bilinear-to-vector in Euclidean norms.
\end{lemma}
\begin{proof}
\textit{Gate.} At onset $w=e^{1/2-[2(2t-1)^2]^{-1}}$ is infinitely flat; at completion
$1-\omega(w)=(1-w)^3$, so $b_+=O((1-t)^3)$ and $1-a_+=O((1-t)^6)$.
For $0\le w\le1$, $w\le \omega(w)\le3w$ and
$\omega(w)\le\sin(\pi \omega(w)/2)\le\pi \omega(w)/2$. Hence the first two
comparisons hold, using $9\pi^2/4<C$, $\alpha=1$ and
$1\le\Psi<e$ on $t\ge1$. Differentiating gives
$\alpha'/\Psi'=\pi \omega'(w)\sin\theta_+\cos\theta_+/(2w)<C$.

\par\textit{Angle and radial maps.} For $v=(2t-1)^{-2}\ge1$,
$w'=2e^{(1-v)/2}v^{3/2}$ and $w''=4e^{(1-v)/2}v^2(v-3)$.
Maximizing $v^ke^{-v/2}$ gives $|w'|<4$, $|w''|<120$ and thus
$|\theta_+'|<19$, $|\theta_+''|<720$.
Writing $c=(1+\norm P^2/R^2)^{-1/2}$ gives
$D\rho=cI-c^3PP^\top/R^2$, with eigenvalues $c,c^3$.
The three second-derivative terms linear in $P$ total at most
$3r(1+r^2)^{-3/2}/R$, and the cubic term is at most
$3r^3(1+r^2)^{-5/2}/R$, where $r=\norm P/R$.
Each scalar factor is at most $1/2$, so
$\norm{D\rho}\le1$, $\norm{D^2\rho}\le3/R$. The disjoint-block
angle map $\Theta_j(P)=\theta_+(z_{j+1}/s)$ therefore obeys
\begin{equation*}
 \norm{D\Theta}<19/s,\qquad \norm{D^2\Theta}<721/s^2.
\end{equation*}
For the second bound use
$\sum_j\norm{U_j}^2\norm{W_j}^2\le\norm U^2\norm W^2$.

\par\textit{Spherical coordinates.} To avoid a length factor, put $p_0=1$, $p_k=\prod_{i\le k}\sin\vartheta_i$
and
\[
 H(\vartheta)=(\cos\vartheta_1,p_1\cos\vartheta_2,\ldots,
 p_{M-1}\cos\vartheta_M,p_M).
\]
Let $T_k$ be the unit tail starting with $\cos\vartheta_k$, padded
by zeros, and $Q_k=\partial_kT_k$. The $Q_k$ are orthonormal,
since earlier derivatives are multiples of a later tail and tangents
are orthogonal to that tail. Moreover,
\[
 \partial_kH=p_{k-1}Q_k,\qquad
 \norm{\nabla p_{k-1}}^2=
 \sum_{i<k}\cos^2\vartheta_i\prod_{r<k,r\ne i}\sin^2\vartheta_r\le1.
\]
The last inequality follows by expanding
$\prod_{i<k}(\sin^2\vartheta_i+\cos^2\vartheta_i)=1$. Thus
$\norm{DH}\le1$, and
\begin{align*}
 D^2H[U,W]={}&-\sum_kU_kW_kp_{k-1}T_k\\
 &+\sum_k[W_k\ip{\nabla p_{k-1}}{U_{<k}}
          +U_k\ip{\nabla p_{k-1}}{W_{<k}}]Q_k
\end{align*}
has norm at most $3\norm U\norm W$, by Cauchy--Schwarz in the first sum and orthogonality in each of the other two.

\par\textit{Forcing map.} Since $h$ contains the
first $M$ coordinates of $H(\Theta)$,
$\norm{Dh}\le19/s$ and $\norm{D^2h}<1804/s^2$. Differentiating
$\mathcal A_j=h_j\Gamma\rho(P_j)$ in orthogonal output blocks gives
\[
 \norm{D\mathcal A}\le1+R\norm{Dh}\le2433,\qquad
 \norm{D^2\mathcal A}\le3/R+2\norm{Dh}+R\norm{D^2h}<231000/s.
\]
\end{proof}

\Needspace{10\baselineskip}
\subsection{Class Membership and Exact Zero-Width Elimination}
\begin{lemma}[Uncut energy and Schur identity]\label{rb:lem:uncut}
Define
\begin{equation}\label{rb:eq:energy}
 \mathcal E_\eta(y)=\frac\nu2\norm y^2-c\sum_v\phi_\eta(t_v)
                         +a\sum_e\phi_\eta(t_e).
\end{equation}
It is $\mu_0$-strongly convex. At $\eta=0$, its hidden-axis Hessian on
every path is $aQ$, and its complementary Hessian is $\nu I$.
Maximizing $\mathcal P(P)+b\ip{\mathcal A(P)}{Y_\partial}-\mathcal E_0(y)$
gives exactly
\begin{equation}\label{rb:eq:zero-value}
 V_0(P)=\lambda s^2J_M(P)+\sum_jh_j^2Q_j(w_j)+\sum_jp(P_j),
\end{equation}
where
\begin{equation}\label{rb:eq:eliminated-quadratic}
 Q_j(w)=6\lambda(x_j-z_{j+1}/2)^2
        +6\lambda\norm{A_{j,\perp}}^2
        +\tfrac32\lambda\norm{B_{j+1,\perp}}^2.
\end{equation}
Its unique uncut maximizer obeys
\begin{equation}\label{rb:eq:optimizer-local-bounds}
 \norm{Y_{\partial,0}}\le2bR/\nu,\qquad
 \norm{y_{v,0}}\le2bR/\nu,\qquad
 \norm{Y_{e,0}}\le2bR/\nu.
\end{equation}
\end{lemma}
\begin{proof}
Let $U$ map the vector of hidden scalar coordinates to their orthogonal
ambient node axes, and let $D_U$ be the incidence map on those scalars;
$\norm{D_U}\le2$. Almost everywhere,
\[
 \nabla^2\mathcal E_\eta
 =\nu I-cU\operatorname{diag}(\mathrm{active\ nodes})U^\top
   +aD_U^\top\operatorname{diag}(\mathrm{active\ edges})D_U
 \succeq(\nu-c)I=\mu_0I.
\]
The gradient is continuous at the hinges, so line integration gives
strong convexity everywhere. At zero width the hidden Hessian is
$\mu_0I+aP_m=aQ$, while the orthogonal complements have Hessian $\nu I$.
For one path its eliminated contribution before compensation is
\[
 \frac{b^2h_j^2}{2a}(u x_j^2-vx_jz_{j+1}+u z_{j+1}^2/4)
 +\frac{b^2h_j^2}{2\nu}
      (\norm{A_{j,\perp}}^2+\norm{B_{j+1,\perp}}^2/4).
\]
Both endpoint diagonal coefficients after subtracting
$d h_j^2(\norm{w_{j,X}}^2+\norm{w_{j,Z}}^2/4)$ are fixed by
$b^2u/(2a)-d=b^2/(2\nu)-d=6\lambda.$
This proves the identity.

The hidden endpoint Green matrix has norm $u+v\le2u$ and the
complementary inverse is $1/\nu$. Summing over the orthogonal bridges
and using $\sum_j\norm{\mathcal A_j}^2\le R^2$ proves the first
optimizer bound. Any two interior rows, such as the two nodes of an edge, restricted to the two source
columns form a $2\times2$ matrix with entries at most $u$, hence norm
at most $2u$. The complementary sources occur only at endpoints, where
the inverse has norm $1/\nu$. Taking the maximum of these orthogonal
operators proves both local bounds.
\end{proof}

\begin{lemma}[Radial calculus]\label{rb:lem:radial}
The following bounds hold globally, where Hessians are understood almost everywhere.
\begin{enumerate}[label=(\roman*),leftmargin=*]
\item If a $C^{1,1}$ scalar $A$ satisfies
$|A|/R_*^2,\norm{\nabla A}/R_*\le4A_0$ on $R_*<\norm z<2R_*$, then
\[
 \nabla^2(\chi_*A)-\chi_*\nabla^2A\preceq720A_0\nabla^2\psi_*.
\]
\item For every $\eta\ge0$, the radial threshold $h_\eta(\norm z)$ is convex and globally
$C^{1,1}$, constant on $\norm z\le R_*$, and satisfies
$\norm{\nabla h_\eta}\le\beta:=\eta/R_*$ and
$h_\eta(r)\ge\max\{\eta,\eta r/(2R_*)\}$. For $r=\norm z>0$, $u=z/r$,
\begin{equation}\label{rd:eq:radial-hess}
 \nabla^2h_\eta=
 \frac{\eta}{R_*^2}\varrho''(r/R_*)uu^\top
 +\frac{\eta}{R_*r}\varrho'(r/R_*)(I-uu^\top)\succeq0.
\end{equation}
Here $0\le\varrho'\le1$ and $\varrho''=\one_{(1,2)}$ almost everywhere.
\item If $n\ge2$, $0\le\eta\le R_*/(16\sqrt n)$ and a vector $e_U$ satisfies
$\norm{e_U}\le\sqrt2$, the owned edge
$G(z)=-\frac{na}{2}(|\ip{e_U}z|-h_\eta(\norm z))_+^2$ is globally $C^{1,1}$, with
\begin{equation}\label{rd:eq:edge-lip}
 \norm{\nabla^2G}\le\tfrac94na,
 \qquad G-a\sqrt n B_{R_*/2}\text{ globally concave}.
\end{equation}
\end{enumerate}
\end{lemma}
\begin{proof}
On $\norm z/R_*=1+r$, $0<r<1$, one has
$|\chi'|\le30r^2$, $|\chi''|\le60r$. In direction $te_r+w$, $w\perp e_r$,
the two extra product rule terms are bounded above by
\[
 A_0(480rt^2+240r^2|t|\norm w+120r^2\norm w^2)
 \le720A_0(rt^2+r^2\norm w^2/4).
\]
The difference is positive semidefinite, since its matrix has diagonals
$240r,60r^2$ and off-diagonal $-120r^2$, with determinant
$14400r^3(1-r)\ge0$. The radial and tangential eigenvalues of
$\nabla^2\psi_*$ are $r$ and $r^2/[2(1+r)]\ge r^2/4$.
Outside this annulus the cutoff is constant, proving (i).

The derivative of $\varrho$ is $0$, $t-1$, and $1$ on its three intervals,
continuous and $1$-Lipschitz. The radial Hessian formula gives (ii),
including regularity at the origin since the threshold is locally constant.
The lower bound uses $\varrho\ge1\ge t/2$ on $[0,2]$ and
$\varrho(t)=t-1/2\ge t/2$ for every $t\ge2$.

For (iii), write $r=\norm z$ and, on the active set, put
$q=|\ip{e_U}z|-h_\eta(r)>0$ and
$v=\operatorname{sign}(\ip{e_U}z)e_U-\nabla h_\eta$. Then
\begin{equation*}
 \nabla G=-naqv,\qquad \nabla^2G=-na\,vv^\top+na\,q\nabla^2h_\eta.
\end{equation*}
For $\eta>0$, away from $\ip{e_U}z=0$ this is the locally $C^{1,1}$
composition $-na(q_+)^2/2$; the gradient vanishes continuously at activation.
Near $\ip{e_U}z=0$ the function is locally zero.
For $\eta=0$, $G=-na\ip{e_U}z^2/2$ directly.
Since $q\le\sqrt2r$, the radial and tangential eigenvalues of
$q\nabla^2h_\eta$ are at most $2\sqrt2\beta$ and $\sqrt2\beta$, respectively;
the radial term is supported on $R_*<r<2R_*$. Thus
\begin{equation}\label{rd:eq:error}
 0\preceq naq\nabla^2h_\eta\preceq2\sqrt2na\beta I
 \preceq(\sqrt2/8)a\sqrt n I,
\end{equation}
vanishing on $r<R_*$. Also, since $\beta\le1/(16\sqrt n)$ and $n\ge2$,
$\norm v^2\le(\sqrt2+\beta)^2\le(\sqrt2+1/(16\sqrt2))^2=1089/512<9/4$ and
$2\sqrt2\beta\le1/8$. The norm of the difference of two positive
semidefinite matrices is at most the larger norm, proving the edge bound.
For $r>R_*/2$,
$\nabla^2B_{R_*/2}=uu^\top+(1-R_*/(2r))(I-uu^\top)$;
it is nonnegative everywhere and at least $I/2$ on $r\ge R_*$,
which dominates~\eqref{rd:eq:error} after multiplication by $a\sqrt n$.
All gradients match at the profile, cutoff and activation joins.
Bounded weak Hessians, mollification and line integration therefore give
the stated global $C^{1,1}$ and curvature claims.
\end{proof}

\begin{lemma}[Componentwise strong concavity and RMS smoothness]\label{rb:lem:class}
The components~\eqref{rb:eq:components} are globally $C^{1,1}$, each component
is $\mu_0$-strongly concave in $y$, and their global joint RMS smoothness
constant is at most
\begin{align}
 L_{\rm rob}\le{}&601\lambda+10^8d
 +32\cdot128\cdot231000\,b^2/\nu+13000b\nonumber\\
 &+\tfrac{13}{2}a\sqrt n+741c+47\nu<4L/5.
 \label{rb:eq:robust-rms}
\end{align}
The pure primal Hessian also obeys, almost everywhere,
\begin{equation}\label{rp:eq:primal-curvature}
 \norm{\nabla^2_{PP}f_{i,\eta}}\le K_P
 :=601\lambda+10^8d+32\cdot128\cdot231000\,b^2/\nu<L/32.
\end{equation}
These bounds hold for every owner assignment, all dimensions and
$0\le\eta\le R_*/(16\sqrt n)$, including when any subset of axes is
replaced by zero consistently in its primal and dual occurrences.
\end{lemma}
\begin{proof}
\textit{Strong concavity.} Lemma~\ref{rb:lem:radial}(iii) makes each owned edge plus its shared
correction concave; unowned corrections are concave. Apply part (i) to
$(A,A_0)=(c\phi_\eta(\ip U{y_v}),c)$ at each node and
$(b\ip{\mathcal A(P)}{Y_\partial},bR/R_*)$ at the endpoints.
Since $bR/R_*=\nu/16$, the barriers remove the extra curvature and leave
$\nabla^2_{yy}f_{i,\eta}\preceq-\nu I+cUU^\top\preceq-\mu_0I$.

\par\textit{Common terms.} For the norm estimates, put $A_1=2433$, $A_2=231000$,
$C_1=15/8$ and $C_2=10/\sqrt3$. Direct differentiation gives
$\norm{\nabla\chi_*}\le C_1/R_*$,
$\norm{\nabla^2\chi_*}\le C_2/R_*^2$ and $\norm{\nabla^2\psi_*}\le1$.
Also $0\le\Psi<3$, $|\Psi'|<5$, $|\Psi''|<100$ and
$0<\Phi<5$, $|\Phi'|<2$, $|\Phi''|\le1$.
In particular, for $v=(2t-1)^{-2}$,
$\Psi''=8e^{1-v}v^2(2v-3)$ gives
$|\Psi''|\le432/e^2+96/e<100$ by maximizing powers times exponentials.
The source-pair Hessian entries are bounded by $500,3,10$, hence its
norm is below $510$; composition with the caps adds less than
$(3/128)\sqrt{6^2+25^2}<1$ and the penalty adds $\lambda/512$.
Together with Lemma~\ref{rb:match:lem:forcing}, this gives the following
bounds for the common terms (the endpoint forcing includes its cutoff):
\begin{center}
\renewcommand{\arraystretch}{1.3}
\begin{tabular}{@{}lll@{}}
\toprule
Term & Hessian block & Bound\\
\midrule
$\lambda s^2J_M+\sum_jp(P_j)$ & $PP$ & $601\lambda$\\
\midrule
$-d\norm{\mathcal A}^2$ & $PP$ & $2d(A_1^2+128A_2)<10^8d$\\
\midrule
$b\chi_*\ip{\mathcal A}{Y_\partial}$ & $PP$ & $2bR_*A_2/s=32\cdot128A_2b^2/\nu$\\
 & $Py$ & $bA_1(1+2C_1)<13000b$\\
 & $yy$ & $(bR/R_*)(2C_1+2C_2)<\nu$\\
\midrule
$\sum_v[c\chi_*\phi_\eta-720c\psi_*]$ & $yy$ & $c(1+4C_1+2C_2)+720c<741c$\\
\midrule
$-\nu\norm y^2/2-45\nu\psi_*(Y_\partial)$ & $yy$ & $46\nu$\\
\bottomrule
\end{tabular}
\end{center}
On node cutoff supports, $|\phi_\eta|\le2R_*^2$ and
$\norm{\nabla\phi_\eta}\le2R_*$; node blocks are disjoint.
The effective endpoint vector has norm at most $2R_*$. The symmetric mixed block has norm equal to
that of its rectangular $Py$ block.

\par\textit{Owned edges.} Write $H_e$ for the unamplified pair Hessian,
$\norm{H_e}\le C_a=9a/4$. Degree two and blockwise Cauchy--Schwarz give
\[
 \sum_i\norm{\sum_{e:\Sigma_e=i}H_e w_e}^2
 \le2\sum_e\norm{H_e w_e}^2\le4C_a^2\norm w^2.
\]
Amplification by $n$ and RMS averaging cost $2\sqrt nC_a$;
the shared pair barriers cost $2a\sqrt n$ by the same degree bound.
Their sum is $(13/2)a\sqrt n$, uniformly over owner assignments and lengths.
Minkowski's inequality in the component-indexed Hilbert space now proves
\eqref{rb:eq:robust-rms}; the three $PP$ rows give~\eqref{rp:eq:primal-curvature}.
All terms are locally $C^{1,1}$. Mollification preserves both component
curvature and the stacked-gradient bound for
$n^{-1/2}(\nabla f_1,\ldots,\nabla f_n)$; line integration and passage to
the limit give the global claims across every hinge and join.

\par\textit{Numerical constants.} Equation~\eqref{rb:eq:common-path-bounds} and $\lambda\le\delta_0L$ bound $\lambda$, $d/6$ and $b^2/(12\nu)$ by $\delta_0L$, so
$K_P/L\le(601+6\cdot10^8+12\cdot32\cdot128\cdot231000)\,\delta_0<2^{34}\delta_0=1/32$. Since $\kappa\ge64$ and
$\nu/L\le1/64+2^{-16}<1/32$, the joint bound is
\[
 \frac{L_{\rm rob}}L\le\frac1{32}
 +13000\sqrt{12\delta_0/32}
 +\frac{47}{64}+\frac{788+13/2}{2^{16}}
 <0.79<\frac45.
\]
This includes the public dual ridge and endpoint correction $47\nu$.

\par\textit{Zeroed axes.} Every estimate used only axis norms at most one; zeroing axes preserves
the source, forcing and derivative bounds, while node curvature remains
at most $cI$ and edge vectors have norm at most $\sqrt2$.
Retain all public radial corrections and the full ambient endpoint
forcing. The same curvature and RMS bounds hold for these zero-axis
formulas, used as censored objectives in Appendix~\ref{app:oracle}.
\end{proof}

\section{The Value Function and Stationarity Barrier}\label{app:barrier}
We first bound the gradient of the zero-width value function $V_0$ near the end of the stage chain (Lemma~\ref{rb:match:lem:uniform-barrier}). A positive width then preserves this barrier, keeps the optimizer inside radius $3R_*/16$, and gives the initial budgets (Lemma~\ref{rb:lem:width-start}).
\subsection{A Uniform Terminal Gradient Bound}\label{rb:match:subsec:barrier}

\begin{lemma}[Uniform terminal-band barrier]\label{rb:match:lem:uniform-barrier}
There is a universal $c_\star>0$, depending only on the displayed scalar
functions and normalized radius $128$, such that, for every finite
length, dimension, owner assignment, and finite uncompressed primal point,
\begin{equation}\label{rb:match:eq:uniform-barrier}
 |z_{M+1}|\le s/8\quad\Longrightarrow\quad
 \norm{\nabla V_0(P)}\ge c_\star\lambda s.
\end{equation}
The gradient is in the uncompressed primal coordinates, with arbitrary
radii, signs, and orthogonal complements of the axes.
\end{lemma}
\begin{proof}
\par\textit{Outline.} After normalization the value is a chain in which block $j$ carries a mass $q_j\in[0,1]$ that can only decrease along the chain. We show that at an approximately stationary point the mass never falls below a fixed level $\ell$, and that a block of mass at least $\ell$ cannot be approximately stationary when its last coordinate lies in the terminal band. Step~3 proves both facts for one or two blocks with exactly vanishing residuals; compactness makes them quantitative (Step~4), and induction along the chain concludes (Step~5).

\par\textit{Step 1: residuals in compressed coordinates.}
Normalize lengths by $s$ and values by $\lambda s^2$, keeping the same symbols for the normalized quantities, and write
$R_0=128$. The cap is a diffeomorphism onto the open radius-$R_0$
ball. In compressed coordinates its primal penalty is
\begin{equation*}
 \pi(r)=\frac1{1024}
 \left(\frac r{\sqrt{1-r^2/R_0^2}}-R_0\right)_+^2,
 \qquad 0\le r<R_0.
\end{equation*}
Set $\tau(r)=\pi'(r)/r$, with $\tau(0)=0$. It is continuous,
nonnegative, and positive exactly for $r>R_0/\sqrt2$. The value is
\begin{equation*}
\begin{aligned}
 V&=J_M+\sum_jq_j[1-\alpha(z_{j+1})]Q_j+\sum_j\pi(\norm{w_j}),\\
 Q_j&=6(x_j-z_{j+1}/2)^2+6\norm{A_{j,\perp}}^2
                              +\tfrac32\norm{B_{j+1,\perp}}^2.
\end{aligned}
\end{equation*}
Define conditional energies by
\begin{equation*}
 E_j=[1-\alpha(z_{j+1})]Q_j+\alpha(z_{j+1})E_{j+1},
 \qquad E_{M+1}=0.
\end{equation*}
The quadratic matrix has norm at most $12$, so telescoping gives
$0\le Q_j,E_j\le E_{\max}:=12R_0^2$, including arbitrary complements.
The source coefficient in block $j$ is $\beta_1=1$ and
$\beta_j=\Psi(z_j)+\Psi(-z_j)$ for $j\ge2$. At a nonterminal
connector $t=z_{j+1}$, with $Y=x_{j+1}$, its source derivative is
$-\mathcal B(t,Y)$, where
\begin{equation*}
 \mathcal B(t,Y)=\Psi'(-t)\Phi(-Y)+\Psi'(t)\Phi(Y)\in[0,25].
\end{equation*}
At the terminal connector it is zero. Put $q_0=1$. If $q_j>0$,
every incoming factor is positive and its connector exceeds $1/2$;
hence, also for block one,
\begin{equation*}
 0<q_j\le q_{j-1}\le1,\qquad
 \frac{q_j}{Cq_{j-1}}\le\beta_j\le\frac{eq_j}{q_{j-1}},\quad C=23.
\end{equation*}
At a current block abbreviate $x=x_j$, $t=z_{j+1}$, $q=q_j$, $q_{\rm prev}=q_{j-1}$,
$H=q[1-\alpha(t)]$, $D=x-t/2$, $E=E_{j+1}$, and
$\tau=\tau(\norm{w_j})$. All compressed residuals are as follows; the suffix enters only through the bounded scalar $E$:
\begin{align*}
 R_x&=-\beta\Phi'(x)+12HD+\tau x,\\
 R_t&=-\mathcal B-6HD+q\alpha'(t)(E-Q)+\tau t,
 \\
 \nabla_{A_\perp}V&=(12H+\tau)A_\perp,&
 \nabla_{B_\perp}V&=(3H+\tau)B_\perp.
\end{align*}

\par\textit{Step 2: a priori bounds on the block states.}
Before the inverse-cap penalty, the gradient of each compressed block has norm at most
$B_0=26+24R_0+(23\cdot5)(12R_0^2)$, the sum of bounds for the source, the direct quadratic and the gate derivative. If the full normalized uncompressed gradient has norm at most one,
$\norm{D\rho}\le1$ implies $p'(\norm{P_j})\le B_0+1$, and therefore
\begin{align*}
 \norm{P_j}&\le\overline R:=R_0+512(B_0+1),&
 \norm{w_j}&\le\overline r:=
 \frac{R_0\overline R}{\sqrt{R_0^2+\overline R^2}}<R_0,
 \\
 \sigma_{\min}(D\rho(P_j))&\ge\sigma_0:=
 (1+\overline R^2/R_0^2)^{-3/2}>0.
\end{align*}
Every compressed block gradient norm is consequently at most
$\norm{\nabla V}/\sigma_0$.

Represent a compressed block by $(x,t,\xi,\zeta)$, with
$\xi=\|A_\perp\|$ and $\zeta=\|B_\perp\|$:
\begin{equation}\label{rb:match:eq:compact-bank}
 \xi,\zeta\ge0,\quad r^2=x^2+t^2+\xi^2+\zeta^2\le\overline r^2,
 \quad \tau=\pi'(r)/r,\quad Q=6(x-t/2)^2+6\xi^2+\tfrac32\zeta^2.
\end{equation}
Let $\mathcal R=\max\{|R_x|,|R_t|,(12H+\tau)\xi,(3H+\tau)\zeta\}$.
Allowing both complementary norms only enlarges this dimension-free set.
The statements below use the relations
\begin{equation}\label{rb:match:eq:closed-interface}
 0<q\le q_{\rm prev}\le1,\qquad
 q/(Cq_{\rm prev})\le\beta\le eq/q_{\rm prev}.
\end{equation}
Fix
\begin{equation*}
 q_*=(24CR_0^2)^{-1},\quad \beta_*=e^{-99},\quad m_*=q_*\beta_*,
 \quad \ell=m_*/4,\quad E_R=e^{-4095},\quad p_{\rm ext}=eE_R/540.
\end{equation*}
Here $\beta_*=\Psi(11/20)$, and $E_R$ bounds $\Phi'(x)=\sqrt e\,e^{-x^2/2}$ when $x^2>R_0^2/2-1$, the region where an aligned block meets the active penalty. In particular $0<p_{\rm ext}<e^{-4100}<\ell<m_*/2<m_*<q_*<1$.

\par\textit{Step 3: exact local statements.}
We prove three statements about blocks whose residuals vanish exactly:
(a) if $\ell\le q\le q_*$, $q_{\rm prev}\ge\ell$ and the residuals of a nonterminal block and its successor vanish, then $t>1$, so the mass does not drop;
(b) if $q,q_{\rm prev}\ge q_*$ and the residuals of a nonterminal block vanish, then $q\alpha(t)\ge m_*$;
(c) if $q,q_{\rm prev}\ge\ell$ and $|t|\le1/8$, the residuals of the terminal block do not all vanish.
We begin with facts about a single block. Assume that all current residuals vanish and $q,\beta>0$.
For $t<1$, $H>0$ eliminates the complements. For $0<t<1$ the
source equation forces $x>t/2$. Indeed, $x\le0$ is impossible, and
$0<x\le t/2$ has radius below two, hence $\tau=0$, also impossible.
At $t=1$, $\tau x=\beta\Phi'(x)>0$ forces $x,\tau>0$ and zero
complements. For $t\le0$, $\alpha=\alpha'=0$ and the source equation
forces $D>0$, whence
\begin{equation*}
 R_t=-\mathcal B-6qD+\tau t<0.
\end{equation*}
For $0<t\le1/2$, $\mathcal B=0$ and $\tau t=6qD$, so $\tau>0$.
Every block with $0<t<1$ and $\tau>0$ satisfies
\begin{equation}\label{rb:match:eq:radial-entry-size}
 x>\sqrt{R_0^2/2-1},\qquad D>x-1/2>90,\qquad \Phi'(x)<E_R.
\end{equation}
Substitute $12q(1-\alpha)D\le\beta\Phi'(x)$ and
$\beta\le eq/q_{\rm prev}$ to obtain
\begin{equation}\label{rb:match:eq:radial-entry}
 q_{\rm prev}[1-\alpha(t)]<eE_R/1080,\qquad
 \alpha(t)\le1/2\ \Longrightarrow\ q_{\rm prev}<p_{\rm ext}.
\end{equation}

At a nonterminal connector $t>1/2$, the adjacent data are
\begin{equation}\label{rb:match:eq:adjacent-interface}
 q'=q\alpha(t)>0,\qquad\beta'=\Psi(t)+\Psi(-t)>0.
\end{equation}
Use $(Y,u,\xi',\zeta')$ for the next block, with its actual
$\tau',Q',H'=q'[1-\alpha(u)]$ and arbitrary farther
$\mathcal B'\in[0,25]$, $E'\in[0,E_{\max}]$.
If its residuals vanish, then $Y>0$. For $Y\le0,u\ge0$ its source residual cannot vanish; for $Y\le0,u<0$ that equation forces
$D'=Y-u/2>0$, making the connector residual
$-\mathcal B'-6q'D'+\tau'u<0$.

\emph{Proof of (a).}
Suppose $\ell\le q\le q_*$, $q_{\rm prev}\ge\ell$, and both blocks
have zero residuals. We show $t>1$. Nonpositive $t$ is excluded above;
$0<t\le1/2$ forces $q_{\rm prev}<p_{\rm ext}<\ell$.
For $1/2<t<1$, $Y>0$ gives $\Phi(Y)>1$, whereas
$q\alpha'(t)|E-Q|\le q_*C\Psi'(t)E_{\max}=\Psi'(t)/2$.
The connector equation and $D>0$ imply $\tau t>\Psi'(t)/2$.
Then \eqref{rb:match:eq:radial-entry-size} and the source equation
imply $\tau\le eE_R/90$ and $\Psi'(t)\le eE_R/45$.
Since $\Psi'=4\Psi/(2t-1)^3$, we have
$\alpha(t)\le C\Psi'(t)/4<1/2$, contradicting
\eqref{rb:match:eq:radial-entry}. At $t=1$, zero complements and the source equation give $x>90$ and $\tau\le eE_R/90<1$. But $\alpha(1)=1$ and $\alpha'(1)=0$ give $H=0$, so the connector equation yields
\begin{equation*}
 \tau=\mathcal B(1,Y)=\Psi'(1)\Phi(Y)=4\Phi(Y)>4.
\end{equation*}

\emph{Proof of (b).}
If $q,q_{\rm prev}\ge q_*$ and only the current nonterminal
residuals vanish (with actual $\mathcal B(t,Y)$), then
\begin{equation}\label{rb:match:eq:exact-first-drop}
 q\alpha(t)\ge m_*.
\end{equation}
For $\alpha\ge1/2$ this is immediate. Otherwise the single-block facts above
force $1/2<t<1$, $\tau=0$, $D>0$.
The source equation with $\beta/q\ge1/C$ gives $D\ge1/(24C)$, since $0<x<3/2$ and $\Phi'(x)>1/2$ when $D<1$.
Since $\alpha(t)<1/2$ gives $H>q/2$ and $\mathcal B\ge0$, the connector equation with $\tau=0$ gives
$q\alpha'(t)(E-Q)=\mathcal B+6HD\ge3qD$, hence
$\alpha'(t)\ge D/(4R_0^2)\ge1/(96CR_0^2)$.
But for $t\le11/20$, $\alpha'(t)\le4000Ce^{-99}$ is smaller.
Thus $t>11/20$ and $\alpha(t)\ge\Psi(t)>e^{-99}$.

\emph{Proof of (c).}
Here $\mathcal B=E=0$. For $q,q_{\rm prev}\ge\ell$ and $|t|\le1/8$,
zero residuals are impossible, since nonpositive $t$ was excluded and positive $t$ is flat and forces $q_{\rm prev}<p_{\rm ext}$.
\par\textit{Step 4: quantitative versions by compactness.}
Every current block obeys \eqref{rb:match:eq:compact-bank} and
\eqref{rb:match:eq:closed-interface}, with $Q,\tau$ the actual functions
of that block. Define:
\begin{enumerate}[label=(\roman*)]
\item $\mathcal K_J$: $q_*\le q\le q_{\rm prev}\le1$,
$q\alpha(t)\le m_*/2$, $0\le E\le E_{\max}$,
$|Y|\le\overline r$, and $\mathcal B=\mathcal B(t,Y)$.
Its residual is $\mathcal R$; statement (b), in the form \eqref{rb:match:eq:exact-first-drop},
excludes a zero.
\item $\mathcal K_L$: $\ell\le q\le q_*$,
$q\le q_{\rm prev}\le1$, $t\le1$, $0\le E\le E_{\max}$,
and a second block $(Y,u,\xi',\zeta')$ satisfying the same compact
radius condition. Retain $\mathcal B=\mathcal B(t,Y)$ and
\eqref{rb:match:eq:adjacent-interface}; allow independent
$E'\in[0,E_{\max}]$, $\mathcal B'\in[0,25]$. The next residuals are
\begin{align*}
 R_Y&=-\beta'\Phi'(Y)+12H'(Y-u/2)+\tau'Y,\\
 R_u&=-\mathcal B'-6H'(Y-u/2)+q'\alpha'(u)(E'-Q')+\tau'u,\\
 R_{\xi'}&=(12H'+\tau')\xi',\qquad
 R_{\zeta'}=(3H'+\tau')\zeta',\qquad H'=q'[1-\alpha(u)].
\end{align*}
Its residual is the maximum of $\mathcal R$ and the absolute values
of these four quantities. For $t\le1/2$ the current flat exclusions
apply; otherwise next-block positivity and statement (a),
including $t=1$, rule out a zero. The coefficient $\beta'=\Psi(t)+\Psi(-t)$ is defined also at $q'=0$.
\item $\mathcal K_T$: $\ell\le q\le q_{\rm prev}\le1$,
$|t|\le1/8$ and $\mathcal B=E=0$; by (c), $\mathcal R$ has no zero.
\end{enumerate}
These are fixed-dimensional compact sets. Masses in denominators are bounded away from zero, $\overline r<R_0$, and residuals are continuous
at gate joins, penalty onset and zero complements. They cover every
actual selected window, including a terminal next block through
$\mathcal B'=E'=0$; independent farther data only enlarge the sets.
Let $\epsilon_J,\epsilon_L,\epsilon_T>0$ be their residual minima, with $\min\emptyset=1$; they are positive by the no-zero arguments. Thus
residual below $\epsilon_J$ forces $q\alpha(t)>m_*/2$ in the high-mass
range, two-block residual below $\epsilon_L$ forces $t>1$ in the low-mass
range, and the terminal-band residual is at least $\epsilon_T$. Set
\begin{equation*}
 \epsilon_{\rm loc}=\min\{1,\epsilon_J,\epsilon_L,\epsilon_T\},\qquad
 c_\star=\tfrac12\min\{1,\sigma_0\epsilon_{\rm loc}\}>0.
\end{equation*}

\par\textit{Step 5: induction along the chain.}
Suppose $\norm{\nabla V}<c_\star$ and $|z_{M+1}|\le1/8$.
Radius control makes every compressed residual smaller than $\epsilon_{\rm loc}$.
We inductively prove $q_j\ge\ell$, starting from $q_1=1$.
For $j<M$, if $q_j>q_*$, separation of $\mathcal K_J$ gives
$q_{j+1}>m_*/2>\ell$. Otherwise $\ell\le q_j\le q_*$;
since $q_{j-1}\ge q_j$, separation of $\mathcal K_L$ gives
$z_{j+1}>1$, so $q_{j+1}=q_j\alpha(z_{j+1})=q_j\ge\ell$.
Thus $q_{M-1}\ge q_M\ge\ell$ (with $q_0=q_1=1$ for $M=1$),
and the terminal band contradicts separation of $\mathcal K_T$.
Restoring units multiplies the
primal gradient by $\lambda s$, proving the result.
\end{proof}

\subsection{Positive Width, Exact Optimizer Localization, and Initialization}\label{rb:sec:width}
Fix the constant $c_\star$ of the preceding barrier proof. Take
\begin{equation}\label{rb:eq:positive-width}
 \eta_0=\frac{c_\star\lambda s\mu_0}{12A_1ab\sqrt K},\qquad \eta=\min\{\eta_0,R_*/(16\sqrt n)\}>0,
 \qquad A_1=2433,\quad K=Mm.
\end{equation}
The width meets both the perturbation budget and the curvature condition
of Lemma~\ref{rb:lem:radial}(iii), without changing the number of stages.

\begin{lemma}[Uniform perturbation and initial budgets]\label{rb:lem:width-start}
For all finite raw $P$, the uncut maximizer of
$\mathcal P(P)+b\ip{\mathcal A(P)}{Y_\partial}-\mathcal E_\eta(y)$
has every relevant node, edge tuple, and complete endpoint tuple strictly inside radius $3R_*/16$. The node and endpoint cutoffs $\chi_*$ equal one, the radial edge width equals the same constant $\eta$, and every radial barrier vanishes on a neighborhood of this point. It is therefore the unique maximizer of the modified mean as well. Its value function $V_\eta$ obeys
\begin{align}
 \norm{\nabla V_\eta(P)-\nabla V_0(P)}&\le c_\star\lambda s/4,
 \label{rb:eq:gradient-transfer}\\
 |V_\eta(P)-V_0(P)|&\le(32c_\star/A_1)\lambda s^2<0.014\lambda s^2.
 \label{rb:eq:value-transfer}
\end{align}
Consequently
\begin{equation}\label{rb:eq:positive-barrier}
 |z_{M+1}|\le s/8\quad\Longrightarrow\quad
 \norm{\nabla V_\eta(P)}\ge\tfrac34c_\star\lambda s=6\eps.
\end{equation}
At the supplied origin $y^*(0)=0$,
\begin{align}
 \norm{\nabla F_\eta(0,0)}^2&=e\lambda^2s^2,\label{rb:eq:initial-gradient}\\
 V_\eta(0)-\inf_PV_\eta(P)&\le(5eM+0.014)\lambda s^2,\label{rb:eq:primal-gap}\\
 G_0:=V_\eta(0)-\inf_PF_\eta(P,0)&\le5eM\lambda s^2+dR^2<\infty.
 \label{rb:eq:full-gap}
\end{align}
The primal--dual gap $G_0$ is independent of $\eta$.
\end{lemma}
\begin{proof}
\textit{Localization.} The clipping identity
$\phi_\eta'(t)=t-\operatorname{clip}_\eta(t)$ and $\norm{D_U}\le2$
give uniformly on the entire dual space
\[
 \norm{\nabla\mathcal E_\eta-\nabla\mathcal E_0}
 \le(c+2a)\eta\sqrt K\le3a\eta\sqrt K.
\]
Strong monotonicity of either uncut energy therefore yields
\begin{equation}\label{rb:eq:optimizer-perturbation}
 \norm{y_\eta^{\rm uncut}(P)-y_0(P)}
 \le3a\eta\sqrt K/\mu_0
 \le\frac{c_\star\lambda s}{4A_1b}.
\end{equation}
The ratio of this error to $bR/\nu$ is at most
$c_\star/(6144A_1)<1$, since $R=128s$ and $b^2/\nu\ge12\lambda$.
Lemma~\ref{rb:lem:uncut} now puts every relevant node, pair and complete
endpoint tuple strictly within $3bR/\nu=3R_*/16<R_*/2$.
All node and endpoint cutoffs $\chi_*$ are identically one, every edge threshold is exactly $\eta$, and all radial barriers vanish on a
neighborhood of this point. Its uncut first-order condition is thus
also the localized first-order condition. Global mean strong concavity
makes it the unique global localized maximizer.

\par\textit{Gradients.} Both maxima are unique and the objective is continuously differentiable,
including at all hinges, so Danskin's theorem gives the value-function
gradients. Their exact difference is
\[
 b[D\mathcal A(P)]^\top\Pi_\partial[y_\eta(P)-y_0(P)],
\]
where $\Pi_\partial$ has norm one. The forcing bound and
\eqref{rb:eq:optimizer-perturbation} prove~\eqref{rb:eq:gradient-transfer},
and the preceding barrier proves~\eqref{rb:eq:positive-barrier}.

\par\textit{Values.} Since $|\phi_\eta(t)-t^2/2|\le\eta|t|$, we have
$|\mathcal E_\eta(y)-\mathcal E_0(y)|\le3a\eta\sqrt K\norm y$.
The gradients of both energies vanish at zero; strong monotonicity
therefore bounds each uncut optimizer's norm by $bR/\mu_0$.
Evaluating either variational problem at the other's optimizer proves
\eqref{rb:eq:value-transfer}.

\par\textit{Initial budgets.} At the origin $\mathcal A=0$ and the unique maximizer is zero; the only
nonzero initial derivative is
$-\sqrt e\lambda s U_{1,1}$ in the first primal subblock.
The values with positive and zero width coincide there.
Since $J_M\ge-5eM$, the nonnegative quadratic and penalty terms in
\eqref{rb:eq:zero-value}, combined with~\eqref{rb:eq:value-transfer}, give
\eqref{rb:eq:primal-gap}. Finally
\[
 F_\eta(P,0)=\lambda s^2J_M(P)-d\norm{\mathcal A(P)}^2+\sum_jp(P_j),
\]
which is width-independent and bounded below by
$-5eM\lambda s^2-dR^2$. Its value at the origin is negative.
This proves~\eqref{rb:eq:full-gap}.
\end{proof}

\par\textit{Normalization and class membership.}
All realizations meet the global smoothness and mean strong concavity
requirements by Lemma~\ref{rb:lem:class}, since $\mu_0\ge\mu$.
By~\eqref{ak:eq:macro-floor}, the primal gap is at most
$\left(\frac{5e}{32}+\frac{0.896}{4096}\right)\Delta<\Delta/2.$
The initial gradient satisfies
$e\lambda^2s^2=64e\eps^2/c_\star^2
\le e\lambda\Delta/64<2L\Delta$.
Only in the high branch, $d<4\lambda$ gives the additional check
$G_0\le\left(\frac{5e}{32}+1024\right)\Delta<1025\Delta.$
Thus the hard family also has finite initial primal--dual gap.

\section{Unrestricted Oracle Lower Bounds}\label{app:oracle}
\par\textit{A public convexification coordinate.}
For exact oracle access, augment the components already constructed by
one common public primal coordinate:
\begin{equation}\label{rp:eq:public-component}
 \widetilde f_{i,\eta}(P,\xi_{\rm pub},y)=f_{i,\eta}(P,y)+B(1-\cos\xi_{\rm pub}),
 \qquad B=L/8.
\end{equation}
We keep $V_\eta(P)$ for the value studied
above; the augmented value and gradient satisfy
\begin{equation*}
 \widetilde V_\eta(P,\xi_{\rm pub})=V_\eta(P)+B(1-\cos\xi_{\rm pub}),\qquad
 \norm{\nabla\widetilde V_\eta(P,\xi_{\rm pub})}^2
 =\norm{\nabla V_\eta(P)}^2+B^2\sin^2\xi_{\rm pub}.
\end{equation*}
The added term is nonnegative and vanishes with its gradient at zero.
Both initial gaps, the initial gradient and the dual maximizer are
unchanged. Orthogonality gives joint RMS constant
$\max\{L_{\rm rob},B\}<4L/5$, and one extra primal dimension.
All preceding widths, constants and value-transfer differences are
unchanged. For every finite $(P,\xi_{\rm pub})$,
\begin{equation}\label{rp:eq:terminal-barrier}
 |z_{M+1}(P)|\le s/8\quad\Longrightarrow\quad
 \norm{\nabla\widetilde V_\eta(P,\xi_{\rm pub})}\ge6\eps.
\end{equation}

\par\textit{Proximal notation.}
At a finite center $\bar w=(\bar P,\bar\xi_{\rm pub},\bar y)$ and inverse
step $\tau=1/\gamma>0$, the regularized objective is
\begin{equation}\label{rp:eq:regularized}
 \widetilde H_i=\widetilde f_{i,\eta}
 +\frac\tau2\bigl(\norm{P-\bar P}^2+|\xi_{\rm pub}-\bar\xi_{\rm pub}|^2
                         -\norm{y-\bar y}^2\bigr).
\end{equation}

\subsection{Exact Proximal Localization and Oracle Complexity}
\label{rp:sec:proof}
\begin{lemma}[Public legality and unique responses]
\label{rp:lem:public-legality}
The exact legal domain of~\eqref{rp:eq:regularized} is
$\tau\in[B,\infty)$, common to all components, hidden realizations and
centers. Every legal response exists and is unique. The same conclusions
hold for every consistent zero-axis formula that keeps all public terms; its response is measurable in its coefficients and center.
\end{lemma}
\begin{proof}
The scalar curvature is $\tau+B\cos\xi_{\rm pub}$, negative at $\xi_{\rm pub}=\pi$
when $\tau<B$. For $\tau\ge B$, its derivative
$\tau(\xi_{\rm pub}-\bar\xi_{\rm pub})+B\sin\xi_{\rm pub}$ tends to $\pm\infty$ and is strictly increasing. At $\tau=B$, its increment on $[\xi_1,\xi_2]$, $\xi_1<\xi_2$, is
$B\int_{\xi_1}^{\xi_2}(1+\cos t)\,dt>0$. Thus the scalar has a unique minimizer,
measurably given by the inverse of
$\xi_{\rm pub}+(B/\tau)\sin\xi_{\rm pub}=\bar\xi_{\rm pub}$. It is not strongly convex at $\tau=B$.

For the remaining variables $z=(P,y)$, let $H$ be the corresponding regularized
objective and put $T=(\nabla_PH,-\nabla_yH)$. Lemma~\ref{rb:lem:class}
gives strong monotonicity with
$m_0=\min\{\tau-K_P,\mu_0+\tau\}>0$ and a Lipschitz bound
$M_0=\sqrt nL+\tau$. Mixed Hessian terms cancel in the monotonicity
quadratic form. For $a_0=m_0/M_0^2$,
\[
 \norm{(I-a_0T)z-(I-a_0T)z'}^2
 \le(1-m_0^2/M_0^2)\norm{z-z'}^2.
\]
Contraction gives a unique zero, hence a unique global saddle by
convexity--concavity. Iteration from zero also makes this root
measurable. The same bounds hold for zeroed axes by
Lemma~\ref{rb:lem:class}. Combining this with the scalar minimizer
proves all claims, including those at the legal boundary.
\end{proof}

\subsubsection{Full-Space Localization by Reflection}

We grant the first source axis and maintain a serial frontier through
$E=M(m-1)$ fixed owned edges. A component-$i$ call tests the frontier:
a hit reveals its successor axis and continues with $i$; completing
a path also grants the next source axis. The first mismatch reveals
only that the owner is not $i$ and ends the run. Granting this entire
run can only help the algorithm.

After exposure we retain full ambient blocks for revealed axes. For an
unrevealed interior dual block retain its current-center span; for an
unrevealed endpoint retain the span of its current dual and paired
primal centers on both sides of the public identity. Their product
is $\mathcal S=\mathcal S_P\times\R_{\xi_{\rm pub}}\times\mathcal S_y$, with each unrevealed
span $S_b$ of dimension at most two. We rebuild it at each call
and replace each unrevealed axis by zero in every primal and dual
occurrence of the full-space component, retaining all public norms,
corrections and identity maps. Let $j$ be the stage whose path contains
the frontier. Its outgoing axis is unrevealed, so its gate factor
becomes $b_+(0)=1$, and we retain it.
We call the resulting function $\widetilde f_i^0$.
The first missing edge is absent because its owner is known to differ
from $i$; every later edge has two zero axes and is identically zero,
regardless of its owner, and unrevealed node terms vanish. Later sources
have $\Psi(\pm0)=0$, and the forcing of every stage after $j$ has an
incoming factor $a_+(0)=0$. The ambient forcing of stage $j$ and its
square are retained. Thus the
censored full-space formula reads only the exposed prefix.

\begin{lemma}[Exact localization of the unique proximal response]\label{rp:lem:localization}
For every legal call, the unique full-space censored saddle lies in $\mathcal S$.
If the actual axes obey
\begin{equation}\label{rp:eq:subspace-good}
 \norm{\operatorname{Proj}_{S_b}U_b}\le\delta,
 \qquad \delta=\min\left\{\frac{s}{8R},\frac{\eta}{8R_*}\right\},
\end{equation}
then it is also the unique actual proximal response. Actual and
censored center values and full joint gradients agree as well.
\end{lemma}
\begin{proof}
The full-space censored saddle exists uniquely by
Lemma~\ref{rp:lem:public-legality}. Reflect each unrevealed block through
$S_b$, using the same reflection on paired primal and dual endpoints;
fix all revealed blocks and $\xi_{\rm pub}$. Norms and radial terms are invariant,
caps and the full ambient forcing are equivariant, the paired endpoint inner product is invariant, and zeroed projections stay zero.
The center is fixed, so this reflection maps the censored saddle to
another censored saddle. Uniqueness therefore places it in
$\mathcal S$.

Under \eqref{rp:eq:subspace-good}, capped primal projections on the entire $\mathcal S$ have magnitude at most $R\delta\le s/8$. The profile satisfies $\varrho(u)\ge u/2$ for all $u\ge0$, since on $[1,2]$ this is equivalent to $u^2-3u+3\ge0$ and the other intervals are immediate. For a hidden--hidden edge with $r=\norm{Y_e}>0$, every point of the retained subspace therefore satisfies
\[
 |t_e|\le\delta(\norm{y_v}+\norm{y_w})\le\sqrt2\delta r
 \le\frac{\sqrt2\eta r}{8R_*}<\frac{\eta r}{2R_*}\le h_\eta(r).
\]
At $r=0$ the margin is $\eta>0$. Thus the edge and its full gradient vanish on an open neighborhood of every such point, at every radius. For an unrevealed node with $\norm{y_v}\le2R_*$, its hidden projection is at most $2R_*\delta\le\eta/4$; outside that radius its cutoff vanishes. The strict margin of the flat region also covers the cutoff boundary. Hence all discarded terms are locally zero in the full space. The actual and censored functions have the same values and full gradients on $\mathcal S$. The censored full-space saddle is therefore stationary for the actual convex--concave objective, hence is its unique saddle. The center itself belongs to $\mathcal S$, proving exact scalar and joint-gradient equality there as well.
\end{proof}

\subsubsection{A Total Ideal Transcript and Uniform Subspace Hiding}

Fix the finite comparison horizon
\begin{equation}\label{rp:eq:prox-horizon}
 E=M(m-1),\qquad H=E(n-1),\qquad q_P=\lfloor H/16\rfloor,
 \qquad p=1/64.
\end{equation}
We compare the algorithm with an ideal process that sees only what the owner runs have exposed.
The instance and the private seed are sampled before execution. The algorithm's query, stopping and output rules are defined on the histories the real oracle can produce. We fix a public extension of them, measurable and using the same seed, to every history of length at most $q_P$ the ideal process can produce, with valid indices, finite centers and outputs, and finite positive inverse steps; the extension need not terminate.
At each call the ideal process chooses its query from the past, exposes the owner run, and returns the full-space censored tuple; if $\tau<B$, it returns a fixed dummy instead. This public fallback reads no hidden data and reveals nothing about the real oracle. All other responses are unique and measurable by Lemma~\ref{rp:lem:public-legality}.
The algorithm is legal almost surely on each fixed instance; averaging over the prior and taking a countable union over calls gives one null event over instances and seeds, so the owner and Haar laws are never conditioned on legality or on projection success.

Conditional on the ideal history and current run exposure, every
unexposed axis is independent Haar. Its current-center span is measurable
from the prefix and private seed and has dimension at most two; earlier
projection successes are not conditioned upon \citep[cf.][Appendix B.1.2]{arjevani2023}. For Haar $U\in S^{D-1}$,
rotational symmetry and the beta integral give
$\E\ip Uv^{2r}=(2r-1)!!\norm v^{2r}/[D(D+2)\cdots(D+2r-2)]$
and zero odd moments. Bounding denominators by $D^r$ in the exponential
series yields $\E e^{\theta\ip Uv}\le e^{\theta^2\norm v^2/(2D)}$,
so Chernoff gives $\Prob\{|\ip Uv|>t\}\le2e^{-Dt^2/(2\norm v^2)}$.
A $1/2$-net of a $d$-dimensional unit sphere has at most $5^d$ points \citep[Corollary 4.2.13]{vershynin2018};
net comparison \citep[Lemma 4.4.1]{vershynin2018} therefore gives, for any fixed subspace $S$,
\begin{equation}\label{rp:eq:subspace-tail}
 \Prob\{\norm{\operatorname{Proj}_S U}>\delta\}
 \le2\,5^d\exp(-D\delta^2/8).
\end{equation}
Conditioning on the ideal history and averaging extends this
to the random current span.

We include the actual proposed output when stopping by $q_P$, and a
fixed virtual output otherwise. In each unrevealed endpoint coordinate block,
its capped primal output subblock spans at most one dimension and is determined by the ideal history, and we test that span separately. There are at most
$K(q_P+1)$ coordinate block tests in total, with $d\le2$, including the output
when $q_P=0$. With
\[
 H_n=\max\{1,R^2/s^2,R_*^2/\eta^2\},\qquad
 \delta^{-2}\le64H_n,
\]
a union bound yields a common event $G_{q_P}$ of probability at least
$1-p$ whenever
\begin{equation}\label{rp:eq:dimension-sufficient}
 D\ge \max\left\{3,\left\lceil512H_n
 \left[2\log5+\log\frac{2K(q_P+1)}p\right]\right\rceil\right\}.
\end{equation}
Localization uses only the current-center
subspace; the transcript keeps all earlier responses.

We choose
$D=\lceil1024H_n\log(1024K(q_P+1))\rceil$.
For $p=1/64$ the logarithmic bracket in
\eqref{rp:eq:dimension-sufficient} is $\log(3200K(q_P+1))$.
Since $K\ge4$,
\begin{equation}\label{rp:eq:dimension-choice}
 (1024K(q_P+1))^2\ge3200K(q_P+1).
\end{equation}
Taking logarithms proves \eqref{rp:eq:dimension-sufficient}, including $D\ge3$.

On $G_{q_P}$, induction using Lemma~\ref{rp:lem:localization} couples all exact real and ideal responses through $q_P$, since equal pasts give the
same legal query, localization gives its same unique tuple, and
fallback is unused. The output test gives \eqref{rp:eq:terminal-barrier}
when the terminal axis is hidden. The proof covers all raw radii
because primal vectors are capped, node cutoffs vanish outside radius $2R_*$, and edge
flatness holds on the entire retained cone.

\subsubsection{Posterior Owner Cost and Integer Stopping}

Keep the owners fixed before execution, independently of the
axes and private seed. The augmented ideal history contains the seed,
exposed axes, internal hits and misses, full censored responses, and
stop/output decisions; it is not conditioned on geometric success or
legality of the real oracle. Induction over internal tests leaves the frontier
owner uniform among its unexcluded labels and all successors independent and uniform, since a miss removes only its label, a hit fixes the current owner,
and a new axis is independent. Full values, gradients and proximal tuples
read only the exposed prefix; the fallback reads none.

\begin{lemma}[Posterior cost of serial owner exposure]
\label{rp:lem:posterior-cost}
For $E\ge1$ independent uniform owners on $[n]$, $n\ge2$, assume the
posterior interface above. A call traverses matching owners until its
first mismatch. Freeze at termination and let $C_q$ be complete chain
exposure within $q$ calls, for integer $q\ge0$. Then
\begin{equation}\label{rp:eq:posterior-completion}
 \Prob(C_q)\le\frac{2q}{H},\qquad H=E(n-1),\qquad q\ge0.
\end{equation}
\end{lemma}
\begin{proof}
With $r$ frontier candidates and $s$ unreached successors, set
$W=(r-1)/2+s(n-1)/2$, and $W=0$ after completion.
Repeated failed labels leave $W$ unchanged. A new label misses with
probability $(r-1)/r$, decreasing $W$ by $1/2$, or hits with probability
$1/r$, removing $(r-1)/2$ and continuing at an independent successor.
The conditional expected decrease during the entire call therefore obeys
\[
 d(r,0)=\frac{r-1}{r},\qquad
 d(r,s)=\frac{r-1}{r}+\frac{d(n,s-1)}r\le1
 \quad(s\ge1),
\]
by induction, including all consecutive hits in one call. The potential
is nonnegative and nonincreasing, starts at $W_0=H/2$, and freezes at
termination. Summing conditional expectations gives $\E[W_0-W_q]\le q$.
Completion implies $W_q=0$. Thus
$\Prob(C_q)W_0\le\E[W_0-W_q]\le q$.
\end{proof}

For $q_P=\lfloor H/16\rfloor$, the lemma gives
\begin{equation}\label{rp:eq:posterior-noncompletion}
 \Prob(C_{q_P}^c)\ge\frac78.
\end{equation}
On $C_{q_P}^c\cap\{N\le q_P\}\cap G_{q_P}$ the terminal axis remains
unexposed, the actual and ideal outputs coincide, and the output barrier
is at least $6\eps$. Uniform expected accuracy on every fixed instance, averaged
over the hard prior, implies by Markov's inequality that this event has probability
at most $1/6$. Thus
\begin{equation}\label{rp:eq:stop-tail}
 \Prob\{N>q_P\}\ge\frac78-\frac1{64}-\frac16
 =\frac{133}{192}.
\end{equation}
This argument also covers $q_P=0$, where the only required geometric test is the proposed zero-call output, already included above. Since the
query count is integer,
\begin{align}
 \E N&\ge(q_P+1)\Prob\{N>q_P\}
 \ge\frac{133H}{3072}\nonumber\\
 &=\frac{133M(m-1)(n-1)}{3072}
 \ge\frac{133nMm}{12288}.
 \label{rp:eq:expected-cost}
\end{align}
The last inequality uses $m,n\ge2$; infinite expected cost is trivial.
Since the instance prior is seed-independent, averaging the real cost selects one fixed realization
with this expected cost, while the assumed accuracy still holds there.

In the high branch, using \eqref{rb:eq:floor-bounds} in \eqref{rp:eq:expected-cost}, with
$12288=3\cdot2^{12}$ and $512=2^9$, gives
\[
 \E N\ge\frac{133c_\star^2}{3\cdot2^{73}}
 n^{3/4}\sqrt\kappa\,\frac{L\Delta}{\eps^2},
\]
proving Theorem~\ref{rb:thm:main}'s lower bound.

\par\textit{Fixed confidence.}
For a deterministic cap $q_P$, the failure probability is at least $7/8-1/64=55/64>1/3$, proving the fixed-confidence lower bound.

\subsection{Han's Protocol and All Early Horizons}\label{app:han}
\begin{corollary}[Han's protocol and one fixed instance for all early horizons]
\label{rp:cor:han-protocol}
In the window~\eqref{rb:eq:window}, the same family gives the lower
bound under \citet[Definitions 11 and 13]{han2024}, whose full-gradient refreshes occur with probability at most $c_0/n$. More precisely, put
\[
 E=M(m-1),\quad H=E(n-1),\quad C=2+c_0,\quad
 T_0=\left\lfloor\frac{H}{512C}\right\rfloor.
\]
For every algorithm in that protocol, there is one fixed hard
realization for which, simultaneously at every step $0\le t\le T_0$,
\begin{equation}\label{rp:eq:han-uniform-horizons}
 \E_{\rm seed}\norm{\nabla\widetilde V_\eta(P_t,\xi_{{\rm pub},t})}
 \ge\frac{159}{32}\eps>\eps.
\end{equation}
Consequently its first accurate horizon, and hence the
$\inf_A\sup_f\inf_t$ complexity of Definition 13, is at least
\begin{equation}\label{rp:eq:han-rate}
 \frac{H}{512(2+c_0)}
 \ge\frac{nMm}{2048(2+c_0)}
 \ge\frac{c_\star^2}{2^{72}(2+c_0)}
       n^{3/4}\sqrt\kappa\,\frac{L\Delta}{\eps^2}.
\end{equation}
The dimension $D$ chosen in Appendix~\ref{rp:sec:proof} suffices.
\end{corollary}
\begin{proof}
\textit{Outline.} We embed Han's protocol in our oracle model, add a subspace test for every early iterate to the ideal simulator, and find one event of probability at least $53/64$ on which the terminal axis stays unexposed at every early step; the terminal barrier then bounds every early gradient.

\par\textit{Embedding the protocol.} Whole-space projections are identities, so Han's fixed sampling and
linear-span operations restrict our model. Supply the common component
smoothness bound and grant the owner-independent initial full gradient,
whose only nonzero block is $-\sqrt e\lambda sU_{1,1}$ in the first primal subblock; the public coordinate has zero derivative. It therefore reveals only the already free source axis. Simulate a full
gradient by $n$ component calls, ignoring their proximal responses.
A Han step has at most two component PIFO calls and an independent
refresh coin $A_j$ with mean $\pi\le c_0/n$. The predrawn budget
$B_T=2T+n\sum_{j\le T}A_j$ dominates its cost, even under mid-step
truncation, and satisfies $\E B_T\le CT$.

\par\textit{Tested early outputs.} Write $d=H/16$ and $q_P=\lfloor d\rfloor$. Use the exact ideal
component simulator for its first $q_P$ calls. In addition to its
rank-at-most-two current-center tests, test the rank-one capped output
span for \emph{each} of the iterates $P_0,\ldots,P_{T_0}$ produced
within that call budget; assign a fixed virtual output to any step not
completed within the budget. This keeps every test measurable from
the ideal history. On $B_{T_0}\le d$, all $T_0$ steps have completed
within $q_P$ calls, so none of their outputs is virtual.

\par\textit{Subspace tests.} Use the same $\delta,H_n$ as in \eqref{rp:eq:dimension-sufficient}.
Each rank-at-most-two test fails with probability at most
$50e^{-D/(512H_n)}$. There are at most
$K(q_P+T_0+1)\le2K(q_P+1)$ tests since $T_0\le q_P$.
The same dimension therefore gives an event $G$ with
\begin{equation*}
 \Prob(G^c)
 \le\frac{100K(q_P+1)}{(1024K(q_P+1))^2}
 =\frac{100}{1024^2K(q_P+1)}<\frac1{64}.
\end{equation*}
This includes $T_0=q_P=0$, since $K\ge4$.

\par\textit{One common event.} Lemma~\ref{rp:lem:posterior-cost} gives
$\Prob(C_{q_P}^c)\ge7/8$ without conditioning on $G$.
Markov's inequality and the definition of $T_0$ give
$\Prob(B_{T_0}>d)\le CT_0/d\le1/32$.
Thus the single common event
\[
 \mathcal E=C_{q_P}^c\cap\{B_{T_0}\le d\}\cap G
\]
has probability at least $7/8-1/32-1/64=53/64$.
On this one event the terminal axis is still unexposed after all
$T_0$ steps, and therefore also at every earlier step. Exact transcript
coupling and the separately tested early outputs give
$\norm{\nabla\widetilde V_\eta(P_t,\xi_{{\rm pub},t})}\ge6\eps$ simultaneously for all
$0\le t\le T_0$.

\par\textit{A fixed realization.} Let $\mathcal B$ be the event, over real instances and seeds, that
$\norm{\nabla\widetilde V_\eta(P_t,\xi_{{\rm pub},t})}\ge6\eps$ for all $0\le t\le T_0$.
The coupling gives $\mathcal E\subseteq\mathcal B$, so
$\Prob(\mathcal B)\ge53/64$.
Averaging $\mathcal B$ selects one fixed realization with this seed
probability. Every early expected gradient there is therefore at least
$6\eps(53/64)=159\eps/32$. Definition 13 indexes the
step-$(T-1)$ point by $T$; thus the first successful horizon is at
least $T_0+1\ge H/(512C)$. The floor bounds on $m,M$ give
\eqref{rp:eq:han-rate}.

\par\textit{Membership in Han's class.} The mean is $L$-smooth by the RMS assumption, so its primal
weak-convexity parameter is at most $L$. Together with the already
proved gap and strong concavity, this places the witness in Han's
nonconvex--strongly-concave class with $\mu_x=L$. Because every legal proximal
subproblem has a unique saddle, the response does not depend on a selection rule.
\end{proof}

\section{The Strongly Concave Lower Bound for Every Condition Number}
\label{ak:app:all}

To complete the proof of Theorem~\ref{ak:thm:all}, we specialize the common path
family to $64\le\kappa\le2^{18}\sqrt n$, and then add a primal family
for $1\le\kappa<64$ and an indexed-search baseline.

\subsection{The Two-Node Specialization}\label{ak:sec:middle}
The common parameters~\eqref{rb:eq:base-parameters}--\eqref{rb:eq:green-parameters}
give $m=2$, $r=\mu/a\ge1/4$, $g=1+r$, and
\begin{equation}\label{ak:eq:middle-base}
 \lambda=\frac{2^{-39}L\kappa}{\kappa+2^{16}\sqrt n}.
\end{equation}
As before, use this family only in the path window~\eqref{ak:eq:path-window},
with the components, widths and oracle simulator of Appendices~\ref{app:robust}--\ref{app:oracle}.

\begin{corollary}[Intermediate admissibility and oracle cost]
\label{ak:cor:middle}
The augmented two-node family has joint RMS smoothness below $4L/5$,
componentwise $\mu$-strong dual concavity, the $6\eps$ terminal barrier,
$y^*(0)=0$, initial squared gradient norm at most $2L\Delta$ and primal gap
less than $\Delta/2$. Its common exact legal domain is $\tau\ge L/8$,
including a unique response at the boundary. A fixed polynomial dimension
yields a fixed hard realization with
\begin{equation}\label{ak:eq:middle-cost}
 \E N\ge\frac{133M(n-1)}{3072}\ge\frac{133nM}{6144},
\end{equation}
and the same fixed-confidence lower bound. By~\eqref{rb:eq:full-gap} and $d=6\lambda r$, the initial
primal--dual gap satisfies $G_0\le(5e/32+1536\mu/a)\Delta$.
\end{corollary}
\begin{proof}
Lemmas~\ref{rb:lem:class}, \ref{rb:lem:width-start} and
\ref{rp:lem:public-legality} apply directly to this branch.
The exact localization, total ideal transcript, posterior owner
cost of Lemma~\ref{rp:lem:posterior-cost}, and arbitrary-output test in Appendix~\ref{rp:sec:proof} then
apply without changes. Here $E=M$, so~\eqref{rp:eq:expected-cost}
gives the sharper first inequality in~\eqref{ak:eq:middle-cost};
the capped failure probability is still $55/64>1/3$.
Finally $g/r=(1+r)/r\le5$ in~\eqref{rb:eq:dimension-ratio}, hence
$H_n=O(n+M)$ and $D=O((n+M)\log(nM))$ with $D$ chosen as in
Appendix~\ref{rp:sec:proof}. Total dimension is $4MD+1$, fixed before execution.
\end{proof}

\subsection{A Primal Family with a Public Exact Convexification Threshold}
\label{ak:sec:primal}

The next family works for every $\kappa\ge1$ and will be used when
$\kappa<64$. It has no hidden dual path. Retain the primal blocks
$P_j=(X_j,Z_{j+1})$, independent Haar axes $U_{j,X},U_{j,Z}$, and the
same $w_j,x_j,z_{j+1},J_M,h_j,\mathcal A_j=h_j\Gamma w_j$.
Define hidden forcing coordinates
\[
 t_{j,1}=\ip{U_{j,X}}{\mathcal A_{j,X}}=h_jx_j,\qquad
 t_{j,2}=\ip{U_{j,Z}}{\mathcal A_{j,Z}}=-h_jz_{j+1}/2,
\]
and the flat cross term
\begin{equation}\label{ak:eq:flat-cross}
 C_\eta(t_1,t_2)=\phi_\eta(t_1+t_2)-\phi_\eta(t_1)-\phi_\eta(t_2),
 \qquad\phi_\eta(t)=\tfrac12(|t|-\eta)_+^2.
\end{equation}
Choose independent fixed owners $\Sigma_j\sim\operatorname{Unif}[n]$.
With public auxiliary primal coordinates $\xi\in\R^n$, set
\begin{align}
 g_i(P,\xi)={}&\lambda s^2J_M(P)+\sum_jp(P_j)
       +6\lambda\norm{\mathcal A(P)}^2\nonumber\\
 &+12n\lambda\sum_{j:\Sigma_j=i}C_\eta(t_{j,1},t_{j,2})
       +B(1-\cos\xi_i),\label{ak:eq:primal-components}\\
 F_i(P,\xi,y)={}&g_i(P,\xi)-\mu y^2/2,\qquad y\in\R.
 \label{ak:eq:decoupled}
\end{align}
The calibration is
\begin{align}
 \lambda&=\frac{L}{2^{34}\sqrt n},& B&=2^{32}n\lambda,&
 s&=\frac{8\eps}{c_\star\lambda},& R&=128s,\label{ak:eq:primal-scales}\\
 M&=\left\lfloor\frac{c_\star^2\lambda\Delta}{2048\eps^2}\right\rfloor,&
 \eta&=\frac{c_\star s}{96\sqrt{2M}A_1},&
 A_1&=2433,& A_2&=231000.\label{ak:eq:primal-width}
\end{align}
Again assume~\eqref{ak:eq:path-window}, with this new $\lambda$, so
$M\ge2$.

\begin{lemma}[Global smoothness and a common exact legal domain]
\label{ak:lem:primal-class}
The components~\eqref{ak:eq:decoupled} satisfy joint RMS smoothness $L$
and componentwise $\mu$-strong dual concavity. Their exact legal
domain of inverse step sizes is $\tau\in[B,\infty)$, independent of all
hidden axes, owners, component indices and centers, and every legal
response is unique. These statements also hold for censored formulas
obtained by setting any unexposed scalar projections to zero.
\end{lemma}
\begin{proof}
Almost everywhere, $\nabla^2C_\eta=a_0\begin{psmallmatrix}1&1\\1&1\end{psmallmatrix}-
\operatorname{diag}(b_0,c_0)$ for some $a_0,b_0,c_0\in\{0,1\}$; hence
$\|\nabla^2C_\eta\|\le3$ and $\|\nabla C_\eta(t)\|\le3\|t\|$.
Put $K_A=A_1^2+128A_2<3.6\cdot10^7$.
The public forcing square costs $12K_A\lambda$ in Hessian norm;
the source and penalty cost $601\lambda$ by Lemma~\ref{rb:lem:class}.
As functions of the forcing vector $A=\mathcal A(P)$, the owned terms are
$T_i(A)=12n\lambda\sum_{j:\Sigma_j=i}C_\eta(\ip{U_{j,X}}{A_{j,X}},\ip{U_{j,Z}}{A_{j,Z}})$.
Before composition, the stage pairs have disjoint
output coordinates, norm-one hidden maps, and exactly one owner.
Thus, for every direction $v$,
$(n^{-1}\sum_i\norm{D^2T_i(A)v}^2)^{1/2}\le36\sqrt n\lambda\norm v$ and
$(n^{-1}\sum_i\norm{\nabla T_i(A)}^2)^{1/2}\le36\sqrt n\lambda\norm A$.
The two chain-rule terms, with $\|A\|\le R$, cost
$36\sqrt n\lambda(A_1^2+RA_2/s)=36K_A\sqrt n\lambda$.
Minkowski's inequality gives
\begin{equation*}
 L_P\le(601+48K_A)\sqrt n\lambda<2^{31}\sqrt n\lambda.
\end{equation*}
Without averaging the same argument gives each $P$-Hessian norm
below $2^{31}n\lambda=B/2$. The $C^{1,1}$ components therefore
have these global gradient bounds. The auxiliary gradients occupy
distinct coordinates, so their RMS constant is $B/\sqrt n=L/4$.
Orthogonality of the $P,\xi,y$ blocks gives total RMS constant
$\max\{L_P,B/\sqrt n,\mu\}\le L$, including $\mu=L$.

For $\tau<B$, the $i$th auxiliary curvature at $\xi_i=\pi$ is
$\tau-B<0$, ruling out global primal convexity at every center.
For $\tau\ge B$ the $P$ part is $(\tau-B/2)$-strongly convex.
The $i$th auxiliary curvature $\tau+B\cos\xi_i$ is nonnegative;
at $\tau=B$ its isolated zeros still give a positive integral on
every nontrivial interval, hence strict convexity. The other
coordinates have positive proximal quadratics. All capped and
auxiliary terms are bounded, $p\ge0$, and the proximal quadratic
is positive, giving coercivity and a unique primal minimizer.
The unique dual maximizer is
\begin{equation}\label{ak:eq:public-dual-prox}
 y^+=\frac{\tau\bar y}{\mu+\tau}.
\end{equation}
Censoring replaces norm-one hidden maps by contractions; all estimates and the argument for the public auxiliary coordinates persist. This proves the exact
domain and unique responses for both actual and censored formulas.
\end{proof}

\begin{lemma}[Value identity, perturbation and initialization]
\label{ak:lem:primal-value}
The value of~\eqref{ak:eq:decoupled} satisfies the terminal barrier
$\norm{\nabla V_\eta(P,\xi)}\ge6\eps$ whenever $|z_{M+1}|\le s/8$,
and belongs to the stated class. Moreover $y^*(P,\xi)=0$ everywhere
and $G_0=V(0)-\inf V\le\Delta$.
\end{lemma}
\begin{proof}
At width zero, $C_0(t_1,t_2)=t_1t_2$. Hence
\begin{align*}
 6\lambda\norm{\mathcal A_j}^2+12\lambda t_{j,1}t_{j,2}
 =h_j^2\bigl[&6\lambda(x_j-z_{j+1}/2)^2
 +6\lambda\norm{A_{j,\perp}}^2\\
 &+\tfrac32\lambda\norm{B_{j+1,\perp}}^2\bigr].
\end{align*}
Since averaging leaves exactly one copy of each bridge, the $P$ value
is exactly $V_0$ in~\eqref{rb:eq:zero-value}. The auxiliary mean
$B\sum_i(1-\cos\xi_i)/n$ is nonnegative and vanishes with its gradient
at $0$.

Use $\phi_\eta'(t)=t-\operatorname{clip}_\eta(t)$ and
$|\phi_\eta(t)-t^2/2|\le\eta|t|$. The map taking each pair to its sum
has norm $\sqrt2$. Let $T_{\eta,\rm mean}=n^{-1}\sum_iT_i$, with $T_i$ from the proof of
Lemma~\ref{ak:lem:primal-class}, and let $T_{0,\rm mean}$ be its value with $C_0$.
The $M$ pair-sum clips and the $2M$ individual clips
give, before composition,
\begin{align*}
 \norm{\nabla_A(T_{\eta,\rm mean}-T_{0,\rm mean})}
 &\le24\sqrt{2M}\lambda\eta,\\
 |T_{\eta,\rm mean}-T_{0,\rm mean}|
 &\le24\sqrt{2M}\lambda\eta\norm A.
\end{align*}
After composition and the choice of $\eta$,
\begin{align*}
 \norm{\nabla_PV_\eta-\nabla V_0}&\le c_\star\lambda s/4,
 \\
 |V_\eta(P,0)-V_0(P)|&\le(32c_\star/A_1)\lambda s^2
 <0.014\lambda s^2.
\end{align*}
Lemma~\ref{rb:match:lem:uniform-barrier} now gives the terminal bound;
the orthogonal auxiliary gradient cannot cancel the $P$ gradient.
At the origin, the sole nonzero derivative has norm $\sqrt e\lambda s$.
The source bound $J_M\ge-5eM$ and~\eqref{ak:eq:macro-floor} imply
\[
 V_\eta(0)-\inf V_\eta\le(5eM+0.014)\lambda s^2<\Delta/2,
 \qquad e\lambda^2s^2\le e\lambda\Delta/64<2L\Delta.
\]
The dual is separable and maximized at zero, proving the identity for $G_0$.
\end{proof}

\subsection{Exact Proximal Hiding for the Primal Family}
\label{ak:sec:primal-oracle}

\begin{lemma}[Localization of the complete exact response]
\label{ak:lem:primal-localization}
\looseness=-1 The family~\eqref{ak:eq:decoupled} admits an exact owner-prefix
simulation for arbitrary finite centers and arbitrary legal proximal
parameters. Every unexposed primal subblock of its censored proximal
response lies in the span of that subblock's current center. If each
unexposed axis has projection norm at most $\delta_h=\eta/(4R)$ onto
that span, the actual and censored value, full gradient, and unique
proximal response agree.
\end{lemma}
\begin{proof}
Grant $U_{1,X}$ and complete each queried component's same-owner run:
a hit reveals $U_{j,Z}$ and the next source $U_{j+1,X}$; the first
mismatch reveals only failure of that label. Censor every unexposed
scalar projection, retaining public norms and maps. For an unexposed
subblock the censored regularized objective is invariant under all
orthogonal transformations fixing its center. Its unique minimizer
(Lemma~\ref{ak:lem:primal-class}) therefore lies in that center's
span, or is zero when the center is zero.

On the projection event, both center and censored minimizer have every
unexposed capped coordinate bounded by $R\delta_h=\eta/4<s/8$,
independently of raw radii, so termwise equality follows. Earlier bridges
use exposed axes. Let $j$ be the first unexposed stage. Its source and incoming connector are known, its outgoing gate factor is locally
$b_+(z_{j+1}/s)=1$, so its forcing square is radial, and its owned
interaction is absent because its owner mismatched. The gate weights
$h_k$ of later stages contain the factor $\alpha(z_{j+1}/s)$, which
vanishes locally.
Each later source vanishes through its own unexposed connector,
where both $\Psi(\pm z_k/s)$ and their derivatives vanish.
For any fully hidden pair, $|t_1|,|t_2|\le\eta/4$ and
$|t_1+t_2|\le\eta/2$, so all three terms of
\eqref{ak:eq:flat-cross} vanish with their full gradients.
These strict neighborhood margins establish equality of center tuples
and of full gradients at the censored minimizer, including omitted
directions. It is thus stationary for the actual globally convex
regularized component and equals its unique minimizer. The dual
response is the public \eqref{ak:eq:public-dual-prox}.
\end{proof}

\begin{lemma}[Primal-family information bound]
\label{ak:lem:primal-cost}
Under~\eqref{ak:eq:path-window}, there is a fixed realization of
\eqref{ak:eq:decoupled} with polynomial dimension such that
$\E N\ge133nM/6144$. The assertion includes arbitrary randomized
query points, legal proximal parameters, stopping times and outputs.
Up to the same smaller constant, it also holds for success probability
$2/3$ under a deterministic cap on the number of queries.
\end{lemma}
\begin{proof}
We run the argument of Appendix~\ref{rp:sec:proof} with three changes.
First, the localizer is Lemma~\ref{ak:lem:primal-localization}, the public legality test is $\tau<B$, and all admissible censored problems have unique responses by Lemma~\ref{ak:lem:primal-class}; fallback and exposed owner runs therefore read only the revealed prefix, and the null-set argument handles almost-sure legality as before.
Second, each current-center span has rank at most one, there are $K=2M$ blocks, and we put $H=M(n-1)$, $q=\lfloor H/16\rfloor$ and
\begin{equation}\label{ak:eq:primal-dimension}
 D=\left\lceil1024\delta_h^{-2}\log(1024K(q+1))\right\rceil.
\end{equation}
The spherical bound \eqref{rp:eq:subspace-tail} and a union bound over at most $K(q+1)$ tests, including a stopped or virtual output, give geometric failure below $1/64$, because $D\delta_h^2/8\ge128\log(1024K(q+1))\ge\log(640K(q+1))$. Off this event an unexposed terminal axis has $|z_{M+1}|\le R\delta_h=\eta/4<s/8$, hence gradient at least $6\eps$.
Third, each same-owner run stops at its first miss, and each miss response reads only the prefix and the failed label, so the augmented ideal history satisfies the posterior interface of Lemma~\ref{rp:lem:posterior-cost} with $E=M$, without conditioning on geometric success.

With these inputs, \eqref{rp:eq:posterior-noncompletion}, \eqref{rp:eq:stop-tail} and \eqref{rp:eq:expected-cost} give $\Prob(N>q)\ge133/192$ and $\E N\ge133H/3072\ge133nM/6144$, including $q=0$. Averaging over the hard prior selects one realization, and under a deterministic cap $q$ the failure probability is at least $7/8-1/64=55/64>1/3$. Since $\delta_h^{-2}=O(M)$, the dimension is $2MD+n+1=O(M^2\log(nM)+n)$.
\end{proof}

\subsection{The Indexed Baseline and Completion of All Accuracy Windows}
\label{ak:sec:assembly}

\begin{lemma}[An indexed baseline with exact proximal responses]
\label{ak:lem:n-baseline}
For $n\ge2$, $\kappa\ge1$, and $\eps^2\le2L\Delta/9$, the class
contains fixed instances requiring worst-case expected cost at least
$n/16$ for expected-norm accuracy. The same lower bound holds for
success probability $2/3$ under a deterministic query cap.
\end{lemma}
\begin{proof}
Choose independent uniform $I\in[n]$ and sign $\sigma\in\{-1,1\}$,
and set (cf.\ the hidden-component quadratic of \citealp[Appendix D.3]{zhou2026ncsc})
\[
 F_i(x,y)=Lx^2/2-3n\sigma\eps\one_{\{i=I\}}x-\mu y^2/2.
\]
The joint RMS constant is $L$, every component is $\mu$-strongly
concave in $y$, and every $\tau>0$ is legal with a unique response.
The mean initial gap and squared joint gradient are
$9\eps^2/(2L)$ and $9\eps^2$, respectively, within the budgets.
Before index $I$ is queried, center values, gradients, and proximal
responses are all sign-independent. At any sign-independent output,
\[
 \tfrac12|Lx-3\eps|+\tfrac12|Lx+3\eps|\ge3\eps\ge2\eps.
\]
For $q=\lfloor n/4\rfloor$, adaptive search hits $I$ within $q$
calls with probability at most $q/n$, since before the hit the label sequence depends only on misses. Thus expected accuracy implies
\[
 \eps\ge2\eps[1-\Prob(N>q)-q/n],\qquad
 \E N\ge(q+1)/4\ge n/16.
\]
For a deterministic cap $q$, the success intervals for the two signs are
disjoint and success is at most $1/2+q/(2n)\le5/8<2/3$.
Averaging over the prior fixes a hard instance.
\end{proof}

\begin{proof}[Proof of Theorem~\ref{ak:thm:all}]
Write $S=L\Delta/\eps^2$ and
$h(n,\kappa)=\min\{\sqrt n\kappa,n^{3/4}\sqrt\kappa\}$.
The common accuracy window is stronger than that of
Lemma~\ref{ak:lem:n-baseline}, so the $n/16$ baseline is always
available. If $\kappa\ge2^{18}\sqrt n$, Theorem~\ref{rb:thm:main}
gives the accuracy term, which is the second branch of $h$.

For $64\le\kappa\le2^{18}\sqrt n$, the two-node coefficient obeys
\[
 \lambda=\frac{2^{-39}L\kappa}{\kappa+2^{16}\sqrt n}
 \ge\frac{L\kappa}{5\cdot2^{55}\sqrt n}.
\]
If its path window holds, Corollary~\ref{ak:cor:middle} and the
floor bound give
\[
 \E N\ge\frac{133c_\star^2}{6144\cdot4096}
 \frac{n\lambda\Delta}{\eps^2}
 \ge\frac{133c_\star^2}{15\cdot2^{78}}\sqrt n\kappa S
 \ge\frac{133c_\star^2}{15\cdot2^{78}}h(n,\kappa)S.
\]
If the path window fails, instead
\[
 h(n,\kappa)S\le\sqrt n\kappa S
 <\frac{5\cdot2^{67}}{c_\star^2}\,n.
\]
The indexed baseline then supplies the requested sum up to a
universal constant.

For $1\le\kappa<64$, use the primal family. In its path window,
Lemma~\ref{ak:lem:primal-cost} gives
\[
 \E N\ge\frac{133c_\star^2}{6144\cdot4096\cdot2^{34}}
 \sqrt nS=\frac{133c_\star^2}{3\cdot2^{57}}\sqrt nS.
\]
Since $h(n,\kappa)\le64\sqrt n$, this bounds the required
accuracy term. If its path window fails,
$\sqrt nS<2^{46}n/c_\star^2$, so
$h(n,\kappa)S<2^{52}n/c_\star^2$, again covered by the
indexed baseline. In a path window, the maximum of the baseline and
accuracy lower bounds is at least half their sum. Outside it, the
accuracy term is at most the displayed universal multiple of $n$.
The common constant $c=c_\star^2/2^{96}$ is valid in
all cases. Both lower bounds are worst cases over instances with the same
public parameters.

The confidence statements of the three families and the baseline
give the same rate with this smaller common constant. All components
are $C^{1,1}$ and $\mu$-strongly concave, have $y^*(0)=0$, and have
unique responses on common exact legal domains. The long-path
dimension is~\eqref{eq:dimension-main}; the intermediate and primal
dimensions were proved above and are polynomial since $M=O(S)$.
For $G_0$, the long-path family
has $G_0\le1025\Delta$, the primal family has $G_0\le\Delta$, and
the intermediate family has the bound of
Corollary~\ref{ak:cor:middle}.
\end{proof}

\Needspace{8\baselineskip}
\section{A Normalized-Start Upper Bound for Every Condition Number}
\label{nu:app:upper}
We use the stochastic projected damped extragradient (SPDE) update of
\citet{zhang2026spde} and the solver for the mean operator
of \citet{alacaoglu2022} and \citet{luo2022}. Fixed differentiable components satisfy
\eqref{eq:rms}, with mean $\mu$-strong concavity, $n\ge2$, $\kappa\ge1$
and $S=L\Delta/\eps^2\ge2$. The finite supplied pair $u_{\rm in}=(x_{\rm in},y_{\rm in})$
obeys $V(x_{\rm in})-V_*\le\Delta$ and
$\|\nabla F(u_{\rm in})\|^2\le2L\Delta$; in this appendix we write $x$ for the
primal variable and $\mathcal M$ for the Moreau envelope below; as in the lower-bound appendices, symbols defined in Appendices~\ref{nu:app:upper} and~\ref{app:upper} are local. Proximal regularizers use exact arithmetic;
each gradient call queries one component $F_i$.

\begin{theorem}[Normalized-start algorithm for all condition numbers]
\label{nu:thm:normalized}
Under these assumptions, a randomized component-gradient algorithm achieves
$\E\|\nabla V(\widehat x)\|^2\le\eps^2$ with expected cost at most
\[
 C_N\bigl[n+\sqrt n\kappa S+
             (n+\sqrt n\kappa)\log(2\kappa)\bigr]
\]
for a universal constant $C_N$. The iteration budgets are finite and fixed from supplied parameters.
This statement holds for every $\kappa\ge1$.
\end{theorem}
We prove the theorem by initializing the proximal energy, running a fixed
SPDE budget with recursive estimates, and refining one uniformly selected
endpoint (Algorithm~\ref{alg:normalized}). All three stages use component gradients.

\begin{algorithm}[t]
\caption{Normalized-start method of Theorem~\ref{nu:thm:normalized}}\label{alg:normalized}
\small
\begin{algorithmic}[1]
\Require $L,\mu,\Delta,\eps$; a start $u_{\rm in}=(x_{\rm in},y_{\rm in})$ with $V(x_{\rm in})-V_*\le\Delta$ and $\norm{\nabla F(u_{\rm in})}^2\le2L\Delta$
\State set the parameters~\eqref{nu:eq:parameters}, \eqref{nu:eq:exact-schedule} and~\eqref{nu:eq:schedules}, and the budgets $j_w$ of~\eqref{nu:eq:warm-budget} and $j_p$ of~\eqref{nu:eq:polish}
\State $u_w\gets$ endpoint of $j_w$ L-SVRE iterations~\eqref{nu:eq:solver-updates} for $Q_{x_{\rm in}}$, from $u_{\rm in}$
\State $(x_0,y_0)\gets u_w$,\quad $z_0\gets x_{\rm in}$,\quad $v_0\gets0$
\For{$t=0,\ldots,T-1$}
\State apply the updates~\eqref{nu:eq:updates}, whose gradient requests $j=3t,3t+1,3t+2$ use
\Statex\hspace{3.2em}$\widehat g_j=\nabla F(u_j)$ if $j\equiv0\pmod q$; otherwise~\eqref{nu:eq:exact-recursion} with $b$ fresh uniform indices
\EndFor
\State draw $J$ uniformly from $\{0,\ldots,T-1\}$
\State $\widehat u\gets$ endpoint of $j_p$ L-SVRE iterations for $Q_{z_J}$ with fresh randomness, from $(x_{J+1},y_{J+1})$
\State \Return the primal coordinate $\widehat x$ of $\widehat u$
\end{algorithmic}
\end{algorithm}
The component inequality is
\begin{equation}\label{nu:eq:rms}
 n^{-1}\sum_i\|\nabla F_i(u)-\nabla F_i(u')\|^2
 \le L^2\|u-u'\|^2.
\end{equation}

\subsection{Unconstrained Proximal Geometry}
\label{nu:sec:geometry}
Strong concavity makes $F(x,\cdot)$ coercive from above, so its maximum
is attained uniquely at $y^*(x)$. Joint mean $L$-smoothness follows
from Jensen's inequality in \eqref{nu:eq:rms}. The two dual optimality
conditions imply $\norm{y^*(x)-y^*(x')}\le\kappa\norm{x-x'}$.
Consequently $V$ is differentiable, $L(1+\kappa)$-smooth and
$L$-weakly convex (cf.\ \citealp[Lemma 4.3]{lin2020gda}, stated there for bounded dual sets). Its infimum $V_*>-\infty$ by the supplied primal
gap. Define the proximal objective $Q_z$, the Moreau envelope $\mathcal M$ of $V$ \citep{davis2019}, and the proximal saddle $u_z^*$ by
\begin{equation}\label{nu:eq:geometry}
 \begin{gathered}
 Q_z(x,y)=F(x,y)+L\norm{x-z}^2,\qquad
 \mathcal M(z)=\min_x\{V(x)+L\norm{x-z}^2\},\\
 u_z^*=(x_z^*,y_z^*)=
  (\argmin_x\{V(x)+L\norm{x-z}^2\},y^*(x_z^*)),\\
 \inf\mathcal M=V_*,\qquad \nabla\mathcal M(z)=2L(z-x_z^*)
                     =\nabla_xF(u_z^*),\qquad \nabla_yF(u_z^*)=0.
 \end{gathered}
\end{equation}
The proximal objective is coercive and $L$-strongly convex, hence has a
unique minimizer. Its first-order condition and dual optimality make
$u_z^*$ the unique unconstrained saddle. Strong monotonicity of the mean proximal saddle
operator gives
\[
 L\norm{x_z^*-x_{z'}^*}^2+\mu\norm{y_z^*-y_{z'}^*}^2
       \le2L\ip{z-z'}{x_z^*-x_{z'}^*}.
\]
Young's inequality therefore yields
\begin{equation}\label{nu:eq:sensitivity}
 \norm{x_z^*-x_{z'}^*}\le2\norm{z-z'},\qquad
 \norm{y_z^*-y_{z'}^*}\le\sqrt{2L/\mu}\norm{z-z'},
 \qquad \operatorname{Lip}(\nabla\mathcal M)\le6L.
\end{equation}

\subsection{Pathwise SPDE Descent and Motion Bounds}
\label{nu:sec:spde}
Set
\begin{equation}\label{nu:eq:parameters}
 \begin{gathered}
 s=\sqrt{2L\mu},\quad h=(192L)^{-1},\quad d=1/64,\quad
 \alpha=(1+hs/16)^{-1},\\
 k=\frac{h(8L-s)}{16+hs},\qquad \beta=hs/4096,\qquad C=140.
 \end{gathered}
\end{equation}
From any initial state, use three gradient requests
$\widehat g_j$ at points $u_j$ per iteration:
\begin{equation}\label{nu:eq:updates}
 \begin{aligned}
 u_{3t}&=(x_t,y_t),&
 \widetilde x_t&=x_t-h[\widehat g_{x,3t}+2L(x_t-z_t)],\\
 \widetilde y_t&=y_t+h(\widehat g_{y,3t}+v_t),&
 u_{3t+1}&=(\widetilde x_t,\widetilde y_t),\\
 \bar v_t&=\alpha v_t+k\widehat g_{y,3t+1},&
 x_{t+1}&=x_t-h[\widehat g_{x,3t+1}+2L(\widetilde x_t-z_t)],\\
 y_{t+1}&=y_t+h(\widehat g_{y,3t+1}+\bar v_t),&
 u_{3t+2}&=(x_{t+1},y_{t+1}),\\
 v_{t+1}&=\alpha v_t+k\widehat g_{y,3t+2},&
 z_{t+1}&=z_t+\beta(x_{t+1}-z_t).
 \end{aligned}
\end{equation}
Always $u_{3t}=u_{3t-1}$ for $t\ge1$. Proximal gradients are
evaluated by exact arithmetic, outside the gradient estimator.

For $w=(x,y,v)$ at center $z$, write
$X=x-x_z^*$, $Y'=y-y_z^*$, $P=\nabla_xQ_z(x,y)$,
$R=-\nabla_yF(x,y)$ and $Q=R-v$. The common potential is
\begin{equation}\label{nu:eq:potential}
 \begin{aligned}
 E_z(w)&=-\left(1-\frac{s}{16L}\right)(Q_z(x,y)-\mathcal M(z))
       +\frac{\norm P^2+\norm Q^2}{L}
       +\frac{s}{16L}\ip v{Y'}+\frac\mu{256}\norm{Y'}^2,\\
 H_z(w)&=E_z(w)+\frac{s}{256L^2}\norm v^2,\qquad
 \mathcal V(z,w)=\mathcal M(z)-V_*+H_z(w).
 \end{aligned}
\end{equation}
Strong saddle inequalities give
\[
 Q_z(x,y)-\mathcal M(z)+\ip R{Y'}\ge\frac L2\norm X^2+\frac\mu2\norm{Y'}^2,
 \qquad
 Q_z(x,y)-\mathcal M(z)\le\frac{\norm P^2}{2L}-\frac\mu2\norm{Y'}^2.
\]
Completing the $Q,Y'$ square yields the following bound; in particular $\mathcal V\ge\mathcal M-V_*\ge0$.
\begin{equation}\label{nu:eq:positive}
 E_z(w)\ge\frac{\norm P^2}{2L}+\frac{s}{32}\norm X^2
       +\frac\mu2\norm{Y'}^2+\frac1L\norm{Q-sY'/32}^2\ge0.
\end{equation}
Let $\eta_j=\widehat g_j-\nabla F(u_j)$ and
$(a_t,b_t,c_t)=(\eta_{3t},\eta_{3t+1},\eta_{3t+2})$. The proof uses
\begin{equation}\label{nu:eq:virtual}
 v_{t+1}^{\circ}=\alpha v_t+
 k\nabla_yF(x_{t+1},y_{t+1}),\qquad
 v_{t+1}=v_{t+1}^{\circ}+kc_{y,t}.
\end{equation}
This is the true-gradient momentum at the realized endpoint. Fix center $z_t$ when defining $P_t,Q_t$ as
above, and set
\[
 P_t^+=\nabla_xQ_{z_t}(u_{3t+2}),\quad
 Q_t^+=-\nabla_yF(u_{3t+2})-v_{t+1}^{\circ},\quad
 X_t^+=x_{t+1}-x_{z_t}^*,\quad Y_t^+=y_{t+1}-y_{z_t}^*,
\]
\begin{equation}\label{nu:eq:dissipation}
 \begin{aligned}
 \mathcal I_t&=\norm{P_t^+-P_t}^2+\norm{Q_t^+-Q_t}^2,\\
 D_t&=\frac h4\norm{P_t^+}^2+\frac{24\beta}{L}\norm{Q_t^+}^2
   +\frac{7\beta}{L}\norm{v_{t+1}^{\circ}}^2
   +32\mu\beta\norm{Y_t^+}^2
   +4\beta(16L-s)\norm{X_t^+}^2+\frac{\mathcal I_t}{4L}.
 \end{aligned}
\end{equation}
\begin{lemma}[Pathwise drift and joint motion]
\label{nu:lem:pathwise}
For arbitrary realized vector errors and every finite initial state,
with $\mathcal V_t=\mathcal V(z_t,w_t)$, the updates satisfy
\begin{align}
 \mathcal V_{t+1}-\mathcal V_t
 &\le-\tfrac14D_t-\frac\beta{16L}\norm{\nabla\mathcal M(z_t)}^2
       +\frac C{L\beta}(\norm{a_t}^2+\norm{b_t}^2+\norm{c_t}^2),
       \label{nu:eq:drift}\\
 \norm{u_{3t+1}-u_{3t}}^2+\norm{u_{3t+2}-u_{3t+1}}^2
 &\le\frac{D_t}{L\beta}
             +\frac{\norm{a_t}^2+\norm{b_t}^2}{L^2}.
       \label{nu:eq:joint-motion}
\end{align}
\end{lemma}
\begin{proof}
The drift and joint-motion estimates are the deterministic inequalities
of \citet[Lemmas~11--12, Eqs.~(117) and~(122)]{zhang2026spde}.
Use their strongly concave substitution from the proof of Lemma~10,
Eq.~(100), as explicitly applied to these lemmas in the proof of
Lemma~15: $G_{\rm sc}=Q_z$, $p_{\rm sc}=\mathcal M$, all normal-cone vectors
are zero, the virtual momentum is $v_{t+1}^{\circ}$, and their
$D_t^0$ is~\eqref{nu:eq:dissipation}. Their numerical parameters are
exactly~\eqref{nu:eq:parameters}, including $C=140$.

The cited lemmas are stated for a compact dual domain.
Equations~\eqref{nu:eq:geometry}--\eqref{nu:eq:sensitivity}
and~\eqref{nu:eq:positive} supply the all-space saddles, envelope and
nonnegative energy. For one finite realized step, choose compact primal
and dual balls whose interiors contain its centers, its query points and
the saddle path between the two centers; such balls exist
by~\eqref{nu:eq:sensitivity}. On these balls the projections are
inactive, and the saddles and envelope values coincide with their
all-space counterparts. The lower energy floor changes by an additive
constant, which cancels in one-step differences. Their fixed-center
Lemma~5 starts from arbitrary feasible states, and the subsequent center
and endpoint estimates, including the endpoint expansion Eq.~(88) and
the Young bounds (118)--(121), are algebraic in the realized errors.
Hence both estimates hold for arbitrary finite initial states and
correlated errors.

\end{proof}

\subsection{A Shared Solver with Mean Strong Monotonicity}
\label{nu:sec:solver}
We use loopless stochastic variance-reduced extragradient (L-SVRE)
\citep{alacaoglu2022,luo2022}, with the following self-contained guarantee.
\begin{lemma}[L-SVRE contraction under mean strong monotonicity]\label{nu:lem:lsvre}
Let $n\ge2$ and $A=n^{-1}\sum_iA_i$ on $\R^d$ have a zero $z^*$,
be $\alpha$-strongly monotone, and satisfy
$n^{-1}\sum_i\|A_i(z)-A_i(z')\|^2\le M_s^2\|z-z'\|^2$.
No component monotonicity is assumed. Starting from $r_0=w_0=u$,
pay $n$ calls for $A(w_0)$ and, with a fresh uniform index $i$, perform
\begin{equation}\label{nu:eq:solver-updates}
 \begin{aligned}
 \bar r&=(1-p_s)r+p_sw,& b&=\bar r-\tau A(w),\\
 v_s&=A(w)+A_i(b)-A_i(w),& r^+&=\bar r-\tau v_s,\\
 p_s&=1/(2n),& \tau&=1/(4\sqrt nM_s).
 \end{aligned}
\end{equation}
Independently refresh $w^+=r^+$ with probability $p_s$, paying $n$
calls for its mean, and otherwise retain $w$. Then, for every $j\ge0$,
\begin{equation}\label{nu:eq:solver-contraction}
 \E\|r_j-z^*\|^2\le4(1-\theta)^j\|u-z^*\|^2,
 \qquad \theta^{-1}=4n+8\sqrt nM_s/\alpha.
\end{equation}
Each iteration costs at most $2+np_s=5/2$ calls in conditional expectation;
the guarantee holds conditionally on preceding history, with fresh randomness.
\end{lemma}
\begin{proof}
Condition on $(r,w)$ and write $p=p_s$, $t=\tau\alpha$,
$R=\|r-z^*\|^2$, $W=\|w-z^*\|^2$. Unbiasedness gives
$\E v_s=A(b)$ and
$\E\|r^+-b\|^2\le\tau^2M_s^2\|b-w\|^2$. The identity
\[
 \E R^+=\|\bar r-z^*\|^2-2\tau\langle A(b),b-z^*\rangle
 -\|\bar r-b\|^2+\E\|r^+-b\|^2
\]
and $R^+\le2\|b-z^*\|^2+2\|r^+-b\|^2$ imply
\[
 (1+t)\E R^+\le\|\bar r-z^*\|^2-\|\bar r-b\|^2
 +(1+2t)\tau^2M_s^2\|b-w\|^2.
\]
Mean Lipschitzness and strong monotonicity give $\alpha\le M_s$;
hence $p,t\le1/4$ and $\tau^2M_s^2=p/8$. Expand
$\|\bar r-z^*\|^2=(1-p)R+pW-p(1-p)\|r-w\|^2$ and use
$\|b-w\|^2\le2\|b-\bar r\|^2+2(1-p)^2\|r-w\|^2$.
Both motion coefficients are nonpositive, so
\begin{equation*}
 (1+t)\E R^+\le(1-p)R+pW.
\end{equation*}
Set $\omega=2(p+t)/[(2p+t)(1-p)]$, $\Lambda=R+\omega W$ and
$\theta=pt/[2(p+t)]$. Since $\E W^+=p\E R^++(1-p)W$,
the relative decreases of the $R$ and $\omega W$ coefficients are
\[
 \frac{t(p+t)}{(2p+t)(1+t)}\ge\theta,
 \qquad
 \frac{pt(1+p+2t)}{2(p+t)(1+t)}\ge\theta.
\]
The first inequality is equivalent to
$t[3p-2p^2+(2-p)t]\ge0$; the second is immediate. Thus
$\E\Lambda^+\le(1-\theta)\Lambda$.
As $1+\omega\le1+2/(1-p)\le11/3<4$, iteration proves
\eqref{nu:eq:solver-contraction}.
\end{proof}

For SPDE, use $A_i=(\nabla_xF_i+2L(x-z),-\nabla_yF_i)$.
Minkowski's inequality gives RMS constant $M_s=3L$; proximal geometry gives
mean curvature $\alpha=\mu$, and the zero of the operator is $u_z^*$. Thus
$\theta^{-1}=4(n+6\sqrt n\kappa)$ in
\eqref{nu:eq:solver-contraction}.

\subsection{Initialization at the Input Center}
\label{nu:sec:initialization}
For a normalized start,
use an initial proximal solve for
$Q_{x_{\rm in}}=F+L\norm{x-x_{\rm in}}^2$. Its unique saddle $u_*=(x_*,y_*)$
exists by \eqref{nu:eq:geometry} using only $V_*>-\infty$, and
$\nabla Q_{x_{\rm in}}(u_*)=0$. At $u_{\rm in}$ the regularizer gradient vanishes,
so strong monotonicity implies the deterministic bound
\begin{equation*}
 \norm{u_{\rm in}-u_*}^2
 \le\frac{\norm{\nabla F(u_{\rm in})}^2}{\mu^2}
 \le\frac{2\kappa^2\Delta}{L}.
\end{equation*}
Run \eqref{nu:eq:solver-updates} on the whole space for
\begin{equation}\label{nu:eq:warm-budget}
 j_w=\left\lceil4(n+6\sqrt n\kappa)\log(512\kappa^2)\right\rceil,
 \qquad
 \E\norm{u_w-u_*}^2\le\Delta/(64L).
\end{equation}
By \eqref{nu:eq:solver-contraction}, the cost including the snapshot is
$O((n+\sqrt n\kappa)\log(2\kappa))$, with all budgets supplied.

Initialize at $u_w$, with
\begin{equation*}
 z_0=x_{\rm in},\qquad v_0=0.
\end{equation*}
Since
$Q_{x_{\rm in}}(u_*)=\mathcal M(x_{\rm in})$, joint $3L$-smoothness and
$\nabla Q_{x_{\rm in}}(u_*)=0$ imply
\[
 |Q_{x_{\rm in}}(u_w)-\mathcal M(x_{\rm in})|\le(3L/2)\norm{u_w-u_*}^2,
 \qquad \norm{\nabla Q_{x_{\rm in}}(u_w)}^2\le9L^2\norm{u_w-u_*}^2.
\]
Using \eqref{nu:eq:positive} gives
\begin{equation*}
 0\le H_{z_0}(u_w,0)\le11L\norm{u_w-u_*}^2,
 \qquad \E\mathcal V_0\le\Delta+11\Delta/64<2\Delta.
\end{equation*}
Main-stage batches are fresh after this fixed-budget initialization.

\subsection{Recursive Gradient Estimation with Full Gradient Refreshes}
\label{nu:sec:estimators}
Write $A_m=\sum_{t<m}\E D_t$ and
$E(m)=\sum_{j<3m}\E\norm{\eta_j}^2$. Let $\mathcal H_j$ be the
pre-query history, which includes
the chosen point, all previous estimates and all past randomness, but
excludes the fresh batches at that request. Fixed finite budgets, finite-support component draws and Bernoulli
refreshes leave finitely many states at each finite horizon, including initialization, so all conditional expectations below exist.

\par\textit{Full-gradient refreshes and recursive updates.}
At $j\equiv0\pmod q$, use $\widehat g_j=\nabla F(u_j)$, at cost $n$.
Otherwise, with $b$ fresh uniform indices sampled with replacement, use the
recursive estimate of \citet{nguyen2017sarah} and \citet{fang2018spider},
\begin{equation}\label{nu:eq:exact-recursion}
 \widehat g_j=\widehat g_{j-1}
   +\frac1b\sum_{i\in I_j}[\nabla F_i(u_j)-\nabla F_i(u_{j-1})].
\end{equation}
Although $\E[\eta_j\mid\mathcal H_j]=\eta_{j-1}$, the centered new
increment and \eqref{nu:eq:rms} give
\[
 \E[\norm{\eta_j}^2\mid\mathcal H_j]
 \le\norm{\eta_{j-1}}^2+(L^2/b)\norm{u_j-u_{j-1}}^2;
\]
at refreshes $\eta_j=0$. Summing within epochs counts each motion at
most $q$ times. Since consecutive iterations share their boundary points,
\eqref{nu:eq:joint-motion} gives
$E(m)\le\frac{qL}{b\beta}A_m+\frac qb E(m).$
Set
\begin{equation}\label{nu:eq:exact-schedule}
 A=16C/\beta^2,\qquad q=\max\{1,\lfloor\sqrt{n/A}\rfloor\},
 \qquad b=\lceil Aq\rceil.
\end{equation}
Since $q/b\le1/A<1/2$, the preceding recursion gives
$E(m)\le2qLA_m/(b\beta)$ and
$CE(m)/(L\beta)\le2CA_m/(A\beta^2)=A_m/8$.
When $q=1$, every request is exact, $E(m)=0$, and the difference
batch is unused. Telescoping \eqref{nu:eq:drift}
from any initialization with $\E\mathcal V_0\le B_0:=2\Delta$ gives
\begin{equation}\label{nu:eq:exact-absorb}
 \E\mathcal V_m+\tfrac18A_m+
 \frac\beta{16L}\sum_{t<m}\E\norm{\nabla\mathcal M(z_t)}^2\le B_0.
\end{equation}

\subsection{Common Budgets, Dissipation, and Output Refinement}
\label{nu:sec:output}
For the main stage choose
\begin{equation}\label{nu:eq:schedules}
 \delta=\eps/16,\qquad B_0=2\Delta,\qquad
 T=\left\lceil\frac{128LB_0}{\beta\delta^2}\right\rceil.
\end{equation}
Choose $J$ independently and uniformly from $\{0,\ldots,T-1\}$.
Then \eqref{nu:eq:exact-absorb} gives
\begin{equation}\label{nu:eq:main-output}
 \E\norm{\nabla\mathcal M(z_J)}^2\le\delta^2/8=\eps^2/2048,
 \qquad A_T\le16\Delta,\qquad
 \E[\mathcal M(z_J)-V_*]\le2\Delta.
\end{equation}

The seed for the final proximal solve is
$u_{\rm seed}=(x_{J+1},y_{J+1})$ at center $z_J$.
The distance terms in \eqref{nu:eq:dissipation} already give,
since $\mu\le L$ and $s\le2L$,
\[
 D_t\ge32\mu\beta(\|X_t^+\|^2+\|Y_t^+\|^2)
 =32\mu\beta\|u_{3t+2}-u_{z_t}^*\|^2.
\]
Here $X_t^+,Y_t^+$ are measured from the saddle at the center $z_t$. As $A_T/T\le\beta\delta^2/(16L)$, averaging over $J$ yields
\begin{equation*}
 \E\|u_{\rm seed}-u_{z_J}^*\|^2
 \le\frac{A_T}{32\mu\beta T}
 \le\frac{\kappa\delta^2}{512L^2}
 \le\frac{\kappa^2\eps^2}{L^2}.
\end{equation*}
Use fresh solver randomness at this center and seed, with the fixed budget
\begin{equation}\label{nu:eq:polish}
 j_p=\left\lceil4(n+6\sqrt n\kappa)
       \log[1024\kappa^2(1+\kappa)^2]\right\rceil,
 \qquad
 \E\|\widehat u-u_{z_J}^*\|^2
 \le\frac{\eps^2}{256L^2(1+\kappa)^2}.
\end{equation}
Indeed, condition on the complete history of the main stage and $J$.
Equation~\eqref{nu:eq:solver-contraction} bounds the conditional
mean-square distance by
$4e^{-\theta j_p}\|u_{\rm seed}-u_{z_J}^*\|^2$; then use
$4e^{-\theta j_p}\le[256\kappa^2(1+\kappa)^2]^{-1}$
and average. This solve includes its own $n$-call snapshot.

Return the primal coordinate $\widehat x$ of $\widehat u$.
The geometry above gives
$\nabla V(x_{z_J}^*)=\nabla\mathcal M(z_J)$ and
$\operatorname{Lip}(\nabla V)\le L(1+\kappa)$, so pathwise
\begin{equation*}
 \|\nabla V(\widehat x)\|
 \le\|\nabla\mathcal M(z_J)\|
   +L(1+\kappa)\|\widehat u-u_{z_J}^*\|.
\end{equation*}
Squaring, taking expectations, and using
\eqref{nu:eq:main-output} and~\eqref{nu:eq:polish} proves
\[
 \E\|\nabla V(\widehat x)\|^2
 \le(1/1024+1/128)\eps^2\le\eps^2.
\]
\subsection{Oracle Costs and Completion of the Upper Bounds}
\label{nu:sec:costs}
With integer rounding, a recursive estimator with refresh
cost $M_0\ge1$, cost two per paired difference, ratio parameter $A\ge1$,
$q=\max\{1,\lfloor\sqrt{M_0/A}\rfloor\}$, and $b=\lceil Aq\rceil$
has $N_r=\lceil3T/q\rceil$ refreshes and cost
\begin{equation}\label{nu:eq:rounding}
 M_0N_r+2b(3T-N_r)\le M_0+18T\min\{M_0,\sqrt{AM_0}\}.
\end{equation}
For $q\ge2$, use $q\ge\tfrac12\sqrt{M_0/A}$,
$b\le2\sqrt{AM_0}$ and $\sqrt{AM_0}\le M_0/2$.
For $q=1$, the cost is $3TM_0$, while $M_0<4A$ implies
$\min\{M_0,\sqrt{AM_0}\}\ge M_0/2$.

Taking $M_0=n$ and $A=16C/\beta^2$ in~\eqref{nu:eq:rounding}
bounds the main cost by
$n+18T\min\{n,\sqrt{An}\}$. Since
$\beta^{-1}=O(\sqrt\kappa)$, $T=O(\sqrt\kappa S)$ and
$A=O(\kappa)$, this is
$O(n+\sqrt n\kappa S)$; the ceiling contributes at most another
$O(n)$ term. Adding the initial proximal solve and the final refinement
proves Theorem~\ref{nu:thm:normalized}. The bound holds for every $\kappa\ge1$ because
$\min\{n,\sqrt{An}\}\le\sqrt{An}$; $S\ge2$ absorbs the ceiling.

For the explicit coefficient, use $S=L\Delta/\eps^2$ and
$\beta^{-1}=(3\cdot2^{18}/\sqrt2)\sqrt\kappa$:
\[
 \beta^{-1}<2^{20}\sqrt\kappa,\qquad
 A<2^{52}\kappa,\qquad T=\lceil2^{16}S/\beta\rceil<2^{37}\sqrt\kappa S.
\]
Thus the cost of the main stage is at most
$n+2^{68}\sqrt n\kappa S$. Each solver iteration costs at most
$5/2$ in expectation, including refreshes. Since
\[
 \log(512\kappa^2)\le9\log(2\kappa),\qquad
 \log[1024\kappa^2(1+\kappa)^2]\le12\log(2\kappa),
\]
the warm and final solves together cost at most
$2n+210(n+6\sqrt n\kappa)\log(2\kappa)+5
\le2^{12}(n+\sqrt n\kappa)\log(2\kappa)$,
including both initial snapshots and both iteration ceilings. Consequently $C_N=2^{72}$
is valid in Theorem~\ref{nu:thm:normalized}.

\subsection{Assembly over All Condition Numbers}
\label{nu:sec:assembly}
Write $h(n,\kappa)=\min\{\sqrt n\kappa,n^{3/4}\sqrt\kappa\}$.
For $1\le\kappa\le\sqrt n$, $h=\sqrt n\kappa$.
For $\sqrt n\le\kappa\le2^{18}\sqrt n$,
$\frac{\sqrt n\kappa}{h(n,\kappa)} =\sqrt{\frac\kappa{\sqrt n}}\le2^9.$
Use Theorem~\ref{nu:thm:normalized} throughout
$1\le\kappa\le2^{18}\sqrt n$.
As $\log^2(2+\kappa)\ge1$ and
$\log(2\kappa)\le2\log^2(2+\kappa)$, its cost is at most
$2^{83}[n+h(n,\kappa)(1+S)]\log^2(2+\kappa).$
For $\kappa>2^{18}\sqrt n$, Appendix~\ref{app:upper} proves the upper
bound of Theorem~\ref{rb:thm:main} for every $S>0$,
with schedules that use only $n,\kappa,S$ and
$T_{\rm out}=\lceil60S\rceil$.
Taking $C=2^{83}+C_H$, where $C_H$ is the universal constant in
\eqref{eq:upper}, proves Theorem~\ref{ak:thm:upper}.
Both branches terminate almost surely with finite expected cost, and their output satisfies the expected squared-gradient criterion.
For a deterministic cap with success probability $2/3$, use target
$\eps^2/6$, cap at six times the resulting bound on the expected cost, and
apply Markov's inequality separately to cost and squared gradient norm.
The two failure probabilities are at most $1/6$ each.

\section{Matching Upper Bound Including All Initialization Costs}\label{app:upper}
We combine the shared L-SVRE solver with \citet{lin2018} and \citet{zhang2021}
to prove \eqref{eq:upperacc}--\eqref{eq:upper}. Algorithm~\ref{alg:high} lists the steps; Appendices~\ref{app:warm}--\ref{app:total} analyze them.

\begin{algorithm}[t]
\caption{Method of Theorem~\ref{rb:thm:main} for large condition numbers}\label{alg:high}
\small
\begin{algorithmic}[1]
\Require $f_i=F_i/L$, $\nu=\mu/L$, $D_0=\Delta/L$, $e_0=\eps/L$; the parameters~\eqref{eq:upperparams} and $\beta=(1-\sqrt q)/(1+\sqrt q)$; $J$ of~\eqref{eq:J}; $K$ and $\alpha$ of Appendix~\ref{app:aux}
\State $y_0\gets0$, $z_1\gets0$
\For{$k=1,\ldots,J$}\Comment{dual warm start, Appendix~\ref{app:warm}}
\State from $y=0$, run certified L-SVRE epochs (Lemma~\ref{lem:lsvre-cert}) on $g_k$ of~\eqref{eq:warmseq} until $\norm{\nabla g_k(y_k)}^2\le2s\eta_k$
\State $z_{k+1}\gets y_k+\beta(y_k-y_{k-1})$
\EndFor
\State $w_1=(p_1,y_1)\gets(0,y_J)$
\For{$t=1,\ldots,T_{\rm out}=\lceil60D_0/e_0^2\rceil$}\Comment{outer steps, Appendices~\ref{app:aux}--\ref{app:total}}
\State compute $R_t^2=\norm{\nabla \widehat f_t(w_t)}^2$ with $n$ calls; \textbf{if} $R_t=0$ \textbf{then return} $p_t$
\State $B_t\gets\nu R_t^2/72$,\quad $\widetilde w_0=(\widetilde p_0,\widetilde y_0)\gets w_t$,\quad $z_1\gets y_t$
\For{$k=1,\ldots,K$}
\State from $\widetilde w_{k-1}$, run certified L-SVRE epochs on $\widehat f_{t,k}$ of~\eqref{eq:deltatk} until $\norm{\nabla \widehat f_{t,k}(\widetilde w_k)}^2\le\delta_{t,k}$
\State $z_{k+1}\gets \widetilde y_k+\beta(\widetilde y_k-\widetilde y_{k-1})$
\EndFor
\State $w_{t+1}\gets \widetilde w_K$\Comment{$\norm{\nabla \widehat f_t(\widetilde w_K)}^2\le\alpha R_t^2$ by~\eqref{eq:uaiinner}}
\EndFor
\State \Return $p_{t+1}$ for $t$ drawn uniformly from $\{1,\ldots,T_{\rm out}\}$
\end{algorithmic}
\end{algorithm}

\subsection{Residual Certification with the Shared Solver}
\begin{lemma}[Certified L-SVRE epochs]\label{lem:lsvre-cert}
Under Lemma~\ref{nu:lem:lsvre}, restart from $r=w=u$ for
$j_0=\lceil\theta^{-1}\log8\rceil$ updates per epoch, paying $n$
calls to test $\|A(u^+)\|^2\le\delta$ afterward. If the test fails,
restart from $u^+$. For $\delta>0$, this terminates almost surely with
a certified output. Writing $H=M_s^2\|u_0-z^*\|^2/\delta$ and
$\log^+0:=0$, the cost $Q_{\rm cert}$ from the given start satisfies
\begin{equation}\label{eq:certcost}
 \begin{aligned}
 \E Q_{\rm cert}
 &\le\bigl[2n+\tfrac52\lceil\theta^{-1}\log8\rceil\bigr]
       \bigl[2+\lceil\log_2^+H\rceil\bigr]\\
 &=O\bigl((n+\sqrt nM_s/\alpha)[1+\log^+H]\bigr).
 \end{aligned}
\end{equation}
\end{lemma}
\begin{proof}
Each epoch halves conditional mean-squared distance by
\eqref{nu:eq:solver-contraction}. Continue virtually after acceptance, and let $N_{\rm ep}$ be the number of epochs.
If $N_{\rm ep}>r$, the $r$th virtual output fails its residual test;
mean Lipschitzness and Markov's inequality therefore give
$\Prob(N_{\rm ep}>r)\le\min\{1,H2^{-r}\}$ for $r\ge1$.
Summation gives $\E N_{\rm ep}\le2+\lceil\log_2^+H\rceil$ and
almost-sure termination. Each epoch costs at most
$C_{\rm ep}=2n+(5/2)j_0$ in conditional expectation, including
its initial snapshot, random refreshes, and final test.
The epoch-start indicator is measurable before that epoch. Multiply
its uniform conditional cost bound by this indicator and sum by the
tower property to obtain $\E Q_{\rm cert}\le C_{\rm ep}\E N_{\rm ep}$.
\end{proof}
Both applications below use $\alpha=s=n^{-1/2}$ and $M_s\le3$,
so their uniform conditional epoch charge is at most
\begin{equation}\label{eq:high-epoch-cost}
 C_{\rm ep}^H:=2n+\tfrac52\lceil28n\log8\rceil.
\end{equation}

\subsection{Initialization Cost and a Pathwise Certificate}\label{app:warm}
We use the normalization and parameters of Section~\ref{sec:upper}, and write $f=F/L$ and $V_f(p)=\max_yf(p,y)=V(p)/L$. For $g(y)=-f(0,y)$ let $g^*=\min_yg(y)$ and $B=D_0/\nu$. Mean strong convexity and \eqref{eq:start} imply
\[
 g(0)-g^*\le\frac{\norm{\nabla g(0)}^2}{2\nu}\le B,
 \qquad \norm{y^*(0)}^2\le\frac{2B}{\nu}.
\]
Starting at $y_0=z_1=0$, set
\begin{align}
 g_k(y)&=g(y)+\frac\tau2\norm{y-z_k}^2,
 &\eta_k&=\frac29B(1-\rho)^k,\label{eq:warmseq}\\
 z_{k+1}&=y_k+\beta(y_k-y_{k-1}),
 &\beta&=\frac{1-\sqrt q}{1+\sqrt q}.\notag
\end{align}
For each $k$, solve from $y=0$ using Lemma~\ref{lem:lsvre-cert} until the computable condition $\norm{\nabla g_k(y_k)}^2\le2s\eta_k$ holds. Each subproblem has strong convexity $s$ and RMS smoothness constant at most $1+\tau\le2$, so this certifies $g_k(y_k)-\min g_k\le\eta_k$.

The Catalyst estimate for strongly convex objectives \citep[Proposition 5]{lin2018} states that the recurrence \eqref{eq:warmseq}, with these subproblem errors and $0<\rho<\sqrt q$, satisfies
\begin{equation}\label{eq:catalyst}
 g(y_k)-g^*\le\frac{8B}{(\sqrt q-\rho)^2}(1-\rho)^{k+1}
 \le\frac{800B}{q}(1-\rho)^k.
\end{equation}
The error sum in the proof of Proposition 5 is monotone in the initial error
and potential, bounded here by $B$ and $2(1-\sqrt q)B$; prescribing
all errors with the same $B$ therefore permits this supplied upper
bound in place of the unknown initial gap.

Choose the deterministic number of accepted subproblems
\begin{equation}\label{eq:J}
 J=\left\lceil\frac{\log(1600/(q\nu^2))}{-\log(1-\rho)}\right\rceil.
\end{equation}
Then \eqref{eq:catalyst} gives $g(y_J)-g^*\le\nu D_0/2$, so $\norm{y_J-y^*(0)}^2\le D_0$ on every terminating run. Each certified subproblem terminates almost surely; the finite composition therefore yields the pathwise certificate \eqref{eq:warm} for the warm start $y_{\rm w}=y_J$.

To charge the cold starts, \eqref{eq:catalyst} and strong convexity bound $\norm{y_k-y^*(0)}^2\le1600B/(q\nu)$ for all accepted $k\le J$. Since $0\le\beta\le1$, $\norm{z_k-y^*(0)}^2\le14400B/(q\nu)$. If $y_k^*=\argmin g_k$, contraction of its proximal map under strong convexity gives
$\norm{y_k^*-y^*(0)}\le\frac{\tau}{\nu+\tau}\norm{z_k-y^*(0)}\le\norm{z_k-y^*(0)}.$
Thus $\norm{0-y_k^*}^2\le CB/(q\nu)$. The threshold is $2s\eta_k=(4/9)sB(1-\rho)^k$. Because $qs=\nu$, the ratio in \eqref{eq:certcost} is at most
$H_k\le C\nu^{-2}(1-\rho)^{-k}\le C\kappa^5.$
The last bound follows from \eqref{eq:J}, $q\ge\nu$, and $1-\rho\ge0.1$. Each subproblem costs $O(n\log(2+\kappa))$ expected calls, uniformly conditional on all previous accepted outputs. Since $J=O(q^{-1/2}\log(2+\kappa))$, startup costs
\begin{equation}\label{eq:warmcost}
 O\bigl(nq^{-1/2}\log^2(2+\kappa)\bigr).
\end{equation}

\subsection{Auxiliary Saddle Solves with Computable Tolerances}\label{app:aux}
At outer iterate $w_t=(p_t,y_t)$ let $\widehat f_t(p,y)=f(p,y)+\norm{p-p_t}^2$ and $R_t^2=\norm{\nabla \widehat f_t(w_t)}^2$. The mean is $(1,\nu)$-strongly convex--strongly concave and has joint smoothness at most three. Let $w_t^*$ be its saddle point and $G_t$ its saddle gap at $w_t$. Then
\[
 G_t\ge\frac\nu2\norm{w_t-w_t^*}^2\ge\frac{\nu R_t^2}{18},\qquad
 G_t\le\frac{R_t^2}{2\nu}.
\]
The first uses strong curvature and $R_t\le3\norm{w_t-w_t^*}$; the second applies strong convexity/concavity separately to the two gap terms. Thus the computable $B_t=\nu R_t^2/72$ satisfies \eqref{eq:gapcert}. If $R_t=0$, uniqueness of the dual maximizer gives $y_t=y^*(p_t)$ and $\nabla V_f(p_t)=0$.

Otherwise start the dual acceleration at $\widetilde w_0=(\widetilde p_0,\widetilde y_0)=w_t$, $z_1=y_t$, write $\widetilde w_k=(\widetilde p_k,\widetilde y_k)$ for its iterates, and, with the same $\beta$ as in \eqref{eq:warmseq}, form
\begin{align}
 \widehat f_{t,k}(p,y)&=\widehat f_t(p,y)-\frac\tau2\norm{y-z_k}^2,\notag\\
 \delta_{t,k}&=\frac\nu{\sqrt2}(1-\rho)^kB_t,\label{eq:deltatk}\\
 z_{k+1}&=\widetilde y_k+\beta(\widetilde y_k-\widetilde y_{k-1}).\notag
\end{align}
Solve $\widehat f_{t,k}$ from the preceding accepted $\widetilde w_{k-1}$ using Lemma~\ref{lem:lsvre-cert}, until $\norm{\nabla \widehat f_{t,k}(\widetilde w_k)}^2\le\delta_{t,k}$. Its mean strong monotonicity is $s$ and its RMS smoothness constant is at most three, because the regularizer Hessian has norm at most two and the RMS smoothness constant of $f$ is one.

Since $B_t\le G_t$, our accepted residual thresholds are no larger than
$(\nu/\sqrt2)(1-\rho)^kG_t$. The residual estimate of
\citet[Theorem 4.2]{zhang2021} therefore gives, pathwise for every
accepted finite prefix,
\begin{equation}\label{eq:uaiinner}
 \|\nabla \widehat f_t(\widetilde w_k)\|^2\le C_R(1-\rho)^kR_t^2,
 \qquad C_R=\frac{5508}{\nu^2(\sqrt q-\rho)^2}
                 +\frac{18\sqrt2}{\nu}.
\end{equation}

The same residual estimate directly controls the next solver's starting
distance. Write $a=1-\rho$ and
$A=(\nabla_p \widehat f_t,-\nabla_y \widehat f_t)$. Its $\nu$-strong monotonicity gives
\[
 d_j:=\|\widetilde w_j-w_t^*\|\le\nu^{-1}\sqrt{C_R}\,a^{j/2}R_t
 \quad(j\ge0),
\]
where $j=0$ follows from $\|A(\widetilde w_0)\|=R_t$ and $C_R\ge1$.
Writing $w_t^*=(p_t^*,y_t^*)$, the auxiliary operator
$A_k=A+(0,\tau(y-z_k))$ is $s$-strongly monotone, so
\[
 \|w_{t,k}^*-w_t^*\|
 \le(\tau/s)\|z_k-y_t^*\|\le\|z_k-y_t^*\|.
\]
For $k\ge2$, $z_k=\widetilde y_{k-1}+\beta(\widetilde y_{k-1}-\widetilde y_{k-2})$ and
$0\le\beta\le1$ consequently give
$\|\widetilde w_{k-1}-w_{t,k}^*\|\le3d_{k-1}+d_{k-2}$.
For $k=1$, $z_1=\widetilde y_0$ gives $A_1(\widetilde w_0)=A(\widetilde w_0)$. Thus
\begin{equation*}
 \|\widetilde w_{k-1}-w_{t,k}^*\|^2\le
 \begin{cases}
 R_t^2/s^2,&k=1,\\
 16C_R\nu^{-2}a^{k-2}R_t^2,&k\ge2.
 \end{cases}
\end{equation*}
Since $\delta_{t,k}=\nu^2R_t^2a^k/(72\sqrt2)$, $a\ge1/10$,
$s\ge\nu$, and
$C_R=550800/(\nu^2q)+18\sqrt2/\nu\le C\nu^{-3}$, the certification
ratio obeys
\[
 \frac{9\|\widetilde w_{k-1}-w_{t,k}^*\|^2}{\delta_{t,k}}
 \le C\kappa^7\qquad(k\ge1).
\]
Here $q=\nu/s\ge\nu$ was used. Conditional on the complete history
at each auxiliary start, all these inputs are fixed and subsequent
solver indices are fresh. Lemma~\ref{lem:lsvre-cert} therefore gives almost-sure
termination and expected cost
\[
 O\bigl((n+3\sqrt n/s)[1+\log^+(C\kappa^7)]\bigr)
 =O(n\log(2+\kappa))
\]
for that call, uniformly over the preceding accepted history, by
induction from $k=1$.

Take $\alpha=\nu^5/576$ and
$K=\left\lceil\frac{\log(C_R/\alpha)}{-\log(1-\rho)}\right\rceil =O(q^{-1/2}\log(2+\kappa)).$
Equation~\eqref{eq:uaiinner} guarantees $\norm{\nabla \widehat f_t(\widetilde w_K)}^2\le\alpha R_t^2$ on every run. The choice of $\alpha$ is no larger than the two outer tolerances in \citet[Theorem 4.1]{zhang2021}. Consequently one outer step costs
\begin{equation}\label{eq:outercost}
 O\bigl(nq^{-1/2}\log^2(2+\kappa)\bigr)
\end{equation}
in conditional expectation.

\subsection{Stationarity, Total Cost, and Fixed Confidence}\label{app:total}
Start the outer method at $(0,y_{\rm w})$ and run $T_{\rm out}=\lceil60D_0/e_0^2\rceil$ outer steps, unless an exactly zero residual permits early return. The outer theorem of \citet[Theorem 4.1]{zhang2021}, under the relative residual conditions used above, bounds the gradients at the outputs $p_{t+1}$ of the outer steps:
\begin{equation*}
 \frac1{T_{\rm out}}\sum_{t=1}^{T_{\rm out}}\norm{\nabla V_f(p_{t+1})}^2
 \le\frac{268D_0}{5T_{\rm out}}+
 \frac{28\norm{y_{\rm w}-y^*(0)}^2}{5T_{\rm out}}
 \le\frac{296D_0}{5T_{\rm out}}\le e_0^2.
\end{equation*}
This holds pathwise for certified accepted outputs. Returning $p_{t+1}$ for a uniform
$t\in\{1,\ldots,T_{\rm out}\}$ and scaling by $L$ give
\eqref{eq:upperacc} for the gradient of the unnormalized value function.

The conditional expected costs \eqref{eq:warmcost} and \eqref{eq:outercost} are uniform. The tower property therefore bounds total expected queries by
\[
 Cnq^{-1/2}(1+S)\log^2(2+\kappa)
 =Cn^{3/4}\sqrt\kappa(1+S)\log^2(2+\kappa).
\]
All auxiliary saddle gaps are finite by strong convexity--concavity.

For a deterministic cap with success probability $2/3$, run with target
squared gradient norm $\eps^2/6$. Choose a universal $C_{\rm cert}\ge1$
bounding both certification ratios above by $C_{\rm cert}\kappa^7$.
With $T_6=\lceil360D_0/e_0^2\rceil$, a valid expected-cost bound is
\begin{equation}\label{eq:high-confidence-cost}
 B_Q=C_{\rm ep}^H(J+T_6K)
       [2+\lceil\log_2^+(C_{\rm cert}\kappa^7)\rceil]+nT_6.
\end{equation}
This uses the L-SVRE epoch charge \eqref{eq:high-epoch-cost} and includes
the residual $R_t$ computed at each outer step. Cap execution at $6B_Q$, returning the
origin on timeout; Markov's inequality then bounds the failure probability
by $1/3$ as in Appendix~\ref{nu:sec:assembly}, at the same rate.

\bibliography{references}

@inproceedings{zhang2021,
 author={Zhang, Siqi and Yang, Junchi and Guzm\'an, Crist\'obal and Kiyavash, Negar and He, Niao},
 title={The Complexity of Nonconvex-Strongly-Concave Minimax Optimization},
 booktitle={Proceedings of the Thirty-Seventh Conference on Uncertainty in Artificial Intelligence},
 series={Proceedings of Machine Learning Research}, volume={161}, pages={482--492}, year={2021}
}

@inproceedings{li2021,
 author={Li, Haochuan and Tian, Yi and Zhang, Jingzhao and Jadbabaie, Ali},
 title={Complexity Lower Bounds for Nonconvex-Strongly-Concave Min-Max Optimization},
 booktitle={Advances in Neural Information Processing Systems}, volume={34}, year={2021}
}

@article{han2024,
 author={Han, Yuze and Xie, Guangzeng and Zhang, Zhihua},
 title={Lower Complexity Bounds of Finite-Sum Optimization Problems: The Results and Construction},
 journal={Journal of Machine Learning Research}, volume={25}, number={2}, pages={1--86}, year={2024}
}

@article{lin2018,
 author={Lin, Hongzhou and Mairal, Julien and Harchaoui, Zaid},
 title={Catalyst Acceleration for First-Order Convex Optimization: From Theory to Practice},
 journal={Journal of Machine Learning Research}, volume={18}, number={212}, pages={1--54}, year={2018}
}

@article{carmon2020,
 author={Carmon, Yair and Duchi, John C. and Hinder, Oliver and Sidford, Aaron},
 title={Lower Bounds for Finding Stationary Points {I}},
 journal={Mathematical Programming}, volume={184}, pages={71--120}, year={2020},
 doi={10.1007/s10107-019-01406-y}
}

@inproceedings{woodworth2016,
 author = {Woodworth, Blake E. and Srebro, Nati},
 title = {Tight Complexity Bounds for Optimizing Composite Objectives},
 booktitle = {Advances in Neural Information Processing Systems},
 volume = {29},
 year = {2016},
 note = {Full version: arXiv:1605.08003v3}
}

@inproceedings{bubeck2019,
 author = {Bubeck, S{\'e}bastien and Jiang, Qijia and Lee, Yin-Tat and Li, Yuanzhi and Sidford, Aaron},
 title = {Complexity of Highly Parallel Non-Smooth Convex Optimization},
 booktitle = {Advances in Neural Information Processing Systems},
 volume = {32},
 year = {2019},
 note = {Full version: arXiv:1906.10655v2}
}

@misc{wu2026ncc,
 author = {Wu, Qilong and Gu, Zhihao and Yang, Junchi},
 title = {Lower Bounds for Nonconvex-Concave Minimax Optimization},
 year = {2026},
 howpublished = {arXiv:2609.14233v1}
}

@article{arjevani2023,
 author={Arjevani, Yossi and Carmon, Yair and Duchi, John C. and Foster, Dylan J. and Srebro, Nathan and Woodworth, Blake},
 title={Lower Bounds for Non-Convex Stochastic Optimization},
 journal={Mathematical Programming}, volume={199}, pages={165--214}, year={2023},
 doi={10.1007/s10107-022-01822-7},
 note={Revised full version: arXiv:1912.02365v2}
}

@misc{zhang2026spde,
 author={Zhang, Huiling and Zhang, Minhao and Xu, Zi},
 title={Single-Loop Stochastic Projected Damped Extragradient Methods for Stochastic Nonconvex--(Strongly) Concave Minimax Optimization},
 year={2026}, howpublished={arXiv:2609.21747v1}
}

@misc{zhou2026ncsc,
 author={Zhou, Qihao},
 title={Near-Optimal Complexity of Finite-Sum Nonconvex-Strongly-Concave Minimax Optimization},
 year={2026}, howpublished={arXiv:2610.04944v1}
}

@inproceedings{alacaoglu2022,
  author={Alacaoglu, Ahmet and Malitsky, Yura},
  title={Stochastic Variance Reduction for Variational Inequality Methods},
  booktitle={Proceedings of Thirty Fifth Conference on Learning Theory}, pages={778--816}, series={Proceedings of Machine Learning Research}, volume={178}, year={2022}
}

@misc{luo2022,
  author={Luo, Luo and Xie, Guangzeng and Zhang, Tong and Zhang, Zhihua},
  title={Near Optimal Stochastic Algorithms for Finite-Sum Unbalanced Convex-Concave Minimax Optimization},
  howpublished={arXiv:2106.01761v2}, year={2022}
}

@misc{peng2026dense,
  author = {Peng, Yuxing and Tang, Zhiqing and Jia, Weijia},
  title = {Dense Weak Hiding: Closing Complexity Gaps in Nonconvex and {PL} Finite-Sum Optimization under Individual Smoothness},
  year = {2026}, howpublished = {arXiv:2609.00045v3}
}

@misc{wu2026ncfinite,
  author={Wu, Wendao and Zhang, Haihan and Chen, Qian and Zhang, Chenheng and Zheng, Chunyuan and Fang, Cong and Li, Haoxuan and Lin, Zhouchen},
  title={Matching Randomized Bounds for Finite-Sum Nonconvex-Concave Optimization},
  year={2026}, howpublished={arXiv:2610.16991v1}
}

@misc{zhou2026inclusion,
  author={Zhou, Qihao},
  title={Optimal Oracle Complexity for Finite-Sum Monotone Inclusions},
  year={2026}, howpublished={arXiv:2610.05038v1}
}

@inproceedings{madry2018,
  author={Madry, Aleksander and Makelov, Aleksandar and Schmidt, Ludwig and Tsipras, Dimitris and Vladu, Adrian},
  title={Towards Deep Learning Models Resistant to Adversarial Attacks},
  booktitle={International Conference on Learning Representations}, year={2018},
  note={arXiv:1706.06083}
}

@inproceedings{namkoong2016,
  author={Namkoong, Hongseok and Duchi, John C.},
  title={Stochastic Gradient Methods for Distributionally Robust Optimization with $f$-divergences},
  booktitle={Advances in Neural Information Processing Systems}, volume={29}, year={2016}
}

@inproceedings{mohri2019,
  author={Mohri, Mehryar and Sivek, Gary and Suresh, Ananda Theertha},
  title={Agnostic Federated Learning},
  booktitle={Proceedings of the 36th International Conference on Machine Learning},
  series={Proceedings of Machine Learning Research}, volume={97}, pages={4615--4625}, year={2019}
}

@inproceedings{lin2020gda,
  author={Lin, Tianyi and Jin, Chi and Jordan, Michael},
  title={On Gradient Descent Ascent for Nonconvex-Concave Minimax Problems},
  booktitle={Proceedings of the 37th International Conference on Machine Learning},
  series={Proceedings of Machine Learning Research}, volume={119}, pages={6083--6093}, year={2020}
}

@inproceedings{luo2020sreda,
  author={Luo, Luo and Ye, Haishan and Huang, Zhichao and Zhang, Tong},
  title={Stochastic Recursive Gradient Descent Ascent for Stochastic Nonconvex-Strongly-Concave Minimax Problems},
  booktitle={Advances in Neural Information Processing Systems}, volume={33}, pages={20566--20577}, year={2020}
}

@inproceedings{nguyen2017sarah,
 author={Nguyen, Lam M. and Liu, Jie and Scheinberg, Katya and Tak\'a\v{c}, Martin},
 title={{SARAH}: A Novel Method for Machine Learning Problems Using Stochastic Recursive Gradient},
 booktitle={Proceedings of the 34th International Conference on Machine Learning},
 series={Proceedings of Machine Learning Research}, volume={70}, pages={2613--2621}, year={2017}
}

@inproceedings{fang2018spider,
 author={Fang, Cong and Li, Chris Junchi and Lin, Zhouchen and Zhang, Tong},
 title={{SPIDER}: Near-Optimal Non-Convex Optimization via Stochastic Path-Integrated Differential Estimator},
 booktitle={Advances in Neural Information Processing Systems}, volume={31}, year={2018}
}

@book{vershynin2018,
 author={Vershynin, Roman},
 title={High-Dimensional Probability: An Introduction with Applications in Data Science},
 publisher={Cambridge University Press}, year={2018},
 doi={10.1017/9781108231596}
}

@article{rockafellar1976,
 author={Rockafellar, R. Tyrrell},
 title={Monotone Operators and the Proximal Point Algorithm},
 journal={SIAM Journal on Control and Optimization},
 volume={14}, number={5}, pages={877--898}, year={1976},
 doi={10.1137/0314056}
}

@article{davis2019,
 author={Davis, Damek and Drusvyatskiy, Dmitriy},
 title={Stochastic Model-Based Minimization of Weakly Convex Functions},
 journal={SIAM Journal on Optimization},
 volume={29}, number={1}, pages={207--239}, year={2019},
 doi={10.1137/18M1178244}
}

@inproceedings{yao1977,
 author={Yao, Andrew Chi-Chih},
 title={Probabilistic Computations: Toward a Unified Measure of Complexity},
 booktitle={Proceedings of the 18th Annual Symposium on Foundations of Computer Science},
 pages={222--227}, year={1977},
 doi={10.1109/SFCS.1977.24}
}

\section*{AI Usage Statement}
We used AI assistants, including OpenAI's GPT models and Anthropic's Claude, to help develop and check proofs, formalize them in Lean, and edit the text. The authors take full responsibility for the content of this paper.
\end{document}